\documentclass[11pt,reqno]{amsart}
\usepackage[margin=1in]{geometry}

\newcommand{\DateOfPub}[1]{}

\usepackage[UKenglish]{babel}

\usepackage{amsfonts,amsmath,amsthm,amssymb} 
\usepackage{mathrsfs} 
\usepackage{stmaryrd} 
\usepackage{mathtools}
\usepackage{bbm}

\usepackage{lscape} 

\usepackage{natbib}
\usepackage{url}

\usepackage{tikz-cd}
\usepackage{ascmac}
\usepackage{soul}

\usetikzlibrary{patterns}

\usepackage{comment} 
\usepackage{graphicx} 
\usepackage{color}

\usepackage{multirow}

\allowdisplaybreaks[1]

\newtheorem{theorem}{Theorem}
\newtheorem{proposition}[theorem]{Proposition}
\newtheorem{lemma}[theorem]{Lemma}
\newtheorem{corollary}[theorem]{Corollary}

\theoremstyle{remark}
\newtheorem{remark}{Remark}

\theoremstyle{definition}
\newtheorem{definition}{Definition}
\newtheorem{assumption}{Assumption}

\newcommand\EatDot[1]{}

\newcommand{\bX}{\boldsymbol{X}}

\newcommand{\bZ}{\boldsymbol{Z}}

\newcommand{\bg}{\boldsymbol{g}}

\newcommand{\bt}{\boldsymbol{t}}

\newcommand{\bw}{\boldsymbol{w}}
\newcommand{\bx}{\boldsymbol{x}}
\newcommand{\by}{\boldsymbol{y}}
\newcommand{\bz}{\boldsymbol{z}}

\newcommand{\btheta}{\boldsymbol{\theta}}

\newcommand{\bmu}{\boldsymbol{\mu}}

\newcommand{\bomega}{\boldsymbol{\bomega}}

\newcommand{\bfA}{\mathbf{A}}

\newcommand{\bfI}{\mathbf{I}}

\newcommand{\bSigma}{\mathbf{\Sigma}}

\newcommand{\bbS}{\mathbb{S}}

\newcommand{\calA}{\mathcal{A}}

\newcommand{\calC}{\mathcal{C}}

\newcommand{\calG}{\mathcal{G}}

\newcommand{\calL}{\mathcal{L}}

\newcommand{\calN}{\mathcal{N}}
\newcommand{\calO}{\mathcal{O}}

\newcommand{\calS}{\mathcal{S}}
\newcommand{\calT}{\mathcal{T}}

\newcommand{\calZ}{\mathcal{Z}}

\newcommand{\N}{\mathbb{N}}

\newcommand{\R}{\mathbb{R}}

\newcommand{\Z}{\mathbb{Z}}

\newcommand{\E}{\mathbb{E}}
\renewcommand{\Pr}{\mathbb{P}}

\newcommand{\Var}{\textnormal{Var}}

\newcommand{\Gauss}{\mathcal{N}}
\newcommand{\Unif}{\textnormal{Unif}}

\newcommand{\Ent}{\textnormal{Ent}}

\newcommand{\tr}{\textnormal{tr}}

\newcommand{\op}{\textnormal{op}}

\newcommand{\diff}{\textnormal{d}}
\newcommand{\Leb}{\textnormal{Leb}}

\renewcommand{\epsilon}{\varepsilon}
\newcommand{\sgn}{\textnormal{sgn}}
\newcommand{\zero}{\mathbf{0}}
\DeclareMathOperator*{\argmin}{arg\,min}

\newcommand{\Poincare}{\textnormal{P}}
\newcommand{\LogSobolev}{\textnormal{LS}}
\newcommand{\KLS}{\textnormal{KLS}}

\newcommand{\Risk}{\mathcal{R}}

\usepackage[justification=centering]{caption}

\title[Generalization bounds for classification]{Improved generalization bounds for binary linear classification via isoperimetry}
\author[S Nakakita]{Shogo Nakakita}
\address{Komaba Institute for Science, University of Tokyo, 3-8-1 Komaba, Meguro-ku, Tokyo 153-8902, Japan}
\date{}

\usepackage{newtxtext,newtxmath}

\begin{document}

\begin{abstract}
We examine the concentration of uniform generalization errors around their expectation in binary linear classification problems via an isoperimetric argument.
In particular, we establish Poincar\'{e} and log-Sobolev inequalities for the joint distribution of the output labels and the label-weighted input vectors, which we apply to derive concentration bounds.
The derived results improve upon existing bounds obtained from general unbounded empirical processes, as well as that tailored specifically to logistic regression.
In asymptotic analysis, we also show that almost sure convergence of uniform generalization errors to their expectation occurs in very broad settings, such as proportionally high-dimensional regimes.
Using this convergence, we establish uniform laws of large numbers under dimension-free conditions.
\end{abstract}

\maketitle

\section{Introduction}

Binary classification is one of the most fundamental tasks in machine learning and statistics.
It concerns estimating a high-performing model for predicting a $\{\pm1\}$-valued random variable $Y$ (output label) from its corresponding $d$-dimensional random vector $\bX$ (input vector) from a sequence of $n$ independent and identically distributed random variables $\{(\bX_{i},Y_{i})\}_{i=1}^{n}$ whose distributions are the same as $(\bX,Y)$.
Binary linear classifiers are fundamental models in this task; we estimate a reasonable  $\hat{\bw}\in\R^{d}$ by the training data and use $\sgn(\langle \bX,\hat{\bw}\rangle)$ as a prediction for the corresponding value of $Y$.
There are multiple approaches to estimating $\hat{\bw}$ from training data; in this paper, we consider an empirical risk minimization problem such that
\begin{equation}\label{eq:empirical}
    \text{minimize }\Risk_{n}(\bw,b)\left(=\frac{1}{n}\sum_{i=1}^{n}\ell\left(Y_{i}\left(\langle \bX_{i},\bw\rangle+b\right)\right)\right)\text{ such that }\|\bw\|\le R_{\bw},\ |b|\le R_{b},
\end{equation}
where $\ell:\R\to[0,+\infty)$ is an $L$-Lipschitz continuous function (for some $L>0$), $\|\cdot\|$ is the Euclidean norm, $R_{\bw}>0$ is the radius of the space of $\bw$, and $R_{b}\ge 0$ is the radius of the space of $b$.
We use the notation $\calT:=\{\bw\in\R^{d}:\|\bw\|\le R_{\bw}\}\times[-R_{b},R_{b}]\subset\R^{d+1}$.
We refer to $\Risk_{n}$ as the empirical risk function and $\ell$ as the loss function.
The logistic loss function $\ell_{\textnormal{L}}(t)=\log\left(1+\exp\left(-t\right)\right)$ and hinge loss function $\ell_{\textnormal{H}}(t)=\max\{0,1-t\}$ serve as prime examples of Lipschitz loss functions.
A basic idea of these loss functions is that their subderivatives satisfy $\ell'(t)\le 0$, and thus they penalize the difference in the signs of $Y_{i}$ and $\langle \bX_{i},\bw\rangle+b$.
Hence, we can interpret the problem \eqref{eq:empirical} as the selection of $(\bw,b)$ so that the signs of $Y_{i}$ and $\langle \bX_{i},\bw\rangle+b$ are the same as possible.

A motivation for the empirical risk minimization \eqref{eq:empirical} is that we expect that the problem should be close to the following population risk minimization problem:
\begin{equation}\label{eq:population}
    \text{minimize }\Risk(\bw,b)\left(=\E\left[\ell\left(Y(\langle \bX,\bw\rangle+b)\right)\right]\right)\text{ such that }\|\bw\|\le R_{\bw},\ |b|\le R_{b},
\end{equation}
where $\Risk(\bw,b)$ is referred to as a population risk function.
Under a mild condition, we know that $\Risk_{n}(\bw,b)\to\Risk(\bw,b)$ almost surely for every $(\bw,b)\in\R^{d+1}$, which is a \emph{pointwise} law of large numbers.
However, we need a \emph{uniform} law on $\calT$ to verify that the optimization problem \eqref{eq:empirical} is indeed close to \eqref{eq:population}.
In particular, the following type of concentration inequalities for given $\delta\in(0,1]$ has been widely used to characterize the discrepancy between them in studies of machine learning and statistics:
\begin{equation*}
    \Pr\left(\sup_{(\bw,b)\in\calT}\left(\Risk(\bw,b)-\Risk_{n}(\bw,b)\right)\ge (\text{residual dependent on }(\delta,n,R_{\bw},R_{b}))\right)\le \delta.
\end{equation*}
The term $\sup_{(\bw,b)\in\calT}(\Risk(\bw,b)-\Risk_{n}(\bw,b))$ is referred to as a \emph{uniform generalization error} or \emph{worst-case generalization error}.
If the residual dependent on $(\delta,n,R_{\bw},R_{b})$ takes small values even for small $\delta>0$, we conclude that the problem \eqref{eq:empirical} is close to \eqref{eq:population}.

In this work, we analyse the discrepancy between the problems \eqref{eq:empirical} and \eqref{eq:population} by showing the following concentration of a uniform generalization error around its expectation:
\begin{equation}\label{eq:ConcAroundExp}
    \Pr\left(\sup_{(\bw,b)\in\calT}\left(\Risk(\bw,b)-\Risk_{n}(\bw,b)\right)\ge \E\left[\sup_{(\bw,b)\in\calT}\left(\Risk(\bw,b)-\Risk_{n}(\bw,b)\right)\right]+\text{(residual)}_{\delta,n,R_{\bw},R_{b}}\right)\le \delta.
\end{equation}
Given a large $n$, we show that uniform generalization errors concentrate around their expectation in many settings, and the residual terms tend to zero.
Hence, our strategy for analysing a uniform generalization error is a divide-and-conquer one; we characterize the behaviour of a uniform generalization error in terms of its expectation, which may be small or large.
If both the expectation and residual converge to zero, then we conclude that the empirical risk minimization problem \eqref{eq:empirical} is a good alternative to the population risk minimization problem \eqref{eq:population}.
On the other hand, if the residual converges to zero, but the expectation does not, then the empirical risk minimization problem does not serve as a good approximation of the population risk minimization, and we can identify that the cause of the difficulty is not the random behaviour of uniform generalization errors but the essential difficulty represented by expected uniform generalization errors.

Such a strategy is reasonable and even classical since the evaluation of expected uniform generalization errors is a classical problem in statistical learning theory.
For example, the Rademacher complexity argument is a very established strategy to give estimates for the expectation \citep{ledoux1991probability,bartlett2002rademacher,wainwright2019high,bach2024learning}, and it is straightforward to derive such a bound in our settings (see Proposition \ref{prop:rademacher}).
Therefore, estimates of the residual term are crucial for evaluating the stochastic behaviour of uniform generalization errors precisely.

The boundedness of loss functions is a widely employed assumption for deriving clean estimates of the residuals.
Under this assumption, one can directly apply McDiarmid's inequality to obtain an explicit and simple residual term $\|\ell\|_{\infty}\sqrt{\log(\delta^{-1})/2n}$ \citep[e.g., see][]{bach2024learning}.
On the other hand, although the boundedness of $\ell$ yields a tractable statistical complexity estimate for the problem, this assumption excludes typical loss functions that render the empirical risk minimization problem \eqref{eq:empirical} a convex optimization problem.
Intuitively, this assumption on $\ell$ makes the problem statistically tractable but computationally intractable.

Therefore, we are motivated to examine the concentration of uniform generalization errors under unbounded loss functions that render $\Risk_n(\bw,b)$ convex.
The logistic and hinge loss functions are typical examples of such loss functions; they correspond to logistic regression and the soft-margin support vector machine, which are central topics in binary linear classification within machine learning.
These problems are computationally tractable, but their statistical complexities have been scarcely investigated.
Although the concentration for general unbounded loss functions has been studied \citep{adamczak2008tail,van2013bernstein,lederer2014new}, previous bounds centre around the scaled expectations of uniform generalization errors and contain residual terms that may diverge in realistic scenarios (see Section \ref{sec:discussion}).
In particular, since the motivating loss functions are Lipschitz continuous, we focus on Lipschitz continuous $\ell$ to circumvent the difficulties encountered in previous studies on empirical processes with unbounded loss functions.

\subsection{Main idea: Isoperimetry in binary linear classification}
This study establishes that uniform generalization errors in binary linear classification problems concentrate rapidly around their expectation, even if loss functions are unbounded.
To obtain residuals significantly smaller than the expectation of uniform generalization errors, we employ isoperimetric properties derived from functional inequalities (Poincar\'{e} and log-Sobolev) for the joint distributions of $(Y_{i}\bX_{i},Y_{i})=:(\bZ_{i},Y_{i})$.
Let us introduce an intuitive example.
Assume that $\bX_{i}\sim\calN(\zero_{d},\bSigma)$ with $\|\bSigma\|_{\text{op}}=1$ and $Y_{i}\sim\text{Unif}(\{\pm1\})$ are independent random variables.
Then, the joint distribution of $(\bZ_i, Y_i)$ coincides with that of $(\bX_i, Y_i)$.
Owing to the independence and tensorization argument \citep{bakry2014analysis}, we have the following logarithmic Sobolev inequality: for any $f:(\R^{d}\times\{\pm1\})^{n}$ being smooth in the Euclidean arguments,
\begin{equation*}
    \Ent[f^{2}((\bZ_{i},Y_{i})_{i=1}^{n})]\le 2\E\left[\Gamma(f)((\bZ_{i},Y_{i})_{i=1}^{n})\right],
\end{equation*}
where $\Ent[f^{2}]=\E[f^{2}\log f^{2}]-\E[f^{2}]\log\E[f^{2}]$ is the entropy operator, 
\begin{equation*}
    \Gamma(f)((\bz_{i},y_{i})_{i=1}^{n})=\sum_{i=1}^{n}\left(\|\nabla_{\bz_{i}}f((\bz_{i},y_{i})_{i=1}^{n})\|^{2}+|\nabla_{y_{i}}f((\bz_{i},y_{i})_{i=1}^{n})|^{2}\right)
\end{equation*}
is a carr\'{e} du champ operator,
$\nabla_{\bz_{i}}$ is the gradient operator with respect to $\bz_{i}$, and $\nabla_{y_{i}}$ is a discrete gradient operator with respect to $y_{i}$ such that $\nabla_{y_{i}}f((\bz_{i},y_{i})_{i=1}^{n})=(f((\bz_{i},y_{i})_{i=1}^{n})-f((\bz_{j},y_{j})_{j=1}^{i-1},(\bz_{i},-y_{i}),(\bz_{j},y_{j})_{j=i+1}^{n}))/2$.
Then, we have
\begin{equation*}
    \left\|\Gamma\left(\sup_{(\bw,b)\in\calT}\left(\Risk(\bw,b)\right)-\frac{1}{n}\sum_{i=1}^{n}\ell(\langle \bz_{i},\bw\rangle+y_{i}b)\right)\right\|_{\infty}\le \frac{L^{2}(R_{\bw}^{2}+R_{b}^{2})}{n};
\end{equation*}
that is, the empirical risk function is $L\sqrt{(R_{\bw}^{2}+R_{b}^{2})/n}$-Lipschitz (with respect to the carr\'{e} du champ operator $\Gamma$).
Noting that $\ell(Y_{i}(\langle \bX_{i},\bw\rangle+b))=\ell(\langle \bZ_{i},\bw\rangle+bY_{i})$ and extending Herbst's argument (see Section \ref{sec:functional:tail}), we obtain the following concentration: for any $\delta\in(0,1]$ with probability at least $1-\delta$,
\begin{align*}
    \sup_{(\bw,b)\in\calT}(\Risk(\bw,b)-\Risk_{n}(\bw,b))&\le \E\left[\sup_{(\bw,b)\in\calT}(\Risk(\bw,b)-\Risk_{n}(\bw,b))\right]\\
    &\quad+\sqrt{\frac{L^{2}(2R_{\bw}^{2}+eR_{b}^{2})\log(\delta^{-1})}{n}}+\frac{4LR_{b}\log(\delta^{-1})}{n}.
\end{align*}
Although $\bZ_{i}$ and $Y_{i}$ are not independent in general, we expect that suitable isoperimetric properties of the joint distributions of $(\bZ_{i}, Y_{i})$ will enable the extension of this argument.

Therefore, the problem is when the joint distributions of $\bZ_{i}$ and $Y_{i}$ are isoperimetric in the sense of Poincar\'{e} or log-Sobolev inequalities.
To develop such functional inequalities, we extend functional inequalities for mixture distributions, which have gathered interest recently \citep{bardet2018functional,schlichting2019poincare,chen2021dimension}.
Let us restrict ourselves to the case with $R_{b}=0$ to obtain an intuition; we need functional inequalities for the marginal distribution of $\bZ_{i}$ in such a setting.
The marginal distribution is a weighted sum of the conditional distributions of $\bZ_{i}$ given $Y_{i}=y\in\{\pm1\}$, which is a mixture distribution.
In general, if the joint state space of two random variables has a good geometric structure (such as an Euclidean space or Riemannian manifold), we can establish functional inequalities for the joint distribution of the variables via standard arguments like the Bakry--{\'E}mery condition, and its marginalization is trivial.
Hence, the fact that marginal distributions are mixture distributions is not problematic under good geometric structures.
However, we need functional inequalities for a marginal distribution of a measure on hybrid continuous--discrete state spaces, which are not manifolds; alternative geometric arguments on such hybrid spaces are not readily available.
As a direct strategy to solve this problem, we employ recent results of functional inequalities for mixture distributions.
More precisely, considering $R_{b}>0$, we extend them to functional inequalities for the joint distribution of continuous and discrete random variables.

\subsection{Contributions}
We derive novel dimension-free uniform concentration bounds around the exact expectation (without any scaling factor) with explicit constants and light tails. 
These bounds improve upon both those derived through the application of dimension-free uniform concentration bounds for general unbounded empirical processes \citep{adamczak2008tail,van2013bernstein,lederer2014new} and that tailored specifically to the logistic regression problem \citep{nakakita2026dimension}; see Table \ref{tab:comparison_bounds}.
\citet{adamczak2008tail} provides light-tailed concentration bounds around expectation with a scaling factor; however, applying this general result to our specific setting leaves the numerical constant implicit, and the remainder term can diverge under certain conditions (see Section \ref{sec:discussion:advantages}).
Although \citet{van2013bernstein} overcome these limitations of \citet{adamczak2008tail}, their concentration bound is centred not around the expectation of the supremum of empirical processes, but rather around a value bounded below by that expectation; furthermore, the construction of generalized brackets is highly non-trivial.
While \citet{lederer2014new} provide a concentration bound centred around expectation (with a scaling factor) with explicit numerical constants, their approach results in a heavy polynomial tail for the derived concentration inequalities.
\citet{nakakita2026dimension}, which is most closely related to our work, derives a dimension-free uniform concentration bound for logistic regression with explicit numerical constants under the isoperimetric property of the marginal distribution $\bX_i$. Although this property is weaker than our assumption, their bound is restricted exclusively to logistic regression, features larger numerical constants than our bounds, and only establishes the concentration of uniform generalization errors around zero.
\begin{table}[htbp]
    \centering
    \begin{tabular}{lccc}
        \hline
        \multirow{2}{*}{Study} & Centred around & Explicit & Light \\
        & expectation & constants & tails \\
        \hline
        \citet{adamczak2008tail}       & \checkmark (up to scale) & $\times$   & \checkmark \\
        \citet{van2013bernstein}       & $\times$   & \checkmark & \checkmark \\
        \citet{lederer2014new}         & \checkmark  (up to scale) & \checkmark & $\times$   \\
        \citet{nakakita2026dimension}  & $\times$   & \checkmark & \checkmark \\
        \textbf{This work}             & \checkmark & \checkmark & \checkmark \\
        \hline
    \end{tabular}
    \vskip 0.05in
    \caption{Comparison of uniform concentration bounds under our problem setup. The term ``light tails'' refers to bounds characterized by subexponential or subgaussian decay rates. ``Up to scale'' implies that the bound centres around the expectation multiplied by a certain scaling factor.}
    \label{tab:comparison_bounds}
\end{table}
 
\subsection{Literature review}
Asymptotic/non-asymptotic behaviours of high-dimensional binary linear classification problems have received much attention in this decade.
Studies on this topic can be mainly classified into two categories: (i) relatively low-dimensional settings $d/n\to0$ or $d^{\ast}/n\to0$ for some intrinsic dimension $d^{\ast}$ \citep[e.g.,][and references therein]{kuchelmeister2024finite,hsu2024sample,nakakita2026dimension}, and (ii) proportional high-dimensional settings $d/n\to \kappa\in(0,\infty)$ \citep[e.g.,][]{sur2019modern,sur2019likelihood,salehi2019impact,candes2020phase,liang2022precise,montanari2025generalization}.
Typical examples of $d^{\ast}$ are the number of non-zero signals in sparse settings and an effective rank $\tr(\E[\bX_{i}\bX_{i}^{\top}])/\|\E[\bX_{i}\bX_{i}^{\top}]\|_{\textnormal{op}}$ in non-sparse settings.
In particular, $d^{\ast}$ can be much smaller than $d$, the dimension of the ambient space, and it characterizes the problem better than $d$ itself.
Our results in the setting $d/n\to0$ or $d^{\ast}/n\to0$ yield powerful conclusions, uniform laws of large numbers; hence, this study is well-contextualized in a series of studies with $d/n\to0$ or $d^{\ast}/n\to0$ (see Section \ref{sec:appl:ulln}).
Moreover, our results can also be applied to asymptotic studies in proportionally high-dimensional settings; the deviation of the empirical risk minimization problem from the population one under $d/n\to\kappa\in(0,\infty)$ can be partially characterized by expected uniform generalization errors (see Section \ref{sec:appl:proportional}).

Let us review studies for relatively low-dimensional settings in detail.
\citet{liang2012maximum} guarantee the asymptotic normality of maximum likelihood estimation in logistic regression under $d/n\to0$.
\citet{kuchelmeister2024finite} identify two distinct rates of convergence of maximum likelihood estimation in logistic regression in small and large noise schemes, the probit model as a data generating model, and $d/n\to0$.
\citet{hsu2024sample} show the phase transition of the sample complexities for logistic regression in three temperature settings and $d/n\to0$.
\citet{van2008high} gives non-asymptotic guarantees for generalized linear models including logistic regression, and the bounds converge to zero under $d^{\ast}/n\to0$ (ignoring logarithmic factors for simplicity), where $d^{\ast}$ is the number of non-zero elements of the true value of the parameter.
\citet{nakakita2026dimension} gives a dimension-free concentration bound for uniform generalized errors around zero in logistic regression, which converges under $d^{\ast}/n\to0$, where $d^{\ast}$ is the effective rank of the covariance matrix of $\bX_{i}$.
Our results resemble the result of \citet{nakakita2026dimension} but have some essential differences; we compare our results with it in Section \ref{sec:appl:ulln}.

It is remarkable that a number of studies in high-dimensional binary linear classification problems do not address the bias of $Y_{i}$ (that is to say, they restrict themselves to the settings $R_{b}=0$ and $\Pr(Y_{i}=+1)=\Pr(Y_{i}=-1)=1/2$).
Bias is the most fundamental parameter in the prediction of $Y$ given a new input vector $\bX$, and large prediction errors can occur if we ignore it. 
If the distribution of $\bX$ is sign-invariant, then for any $\hat{\bw}$ independent of $\bX$, $\Pr(\sgn(\langle \bX,\hat{\bw}\rangle)=+1)=\Pr(\sgn(\langle \bX,\hat{\bw}\rangle)=-1)=1/2$; then, the prediction results in large errors if $\Pr(Y=+1)\neq\Pr(Y=-1)$.
This study is not the first one enabling us to characterize such problems in the presence of both bias and high dimensionality \citep[e.g.,][can address it by considering non-centred input vectors]{nakakita2026dimension}, but is one of the few studies to estimate the effect of biases under high-dimensional settings explicitly.

Beyond the context of binary classification problems, concentration inequalities under dependence also provide a relevant background for our study, as the derived concentration inequalities concern the joint distributions of dependent random variables $(\bZ_i,Y_i)$.
Previous studies can be categorized into four methodological approaches: the martingale method \citep{kontorovich2008concentration}, the mixing or spectral gap method \citep{paulin2015concentration,fan2021hoeffding,merlevede2011bernstein}, the functional inequality method \citep{nakakita2025corrigendum,nakakita2026sample}, and the information-theoretic method \citep{esposito2024concentration}.
Among studies on functional inequalities for the joint distribution of dependent random variables, the works by \citet{nakakita2025corrigendum} and \citet{nakakita2026sample} are the most closely related to this study. They investigate log-Sobolev inequalities for the joint distributions of dependent stochastic processes, such as causal Bernoulli shift processes and the outputs of the unadjusted Langevin algorithm.
Although our study assumes that $(\bZ_i,Y_i)$ and $(\bZ_j,Y_j)$ with $i\neq j$ are independent, it shares a common focus with these previous works in terms of carefully examining dependence structures.

\subsection{Paper organization}
Section \ref{sec:preliminaries} provides notation and definitions used in this paper.
Section \ref{sec:functional} exhibits functional inequalities for distributions on $\R^{d}\times\{\pm1\}$, which are the main technical contributions of this study, and their applications to tail bounds.
Section \ref{sec:concentration} gives concentration inequalities for uniform generalization errors in binary linear classification problems.
Section \ref{sec:appl} is for applications of our non-asymptotic results to asymptotic studies in two settings: a dimension-free regime and proportionally high-dimensional regime.
Section \ref{sec:discussion} is devoted to discussions regarding the advantages and remaining problems of our study.
Section \ref{sec:concluding} gives some concluding remarks and possible future directions of research.
All the technical proofs are deferred to the Appendix.

\section{Notation and definitions} \label{sec:preliminaries}
In this section, we introduce notation and definitions used throughout this paper.

For any matrix $\bfA\in\R^{p\times q}$, $\|\bfA\|_{\textnormal{op}}:=\sup_{\bx\in\R^{q};\|\bx\|\neq0}\|\bfA\bx\|/\|\bx\|$.
We let $\Leb(\cdot)$ and $\sharp(\cdot)$ denote the Lebesgue measure on $\R^{d}$ and the counting measure on $\{\pm1\}$.
For $f:\R^{d}\times\{\pm1\}\to\R$, we define a discrete gradient $\nabla_{y}f(\bz,y):=(f(\bz,y)-f(\bz,-y))/2$.
For $f:(\R^{d}\times\{\pm1\})^{n}\to\R$, we also define a discrete gradient $\nabla_{y_{i}}f((\bz_{i},y_{i})_{i=1}^{n}):=(f((\bz_{i},y_{i})_{i=1}^{n})-f((\bz_{j},y_{j})_{j=1}^{i-1},(\bz_{i},-y_{i}),(\bz_{j},y_{j})_{j=i+1}^{n})))/2$.
$\calA_{1}$ denotes the family of all functions $f:\R^{d}\times\{\pm1\}\to\R$ (resp.\ $f:(\R^{d}\times\{\pm1\})^{n}\to\R$) such that for both $y\in\{\pm1\}$, $f(\bz,y)$ is locally Lipschitz continuous with respect to $\bz\in\R^{d}$.
Similarly, $\calA_{n}$ is the family of all functions $f:(\R^{d}\times\{\pm1\})^{n}\to\R$ such that for any $(y_{i})_{i=1}^{n}\in\{\pm1\}^{n}$, $f((\bz_{i},y_{i})_{i=1}^{n})$ is locally Lipschitz with respect to $(\bz_{i})_{i=1}^{n}\in\R^{dn}$.

For any nonnegative random variable $\xi$, we let $\Ent_{\rho}(\xi)$ denote the entropy of $\xi$ with a probability measure $\rho$ defined as follows:
\begin{equation*}
    \Ent_{\rho}(\xi):=\E_{\rho}[\xi\log\xi]-\E_{\rho}[\xi]\log\E_{\rho}[\xi].
\end{equation*}

We use subscripts for the expectation $\E$, variance $\Var$, and entropy $\Ent$ to indicate the underlying measure. In particular, when these measures are equipped with subscripts (e.g., $P_{\bX}$), we adopt the same subscripts for the corresponding expectation, variance, and entropy to specify the measure they are taken with respect to (e.g., $\E_{\bX}$, $\Var_{\bX}$, and $\Ent_{\bX}$).

We call a symmetric positive bilinear operator $\Gamma$ on $\calA_{1}\times\calA_{1}$ (resp.~$\calA_{n}\times\calA_{n}$) such that (i) $\Gamma(f_{1},f_{2}):\R^{d}\times\{\pm1\}\to\R$ for $f_{1},f_{2}\in\calA_{1}$ (res.~$\Gamma(f_{1},f_{2}):(\R^{d}\times\{\pm1\})^{n}\to\R$ for $f_{1},f_{2}\in\calA_{n}$) is measurable, (ii) $\Gamma(f_{1},f_{1})\ge0$, (iii) $\Gamma(f_{1},f_{2})=\Gamma(f_{2},f_{1})$, and (iv) $\Gamma(1,1)=0$ as a \emph{carr\'{e} du champ operator}.
A typical example is $\Gamma(f_{1},f_{2})=\langle \nabla_{\bz} f_{1},\nabla_{\bz} f_{2}\rangle+\langle \nabla_{y} f_{1},\nabla_{y} f_{2}\rangle$ for $f_{1},f_{2}\in\calA_{1}$.
For notational convenience, we write $\Gamma(f)=\Gamma(f,f)$.

We provide the formal definitions of the evenness of functions and $m$-convexity.
\begin{definition}
    $f:\R^{d}\to\R$ is an \emph{even} function if $f(\bx)=f(-\bx)$ for all $\bx\in\R^{d}$. 
\end{definition}

\begin{definition}
    A convex function $f:\R^{d}\to\R$ is \emph{$m$-convex} with $m\ge0$ if for any $\bx,\by\in\R^{d}$ and any $\bg\in\R^{d}$ being a subgradient of $f$ at $\bx\in\R^{d}$,
    \begin{equation*}
        f(\by)-f(\bx)-\langle \bg,\by-\bx\rangle\ge \frac{m}{2}\left\|\by-\bx\right\|^{2}.
    \end{equation*}
\end{definition}

All convex functions are $0$-convex by definition.

\section{Fundamental results for functional inequalities and tail bounds} \label{sec:functional}
In this section, we discuss functional inequalities for the joint distribution of general random variables $(\bZ,Y)$ on $\R^{d}\times\{\pm1\}$ and tail bounds as their application.
We suppose that $(\bZ,Y)$ are absolutely continuous with respect to the product measure of the $d$-dimensional Lebesgue measure $\Leb$ and the counting one $\sharp$.

\subsection{Settings}
We introduce the notation of the joint, marginal, and conditional measures of $Y$ and $\bZ$.
We let $P_{\bZ Y}(\diff\bz,\diff y)$ denote the joint measure of $(\bZ,Y)$ on $\R^{d}\times\{\pm1\}$.

We use the following notation for marginal/conditional measures and densities: 
\begin{align*}
    P_{\bZ}(\diff\bz)&:=P_{\bZ Y}(\diff\bz,\{\pm1\}), & 
    P_{Y}(\diff y)&:=P_{\bZ Y}(\R^{d}, \diff y)=:p_{Y}(y)\sharp(\diff y),\\
    P_{\bZ|Y}(\diff \bz|y)&:=\frac{1}{p_{Y}(y)}P_{\bZ Y}(\diff\bz,\{y\}),& 
    P_{Y|\bZ}(\diff y|\bz)&:=\frac{P_{\bZ Y}(\diff \bz, \diff y)}{P_{\bZ}(\diff \bz)}=:p_{Y|\bZ}(y|\bz)\sharp(\diff y).
\end{align*}

We also set the following carr\'{e} du champ operators on $\calA_{1}\times\calA_{1}$:
\begin{align*}
    \Gamma_{\bZ}(f_{1},f_{2})(\bz,y):=\left\langle\nabla_{\bz}f_{1}(\bz,y),\nabla_{\bz}f_{2}(\bz,y)\right\rangle,\ 
    \Gamma_{Y}(f_{1},f_{2})(\bz,y):=\nabla_{y}f_{1}(\bz,y)\times\nabla_{y}f_{2}(\bz,y).
\end{align*}
We define the following constants:
\begin{align*}
    K_{\Poincare}:=\max_{y\in\{\pm1\}}C_{\Poincare}(P_{\bZ|Y}(\cdot|y)),
    K_{\LogSobolev}=\max_{y\in\{\pm1\}}C_{\LogSobolev}(P_{\bZ|Y}(\cdot|y)),
\end{align*}
where $C_{\Poincare}(P)\in[0,\infty]$ and $C_{\LogSobolev}(P)\in[0,\infty]$ (they are permitted to be $\infty$, although they render the following functional inequalities vacuous) with $P$, an arbitrary probability measure on $\R^{d}$, are defined as the minimal constants satisfying a Poincar\'{e} inequality and log-Sobolev inequality such that for any locally Lipschitz $f:\R^{d}\to\R$,
\begin{equation*}
    \Var_{P}(f)\le C_{\Poincare}(P)\E_{P}\left[\left\|\nabla f\right\|^{2}\right],\ \Ent_{P}(f)\le 2C_{\LogSobolev}(P)\E_{P}\left[\left\|\nabla f\right\|^{2}\right].
\end{equation*}
We also set the following notation:
\begin{equation*}
    K_{\chi^{2}}:=\max_{y\in\{\pm1\}}\chi^{2}\left(P_{\bZ}(\cdot)\|P_{\bZ|Y}(\cdot|y)\right),\ 
    K_{\textnormal{V}}:=4p_{Y}(+1)p_{Y}(-1),\ 
    K_{\textnormal{U}}:=\max\left\{\frac{1}{p_{Y}(+1)},\frac{1}{p_{Y}(-1)}\right\}.
\end{equation*}
$K_{\chi^{2}}$ quantifies the dependence between $\bZ$ and $Y$.
If $\bZ\perp Y$, then $K_{\chi^{2}}=0$.
$K_{\textnormal{V}}$ is the variance of $Y$.
The constant $K_{\textnormal{U}}$ represents the degree of deviation from the uniform distribution on $\{\pm1\}$.
As well as $K_{\Poincare}$ and $K_{\LogSobolev}$, if these constants are not finite, the derived functional inequalities below are vacuous.

\begin{remark}
    One can consider examples where $\chi^{2}(P_{\bZ}(\cdot)\|P_{\bZ|Y}(\cdot|y))=\infty$ and $K_{\textnormal{U}}=\infty$, which render the functional inequalities in Theorem \ref{thm:functional} below vacuous.
    For instance, if $\Pr(Y\langle \bZ,\bw\rangle > 0)=\Pr(\langle \bX,\bw\rangle>0)=1$ for some $\bw\neq0$, then the conditional measures $P_{\bZ|Y}(\cdot|+1)$ and $P_{\bZ|Y}(\cdot|-1)$ are mutually singular.
\end{remark}

\subsection{Functional inequalities}
We present Poincar\'{e} and log-Sobolev inequalities for weighted sums of the carr\'{e} du champ operators $\Gamma_{\bZ}$ and $\Gamma_{Y}$.
Although the inequalities formally appear with constant 1, the associated carré du champ operators may carry large scaling factors.
We adopt the convention that $\infty \times 0=0$.

\begin{theorem}\label{thm:functional}
    The following Poincar\'{e} and log-Sobolev inequalities hold: for any $f\in\calA_{1}$, 
    \begin{align*}
        \Var(f(\bZ,Y))&\le \E\left[\Gamma_{\Poincare}(f)(\bZ,Y)\right],\\
        \Ent(f^{2}(\bZ,Y))&\le 2\E\left[\Gamma_{\LogSobolev}(f)(\bZ,Y)\right],
    \end{align*}
    where $\Gamma_{\Poincare}$ and $\Gamma_{\LogSobolev}$ are the following weighted carr\'{e} du champ operators for any conjugate pair $c,c^{\ast}\in[1,\infty]$ such that $1/c+1/c^{\ast}=1$: for any $f_{1},f_{2}\in\calA_{1}$,
    \begin{align*}
        \Gamma_{\Poincare}(f_{1},f_{2})&:=K_{\Poincare}\left(1+cK_{\chi^{2}}\right)\Gamma_{\bZ}(f_{1},f_{2})+c^{\ast}K_{\textnormal{V}}\Gamma_{Y}(f_{1},f_{2}),\\
        \Gamma_{\LogSobolev}(f_{1},f_{2})&:=\left(1+\frac{1}{2}\log K_{\textnormal{U}}\right)\Gamma_{\Poincare}(f_{1},f_{2})+2K_{\LogSobolev}\Gamma_{\bZ}(f_{1},f_{2}).
    \end{align*}
\end{theorem}

If $\bZ\perp Y$ and thus $K_{\chi^{2}}=0$, then we can select $c=\infty$ and $c^{\ast}=1$ and obtain
\begin{equation*}
    \Gamma_{\Poincare}(f_{1},f_{2})=C_{\Poincare}(P_{\bZ})\Gamma_{\bZ}(f_{1},f_{2})+\Gamma_{Y}(f_{1},f_{2}),
\end{equation*}
which exactly recovers the result by tensorization.
$\Gamma_{\text{LS}}$ can also be recovered up to a constant $1+(1/2)\log K_{\textnormal{U}}$, which should be multiplied with $\Gamma_{Y}$ due to the well-known logarithmic Sobolev constant for binomial distributions \citep[see][]{boucheron2013concentration}; the multiplicative constant of $\Gamma_{\bZ}$ in the logarithmic Sobolev inequality above is optimal up to a numerical constant and $\log K_{\textnormal{U}}$.

As we consider a sequence of $\R^{d}\times\{\pm1\}$-valued i.i.d.~random variables, the following tensorization is an important tool for deriving concentration bounds.
\begin{lemma}[Propositions 4.3.1 and 5.2.7 of \citealp{bakry2014analysis}; see also Theorems 2.3 and 3.14 of \citealp{van2014probability}]\label{lem:tensor}
    Let $P_{i}$ ($i=1,\ldots,n$) be a probability measure on a state space $\R^{d}\times\{\pm1\}$ and suppose that every $P_{i}$ satisfies a Poincar\'{e} inequality (resp.~log-Sobolev inequality) for constant $C>0$ with a carr\'{e} du champ operator $\Gamma_{i}$.
    Then, $\bigotimes_{i=1}^{n}P_{i}$ on $(\R^{d}\times\{\pm1\})^{n}$ satisfies a Poincar\'{e} inequality (resp.~log-Sobolev inequality) for constant $C$ with the  carr\'{e} du champ operator $\Gamma=\bigoplus_{i=1}^{n}\Gamma_{i}$.
\end{lemma}

Using tensorization, we obtain the following result as a corollary of Theorem \ref{thm:functional}.
\begin{corollary}
    Let $((\bZ_i,Y_i))_{i=1}^{n}$ be i.i.d.~random variables that are independent copies of $(\bZ,Y)$. Then, the following Poincar\'{e} and log-Sobolev inequalities hold: for any $f\in\calA_{n}$, 
    \begin{align*}
        \Var(f((\bZ_{i},Y_{i})_{i=1}^{n}))&\le \E\left[\Gamma_{\Poincare,n}(f)((\bZ_{i},Y_{i})_{i=1}^{n})\right],\\
        \Ent(f^{2}((\bZ_{i},Y_{i})_{i=1}^{n}))&\le 2\E\left[\Gamma_{\LogSobolev,n}(f)((\bZ_{i},Y_{i})_{i=1}^{n})\right],
    \end{align*}
    where, for any conjugate pair $c,c^{\ast}\in[1,\infty]$ such that $1/c+1/c^{\ast}=1$ and any $f_{1},f_{2}\in\calA_{n}$, the weighted carr\'{e} du champ operators $\Gamma_{\Poincare,n}$ and $\Gamma_{\LogSobolev,n}$ are defined as
    \begin{align*}
        \Gamma_{\Poincare,n}(f_{1},f_{2})&:=K_{\Poincare}\left(1+cK_{\chi^{2}}\right)\sum_{i=1}^{n}\Gamma_{\bZ_i}(f_{1},f_{2})+c^{\ast}K_{\textnormal{V}}\sum_{i=1}^{n}\Gamma_{Y_{i}}(f_{1},f_{2}),\\
        \Gamma_{\LogSobolev,n}(f_{1},f_{2})&:=\left(1+\frac{1}{2}\log K_{\textnormal{U}}\right)\Gamma_{\Poincare,n}(f_{1},f_{2})+2K_{\LogSobolev}\sum_{i=1}^{n}\Gamma_{\bZ_i}(f_{1},f_{2}).
    \end{align*}
    Here, $\Gamma_{\bZ_i}(f_{1},f_{2})=\langle \nabla_{\bz_{i}}f_{1},\nabla_{\bz_{i}}f_{2}\rangle$ and $\Gamma_{Y_i}(f_{1},f_{2})=\nabla_{y_{i}}f_{1}\times\nabla_{y_{i}}f_{2}$.
\end{corollary}

\subsection{Tail bounds}\label{sec:functional:tail}
To derive subexponential and subgaussian tail bounds via the functional inequalities, we need extensions of well-known results under Euclidean settings.
It is because the chain rule for discrete gradients holds not exactly but just approximately.

We derive the following subexponential and subgaussian tail bounds; their multiplicative numerical constants are quite mild in comparison to the case where the discrete space is absent.
\begin{lemma}[subexponential concentration; see also \citealp{bobkov1997poincare,bakry2014analysis}]\label{lem:bobkov}
    Let $P$ be a probability measure on a state space $(\R^{d}\times\{\pm1\})^{n}$ absolutely continuous with respect to $\bigotimes_{i=1}^{n}(\Leb\times\sharp)$ such that the following Poincar\'{e} inequality holds true: for some positive constants $C_{\bZ},C_{Y}>0$, for any $f\in\calA_{n}$,
    \begin{equation*}
        \Var_{P}(f)\le \E_{P}\left[C_{\bZ}\sum_{i=1}^{n}\Gamma_{\bZ_{i}}(f)+C_{Y}\sum_{i=1}^{n}\Gamma_{Y_{i}}(f)\right],
    \end{equation*}
    where  $\Gamma_{\bZ_i}(f_{1},f_{2})=\langle \nabla_{\bz_{i}}f_{1},\nabla_{\bz_{i}}f_{2}\rangle$ and $\Gamma_{Y_i}(f_{1},f_{2})=\nabla_{y_{i}}f_{1}\times\nabla_{y_{i}}f_{2}$.
    Suppose that a function $f\in\calA_{n}$ satisfies $\|\Gamma_{\bZ_{i}}(f)\|_{L^{\infty}}<\infty$ and $\|\Gamma_{Y_{i}}(f)\|_{L^{\infty}}<\infty$.
    Then, letting $L_{\bZ}=\sqrt{\|\Gamma_{\bZ_{i}}(f)\|_{L^{\infty}}}$ and $L_{Y}=\sqrt{\|\Gamma_{Y_{i}}(f)\|_{L^{\infty}}}$, we have that for all $t\ge0$,
    \begin{equation*}
        P\left(f-\E_{P}[f]\ge t\right)\le 2\exp\left(-\frac{t}{\max\left\{\sqrt{n(C_{\bZ}L_{\bZ}^{2}+2C_{Y} L_{Y}^{2})},2L_{Y}\right\}}\right).
    \end{equation*}
    In particular, for any $\delta\in(0,1]$,
    \begin{equation*}
        P\left(f\le \E_{P}[f]+\max\left\{\sqrt{n(C_{\bZ}L_{\bZ}^{2}+2C_{Y} L_{Y}^{2})},2L_{Y}\right\}\log(2\delta^{-1})\right)\ge 1-\delta.
    \end{equation*}
\end{lemma}

\begin{lemma}[subgaussian concentration; see also \citealp{ledoux1999concentration,bakry2014analysis}]\label{lem:herbst}
    Let $P$ be a probability measure on a state space $(\R^{d}\times\{\pm1\})^{n}$  absolutely continuous with respect to $\bigotimes_{i=1}^{n}(\Leb\times\sharp)$ such that the following log-Sobolev inequality holds true: for some positive constants $C_{\bZ},C_{Y}>0$, for any $f\in\calA_{n}$,
    \begin{equation*}
        \Ent_{P}(f^{2})\le 2\E_{P}\left[C_{\bZ}\sum_{i=1}^{n}\Gamma_{\bZ_{i}}(f)+C_{Y}\sum_{i=1}^{n}\Gamma_{Y_{i}}(f)\right],
    \end{equation*}
    where $\Gamma_{\bZ_i}(f_{1},f_{2})=\langle \nabla_{\bz_{i}}f_{1},\nabla_{\bz_{i}}f_{2}\rangle$ and $\Gamma_{Y_i}(f_{1},f_{2})=\nabla_{y_{i}}f_{1}\times\nabla_{y_{i}}f_{2}$.
    Suppose that a function $f\in\calA_{n}$ satisfies $\|\Gamma_{\bZ_{i}}(f)\|_{L^{\infty}}<\infty$ and $\|\Gamma_{Y_{i}}(f)\|_{L^{\infty}}<\infty$.
    Then, letting $L_{\bZ}=\sqrt{\|\Gamma_{\bZ_{i}}(f)\|_{L^{\infty}}}$ and $L_{Y}=\sqrt{\|\Gamma_{Y_{i}}(f)\|_{L^{\infty}}}$, we have that for all $t\ge0$,
    \begin{equation*}
        P\left(f-\E_{P}[f]\ge t\right)\le \exp\left(-\min\left\{\frac{t^{2}}{n(2C_{\bZ}L_{\bZ}^2+eC_{Y}L_{Y}^2)},\frac{t}{4L_{Y}}\right\}\right).
    \end{equation*}
    In particular, for any $\delta\in(0,1]$,
    \begin{equation*}
        P\left(f\le \E_{P}[f]+\max\left\{\sqrt{n(2C_{\bZ}L_{\bZ}^2+eC_{Y}L_{Y}^2)\log(\delta^{-1})},4L_{Y}\log(\delta^{-1})\right\}\right)\ge 1-\delta.
    \end{equation*}
\end{lemma}
Using these extensions, we can derive concentration inequalities for $(\R^{d}\times\{\pm1\})$-valued random variables similar to ones for $\R^{d}$-valued random variables.

\section{Concentration inequalities for uniform generalization errors}
\label{sec:concentration}

We now apply Theorem \ref{thm:functional} to concentration bounds on uniform generalization errors for binary linear classification problems.

We first introduce a basic estimate for the expectation of uniform generalization errors, which should be the main term of the bound \eqref{eq:ConcAroundExp}.
\begin{proposition}\label{prop:rademacher}
    We have
    \begin{equation*}
        \E\left[\sup_{(\bw,b)\in\calT}(\Risk(\bw,b)-\Risk_{n}(\bw,b))\right]\le \frac{2L\left(R_{\bw}\sqrt{\tr(\E[\bX_{i}\bX_{i}^{\top}])}+R_{b}\right)}{\sqrt{n}}.
    \end{equation*}
\end{proposition}
Note that this result is dimension-independent, but depends on a possibly large quantity $\tr(\E[\bX_{i}\bX_{i}^{\top}])$.
The purpose of this section is to show that the residual term in Equation \eqref{eq:ConcAroundExp} is much smaller than this standard estimate for expected uniform generalization errors.

\subsection{Basic settings}\label{sec:concentration:settings}
We set the probability measure of $\bX_{i}$ as
\begin{equation*}
    P_{\bX}(\diff \bx)=\calZ^{-1}\exp\left(-U\left(\bx\right)\right)\diff \bx,
\end{equation*}
where $U:\R^{d}\to[0,\infty)$ is the potential function, and $\calZ=\int_{\R^{d}}\exp\left(-U\left(\bx\right)\right)\diff \bx$ is the normalizing constant assumed to be finite.
Although theorems below assume the evenness of $U$, we can relax them to the settings with non-central input vectors; see Section \ref{sec:concentration:noncentral} for details.
We also set the conditional probability mass function of $Y_{i}$ on $\{\pm1\}$ given $\bX_{i}=\bx_{i}\in\R^{d}$ as
\begin{equation*}
    p_{Y|\bX}(y|\bx)=g(y(\langle\bx,\btheta_{1}\rangle+\theta_{0})),
\end{equation*}
where $g:\R\to[0,1]$ is a function called a link function, $\btheta_{1}\in\R^{d}$ is a fixed vector, and $\theta_{0}\in\R$ is a fixed number.
By definition, we have $g(t)+g(-t)=1$ for all $t\in\R$.

Under these settings, we derive the following joint probability measure of $(\bZ_{i},Y_{i})$:
\begin{equation*}
    P_{\bZ Y}(\diff\bz,\diff y)=\calZ^{-1}g(\langle\bz,\btheta_{1}\rangle+y\theta_{0})\exp(-U(y\bz))\diff\bz\sharp(\diff y).
\end{equation*}
The marginal density/mass functions and conditional density/mass functions are given as
\begin{align*}
    p_{\bZ}(\bz)&:=\frac{P_{\bZ}(\diff\bz)}{\diff\bz}=\sum_{y\in\{\pm1\}}\calZ^{-1}g(\langle\bz,\btheta_{1}\rangle+y\theta_{0})\exp(-U(y\bz)),\\
    p_{Y}(y)&:=\frac{P_{y}(\diff y)}{\sharp(\diff y)}=\frac{1}{\calZ}\int_{\R^{d}} g(\langle\bz,\btheta_{1}\rangle+y\theta_{0})\exp(-U(y\bz))\diff\bz=:\frac{\calZ_{y}}{\calZ},\\
    p_{\bZ|Y}(\bz|y)&=\calZ_{y}^{-1}g(\langle\bz,\btheta_{1}\rangle+y\theta_{0})\exp(-U(y\bz)),\\
    p_{Y|\bZ}(y|\bz)&=\frac{g(\langle\bz,\btheta_{1}\rangle+y\theta_{0})\exp(-U(y\bz))}{\sum_{y'\in\{\pm1\}}g(\langle\bz,\btheta_{1}\rangle+y'\theta_{0})\exp(-U(y'\bz))}.
\end{align*}
Here, we employ the fact $\calZ=\int_{\R^{d}}\exp(-U(\bx))\diff\bx=\sum_{y\in\{\pm 1\}} \int_{\R^{d}}g(y(\langle\bx,\btheta_{1}\rangle+\theta_{0}))\exp(-U(\bx))\diff \bx=\sum_{y\in\{\pm 1\}} \int_{\R^{d}}g(\langle\bz,\btheta_{1}\rangle+y\theta_{0})\exp(-U(y\bz))\diff \bz$.
Our loss function $\ell(\cdot)$ does not necessarily coincide with $-\log(g(\cdot))$.

We set the assumptions on functional inequalities of the conditional distributions $P_{\bZ|Y}(\cdot|y)$ for $y\in\{\pm1\}$.

\begin{assumption}[Poincar\'{e} inequality]\label{assum:pi}
    For both $y\in\{\pm1\}$, the conditional distribution $P_{\bZ|Y}(\cdot|y)$ satisfies a Poincar\'{e} inequality (with respect to $\Gamma_{\bZ}$) for some constant $K_{\Poincare}>0$.
\end{assumption}

\begin{assumption}[log-Sobolev inequality]\label{assum:lsi}
    For both $y\in\{\pm1\}$, the conditional distribution $P_{\bZ|Y}(\cdot|y)$ satisfies a log-Sobolev inequality  (with respect to $\Gamma_{\bZ}$) for some constant $K_{\LogSobolev}>0$.
\end{assumption}

Note that a Poincar\'{e} inequality yields subexponential concentration of arbitrary Lipschitz maps, while a log-Sobolev inequality yields subgaussian concentration of arbitrary Lipschitz maps.
Since $K_{\Poincare}\le K_{\LogSobolev}$ in general, sufficient conditions for a log-Sobolev inequality are stronger than those for a Poincar\'{e} inequality.

Let us introduce some natural sufficient conditions for Assumption \ref{assum:lsi}.
For example, if $U$ is $K^{-1}$-convex, $\log\circ\ g$ is concave, and they are of $\calC^{2}$, then the potential of the conditional distribution is $K^{-1}$-convex, and the conditional distribution satisfies a log-Sobolev inequality with constant $K$ by the Bakry--\'{E}mery theory.
Note that $U(\bx)=(1/2)\langle\bx,\bSigma^{-1}\bx\rangle $, the potential function of $\Gauss_{d}(\zero_{d},\bSigma)$ for positive definite $\bSigma$, is $\|\bSigma\|_{\op}$-convex.
Moreover, if $U$ is even, $\log\circ\ g$ is $G$-Lipschitz, and a log-Sobolev inequality for the probability measure $(\int_{\R^{d}}g(\langle\bz,\btheta_{1}\rangle )e^{-U(\bz)}\diff\bz)^{-1}g(\langle\bz,\btheta_{1}\rangle )e^{-U(\bz)}\diff\bz$ holds with constant $Ke^{-2G|\theta_{0}|}$, then the perturbation theory \citep{bakry2014analysis,cattiaux2022functional} gives Assumption \ref{assum:lsi}.
It is because $g(t+|\theta_{0}|)/g(t)\le e^{G|\theta_{0}|}$ and $g(t-|\theta_{0}|)/g(t)\ge e^{-G|\theta_{0}|}$ for all $t\in\R$.
For example, the $1$-Lipschitz continuity of $\log\circ\ \sigma$ holds, where $\sigma(t):=1/(1+\exp(-t))$ is the logistic link function.

We also introduce sufficient conditions for Assumption \ref{assum:pi}.
Since $K_{\Poincare}\le K_{\LogSobolev}$, sufficient conditions for Assumption \ref{assum:lsi} are also sufficient for Assumption \ref{assum:pi}.
A log-Sobolev inequality in Theorem \ref{thm:functional} includes the multiplicative constant $(1+(1/2)\log K_{U})$, which can be large in some situations (Theorems \ref{prop:concentration:2} and \ref{prop:concentration:3}); hence, considering subexponential concentration is sometimes beneficial even if we know that Assumption \ref{assum:lsi} holds.
In addition, if both $U$ and $(-1)\log\circ\ g$ are convex (here, we do not require the $m$-convexity of $U$ for some positive $m>0$), we can give estimates for the Poincar\'{e} constants by applying recent results on the Kannan--Lov\'{a}sz--Simonovits (KLS) conjecture \citep{kannan1995isoperimetric}, which is a topical problem  \citep{lee2018kannan,chen2021almost,klartag2022bourgain,jambulapati2022slightly,klartag2023logarithmic,lee2024eldan}.
Let us introduce a constant $C_{\KLS}(d)=\sup_{\mu:\text{isotropic and log-concave on }\R^{d}}C_{\Poincare}(\mu)$, where $C_{\Poincare}(\mu)$ is the minimal constant in a Poincar\'{e} inequality of $\mu$ such that for any locally Lipschitz $f:\R^{d}\to\R$, $\Var_{\mu}(f)\le C_{\Poincare}(\mu)\E_{\mu}[\|\nabla f\|^{2}]$.
We notice that $C_{\Poincare}(P_{\bZ_{i}|Y}(\cdot|y))\le \|\E[\bZ_{i}\bZ_{i}^{\top}]-\E[\bZ_{i}]\E[\bZ_{i}]^{\top}\|_{\text{op}}C_{\KLS}(d)\le \|\E[\bX_{i}\bX_{i}^{\top}]\|_{\text{op}}C_{\KLS}(d)$.
The statement of the KLS conjecture is ``$C_{\KLS}:=\sup_{d\in\N}C_{\KLS}(d)<\infty$''; to our best knowledge, the most recent estimate is $C_{\KLS}(d)\le c(1+\log (d))$ for some absolute constant $c>0$ by \citet{klartag2023logarithmic}.
Hence, we can let $K_{\Poincare}=c(1+\log (d))\|\E[\bX_{i}\bX_{i}^{\top}]\|_{\textnormal{op}}$ under the (weak) convexity of $U$ and $(-1)\log\circ\ g$.
Moreover, we can apply an upper bound given by \citet{lee2024eldan} in the place of $\|\E[\bX_{i}\bX_{i}^{\top}]\|_{\textnormal{op}}C_{\KLS}(d)$; they yield $C_{\Poincare}(P_{\bZ|Y}(\cdot|y))\le c'\tr(\E[\bX_{i}\bX_{i}^{\top}]^{2})^{1/2}$ for an absolute constant $c'>0$.
Hence, we can let $K_{\Poincare}=c'\tr(\E[\bX_{i}\bX_{i}^{\top}]^{2})^{1/2}$.
The residual derived by this result is smaller than that by McDiarmid's inequality and the Rademacher complexity bound for the expected uniform generalization error given by Proposition \ref{prop:rademacher} up to $\log(3/\delta)$ and a numerical constant.

\subsection{Concentration under no bias via tensorization}\label{sec:concentration:base}
First, we consider the case under $\theta_{0}=0$ as the baseline analysis; to our best knowledge, this result itself is novel as dimension-free concentration inequalities for uniform generalization errors in binary linear classification problems without boundedness. 
The setting $\theta_{0}=0$ results in the independence of $\bZ_{i}$ and $Y_{i}$.

\begin{proposition}\label{prop:independence}
    Assume that the potential function $U$ is even and $\theta_{0}=0$. Then, $\bZ_{i}$ and $Y_{i}$ are independent.
\end{proposition}

Using Proposition \ref{prop:independence}, we derive the following concentration inequality via the classical tensorization argument without our Theorem \ref{thm:functional}.
Hence, we take this result as a baseline for comparison, and we will give concentration inequalities similar to this inequality in more general settings.

\begin{proposition}[concentration under no bias]\label{prop:concentration:0}
    Suppose that the same assumption as Proposition \ref{prop:independence} holds true.
    
    (i) Under Assumption \ref{assum:pi}, for all $\delta\in(0,1]$, with probability at least $1-\delta$,
    \begin{align*}
        &\sup_{(\bw,b)\in\calT}(\Risk(\bw,b)-\Risk_{n}(\bw,b))-\E\left[\sup_{(\bw,b)\in\calT}(\Risk(\bw,b)-\Risk_{n}(\bw,b))\right]\\
        &\le \max\left\{\sqrt{\frac{L^{2}(K_{\Poincare}R_{\bw}^{2}+2R_{b}^{2})}{n}},\frac{2LR_{b}}{n}\right\}\log(2\delta^{-1}).
    \end{align*}
    (ii) Under Assumption \ref{assum:lsi}, for all $\delta\in(0,1]$, with probability at least $1-\delta$,
    \begin{align*}
        &\sup_{(\bw,b)\in\calT}(\Risk(\bw,b)-\Risk_{n}(\bw,b)) - \E\left[\sup_{(\bw,b)\in\calT}(\Risk(\bw,b)-\Risk_{n}(\bw,b))\right]\\
        &\le \max\left\{\sqrt{\frac{L^{2}(2K_{\LogSobolev}R_{\bw}^{2}+eR_{b}^{2})\log(\delta^{-1})}{n}},\frac{4LR_{b}\log(\delta^{-1})}{n}\right\}.
    \end{align*}
\end{proposition}
The bound (ii) is sharp in the sense that it recovers the tail behaviour of the standard Gaussian distribution. 
That is, we can choose $R_{b}=0$ and $K_{\LogSobolev}=1$ and observe that it matches the standard Gaussian concentration.

\begin{remark}
    The typical examples of log-concave $g$ is the logistic link function $\sigma$ and probit link function $\Phi$ defined as for all $t\in\R$,
    \begin{equation}\label{eq:LogitAndProbit}
        \sigma(t):=\frac{1}{1+\exp\left(-t\right)},\ 
        \Phi(t):=\int_{-\infty}^{t}\frac{\exp(-s^{2}/2)}{\sqrt{2\pi}}\diff s.
    \end{equation}
\end{remark}

\subsection{Concentration in general settings via Theorem \ref{thm:functional}}\label{sec:concentration:main}
Proposition \ref{prop:concentration:0} yields clean and rate-optimal bounds owing to tensorization, but it needs the assumption $\theta_{0}=0$.
We use our Theorem \ref{thm:functional} to investigate the concentration of uniform generalization errors in more general settings.
We exhibit four concentration inequalities, which are appealing under the following four edge cases respectively (see Figure \ref{fig:edge}): weak bias $|\theta_{0}|=\calO(1)$; strong bias $|\theta_{0}|\gg1$; weak signal $\|\btheta_{1}\|=\calO(1)$; and strong signal $\|\btheta_{1}\|\gg1$. 
In particular, the cases $|\theta_{0}|=\calO(1)$ (Proposition \ref{prop:concentration:1}) and $\|\btheta_{1}\|\gg1$ (Proposition \ref{prop:concentration:4}) yield upper bounds matching that of Proposition \ref{prop:concentration:0} up to mild multiplicative constants.
\begin{figure}[ht]
    \centering
\begin{tikzpicture}[scale=1.0, every node/.style={font=\small}]
    \coordinate (O) at (0,0);
    \coordinate (A) at (4,0);
    \coordinate (B) at (4,4);
    \coordinate (C) at (0,4);
    \coordinate (M) at (2,2);

    \fill[green!50!black!60] (O) -- (A) -- (M) -- cycle;
    
    \fill[orange!80!black!70] (A) -- (B) -- (M) -- cycle;
    
    \fill[blue!60!cyan!80] (O) -- (M) -- (C) -- cycle;
    
    \fill[magenta!70!black!60] (C) -- (M) -- (B) -- cycle;

    \draw[thick, white, line width=1.5pt] (O) -- (B);
    \draw[thick, white, line width=1.5pt] (C) -- (A);

    \draw[line width=6pt, orange, dashed] (O) -- (C);

    \node[text=white, font=\bfseries] at (1.0, 2.0) {Prop. 6};
    \node[text=white, font=\bfseries] at (3.0, 2.0) {Prop. 7};
    \node[text=white, font=\bfseries] at (2.0, 0.8) {Prop. 8};
    \node[text=white, font=\bfseries] at (2.0, 3.2) {Prop. 9};
    
    \node[orange!80!black, left, font=\bfseries, align=right] (P5) at (-0.4, 2.0) {Prop. 5\\($|\theta_0|=0$)};
    \draw[->, line width=0.8pt, orange!80!black] (P5.east) -- (-0.05, 2.0);
    
    \draw[->, line width=0.8pt] (O) -- (4.5, 0) node[right] {$|\theta_{0}|$};
    \draw[->, line width=0.8pt] (O) -- (0, 4.5) node[above] {$\|\theta_{1}\|$};
    \node[below left] at (O) {$O$};
\end{tikzpicture}
    \caption{A schematic representation of the regimes covered in Propositions \ref{prop:concentration:1}--\ref{prop:concentration:4}.
    The orange dotted line represents the zero bias region (studied by Proposition \ref{prop:concentration:0}). 
    Situations in the region close to the left edge (that is, small $|\theta_{0}|$) can be evaluated by Proposition \ref{prop:concentration:1}; the same applies to other cases.}
    \label{fig:edge}
\end{figure}

\subsubsection{Concentration under weak bias}
In the next place, we examine the case where $|\theta_{0}|$ is non-zero but not very large.
If $\theta_{0}\neq0$, then $\bZ_{i}$ and $Y_{i}$ are dependent in general.
Therefore, we apply Theorem \ref{thm:functional} instead of tensorization and derive the following result.

\begin{proposition}[concentration under weak bias]\label{prop:concentration:1}
    Assume that the potential function $U$ is even, $g$ is non-decreasing, and $\log\circ\ g$ is $G$-Lipschitz.
    
    (i) Suppose that Assumption \ref{assum:pi} holds true.
    Then, for all $\delta\in(0,1]$, with probability at least $1-\delta$,
    \begin{align*}
        &\sup_{(\bw,b)\in\calT}(\Risk(\bw,b)-\Risk_{n}(\bw,b))-\E\left[\sup_{(\bw,b)\in\calT}(\Risk(\bw,b)-\Risk_{n}(\bw,b))\right]\\
        &\le\max\left\{\sqrt{\frac{L^{2}\left[K_{\Poincare}\left(e^{2G|\theta_{0}|}+\sqrt{e^{2G|\theta_{0}|}-1}\right)R_{\bw}^{2}+2(1+\sqrt{e^{2G|\theta_{0}|}-1})R_{b}^{2}\right]}{n}},\frac{2LR_{b}}{n}\right\}\log(2\delta^{-1}).
    \end{align*}
    
    (ii) Suppose that Assumption \ref{assum:lsi} holds true.
    Then, for all $\delta\in(0,1]$, with probability at least $1-\delta$,
    \begin{align*}
        &\sup_{(\bw,b)\in\calT}(\Risk(\bw,b)-\Risk_{n}(\bw,b))-\E\left[\sup_{(\bw,b)\in\calT}(\Risk(\bw,b)-\Risk_{n}(\bw,b))\right]\\
        &\le\max\left\{\sqrt{\frac{L^{2}\left(3+2G|\theta_{0}|\right)\left[K_{\LogSobolev}\left(1+2e^{2G|\theta_{0}|}\right)R_{\bw}^{2}+eR_{b}^{2}\right]\log(\delta^{-1})}{n}},\frac{4LR_{b}\log(\delta^{-1})}{n}\right\}.
    \end{align*}
\end{proposition}

\begin{remark}
    It is easy to see that $\log\circ\ \sigma$, the composition of the logarithmic function and the logistic link function \eqref{eq:LogitAndProbit}, is $1$-Lipschitz.
    On the other hand, $\log\circ\ \Phi$, the composition of the logarithmic function and the probit link function \eqref{eq:LogitAndProbit}, is not Lipschitz.
    Hence, we can assume logistic models as the true model, but cannot set probit models to apply Proposition \ref{prop:concentration:1} under $\theta_{0}\neq0$.
\end{remark}

\subsubsection{Concentration under strong bias}\label{sec:concentration:largebias}
If $|\theta_{0}|$ is large, then the upper bound given by Proposition \ref{prop:concentration:1} can be vacuous; however, we conjecture that $Y_{i}$ would be almost constant when $|\theta_{0}|$ is sufficiently large, and thus $\bZ_{i}$ and $Y_{i}$ should again be approximately independent (we examine this conjecture numerically in Section \ref{sec:discussion:further}).
In this section, we restrict ourselves to discussing the logistic link function $g=\sigma$ and see that the guess is certainly correct in a specified setting.

We define the following moment-generating function:
\begin{equation*}
    M_{\bX}(\bt):=\E\left[\exp\left(\langle\bX,\bt\rangle\right)\right],\ \bt\in\R^{d}.
\end{equation*}
If $U$ is even, then $M_{\bX}$ is also an even function.

\begin{proposition}[concentration under strong bias and logistic models]\label{prop:concentration:2}
    Assume that the potential function $U$ is even and $g(t)=\sigma(t)=1/(1+\exp(-t))$.
    
    (i)
    Suppose that Assumption \ref{assum:pi} holds.
    Then, for all $\delta\in(0,1]$, with probability at least $1-\delta$,
    \begin{align*}
        &\sup_{(\bw,b)\in\calT}(\Risk(\bw,b)-\Risk_{n}(\bw,b))
        - \E\left[\sup_{(\bw,b)\in\calT}(\Risk(\bw,b)-\Risk_{n}(\bw,b))\right]\\
        &\le \max\left\{\sqrt{\frac{L^{2}\left[K_{\Poincare}\left(8M_{\bX}^{2}(\btheta_{1})-1\right)R_{\bw}^{2}+16e^{-|\theta_{0}|}M_{\bX}(\btheta_{1})R_{b}^{2}\right]}{n}},\frac{2LR_{b}}{n}\right\}\log(2\delta^{-1}).
    \end{align*}
    
    (ii)
    Suppose that Assumption \ref{assum:lsi} holds.
    Then, for all $\delta\in(0,1]$, with probability at least $1-\delta$,
    \begin{align*}
        &\sup_{(\bw,b)\in\calT}(\Risk(\bw,b)-\Risk_{n}(\bw,b))
        - \E\left[\sup_{(\bw,b)\in\calT}(\Risk(\bw,b)-\Risk_{n}(\bw,b))\right]\\
        &\le\max\left\{\sqrt{\frac{L^{2}\left(e+|\theta_{0}|\right)\left[K_{\LogSobolev}\left(1/2+8M_{\bX}^{2}(\btheta_{1})\right)R_{\bw}^{2}+4e^{1-|\theta_{0}|}M_{\bX}(\btheta_{1})R_{b}^{2}\right]\log(\delta^{-1})}{n}},\frac{4LR_{b}\log(\delta^{-1})}{n}\right\}.
    \end{align*}
\end{proposition}

The right-hand side in Proposition \ref{prop:concentration:2}-(i) is decreasing in $|\theta_{0}|$; therefore, large $|\theta_{0}|$ is beneficial to yield a better bound from this proposition.
Proposition \ref{prop:concentration:2}-(ii) gives a residual with dependence on $\sqrt{|\theta_{0}|}$ (ignoring other parameters), which is much milder than $\exp(G|\theta_{0}|)$ of Proposition \ref{prop:concentration:1}-(ii).
For instance, if $|\theta_{0}|=o(n)$ in statistical experiments, then Proposition \ref{prop:concentration:2}-(ii) concludes that a uniform generalization error concentrates around its expectation well.  
Removing the dependence on $|\theta_{0}|$ via the improvement of Theorem \ref{thm:functional} is an important direction of further investigation.

\subsubsection{Concentration under weak signal}
We can guess that the residual in Equation \eqref{eq:ConcAroundExp} should be close to that of Proposition \ref{prop:concentration:0} if $\|\btheta_{1}\|$ is very small; it is because the dependence between $\bX_{i}$ and $Y_{i}$ or that between $\bZ_{i}$ and $Y_{i}$ should be weak.

Let us define the following functions resembling the moment-generating function of $\bX$:
\begin{equation*}
    M_{\bX}^{\textnormal{num}}(\bt):=\E\left[\exp\left(\max\{0,\langle \bX,\bt\rangle\}\right)\right],\ M_{\bX}^{\textnormal{den}}(\bt):=\E\left[\exp\left(\min\{0,\langle \bX,\bt\rangle\}\right)\right],\ \bt\in\R^{d}.
\end{equation*}
We know that $M_{\bX}^{\textnormal{den}}(\bt)\ge \exp(\E[\min\{0,\langle \bX,\bt\rangle\}])=\exp(-\E[|\langle \bX,\bt\rangle|]/2)$ by Jensen's inequality if $U$ is even.
If $U$ is even, $M_{\bX}^{\textnormal{num}}(\bt)\le \E[\exp(|\langle \bX,\bt\rangle|)]\le \E[\exp(\langle \bX,\bt\rangle)+\exp(\langle -\bX,\bt\rangle)]\le 2M_{\bX}(\bt)$.
Furthermore, we can derive the exact representations of $M_{\bX}^{\textnormal{num}}(\bt)$ and $M_{\bX}^{\textnormal{den}}(\bt)$ if $\bX\sim\calN_{d}(\zero_{d},\bSigma)$.
As $X_{\bt}:=\langle \bX,\bt\rangle\sim \calN_{1}(0,
\sigma_{\bt}^{2})$ with $\sigma_{\bt}^{2}=\langle\bt,\bSigma\bt\rangle$, it holds true that
$\E[e^{\max\{0,X_{\bt}\}}]=(2\pi\sigma_{\bt}^2)^{-1/2}\int_{0}^{\infty}\exp(x-x^{2}/(2\sigma_{\bt}^{2}))\diff x+1/2=\exp(\sigma_{\bt}^{2}/2)(2\pi\sigma_{\bt}^2)^{-1/2}\int_{0}^{\infty}\exp(-(x-\sigma_{\bt}^{2})^{2}/(2\sigma_{\bt}^{2}))\diff x+1/2=\exp(\sigma_{\bt}^{2}/2)(2\pi)^{-1/2}\int_{-\sigma_{\bt}}^{\infty}\exp(-y^2/2)\diff y+1/2=\exp(\sigma_{\bt}^{2}/2)(1-\Phi(-\sigma_{\bt}))+1/2=\exp(\langle \bt,\bSigma\bt\rangle/2)(1-\Phi(-\sqrt{\langle \bt,\bSigma\bt\rangle}))+1/2=\exp(\langle \bt,\bSigma\bt\rangle/2)\Phi(\sqrt{\langle \bt,\bSigma\bt\rangle})+1/2$, and similarly 
$\E[e^{\min\{0,X_{\bt}\}}]=(2\pi\sigma_{\bt}^2)^{-1/2}\int_{-\infty}^{0}\exp(x-x^{2}/(2\sigma_{\bt}^{2}))\diff x+1/2=\exp(\sigma_{\bt}^{2}/2)(2\pi)^{-1/2}\int_{-\infty}^{-\sigma_{\bt}}\exp(-y^2/2)\diff y+1/2=\exp(\langle \bt,\bSigma\bt\rangle/2)\Phi(-\sqrt{\langle \bt,\bSigma\bt\rangle})+1/2$.

\begin{proposition}[concentration under weak signal]\label{prop:concentration:3}
     Assume that the potential function $U$ is even, $g$ is non-decreasing, and $\log\circ\ g$ is $G$-Lipschitz.

    (i) Suppose that Assumption \ref{assum:pi} holds true.
    We have for all $\delta\in(0,1]$, with probability at least $1-\delta$,
    \begin{align*}
        &\sup_{(\bw,b)\in\calT}(\Risk(\bw,b)-\Risk_{n}(\bw,b))-\E\left[\sup_{(\bw,b)\in\calT}(\Risk(\bw,b)-\Risk_{n}(\bw,b))\right]\\
        &\le \max\left\{\sqrt{\frac{L^{2}\left[K_{\Poincare}\left(4M_{\bX}^{\textnormal{num}}(G\btheta_{1})M_{\bX}^{\textnormal{num}}(2G\btheta_{1})-1\right)R_{\bw}^{2}+16g(+\theta_{0})g(-\theta_{0})M_{\bX}^{\textnormal{num}}(G\btheta_{1})R_{b}^{2}\right]}{n}},\frac{2LR_{b}}{n}\right\}\\
        &\quad\times\log(2\delta^{-1}).
    \end{align*}
     
    (ii) Suppose that Assumption \ref{assum:lsi} holds true.
    We have for all $\delta\in(0,1]$, with probability at least $1-\delta$,
    \begin{align*}
        \sup_{(\bw,b)\in\calT}(&\Risk(\bw,b)-\Risk_{n}(\bw,b))-\E\left[\sup_{(\bw,b)\in\calT}(\Risk(\bw,b)-\Risk_{n}(\bw,b))\right]\\
        \le \max&\left\{
        \sqrt{\frac{2L^{2}\log(\delta^{-1})}{n}}\sqrt{1+\frac{1}{2}\left(\log\frac{1}{g(-|\theta_{0}|)}+\log\frac{1}{M_{\bX}^{\textnormal{den}}(G\btheta_{1})}\right)}\right.\\
            &\times\left.\sqrt{5K_{\LogSobolev}M_{\bX}^{\textnormal{num}}(G\btheta_{1})M_{\bX}^{\textnormal{num}}(2G\btheta_{1})R_{\bw}^{2}+4eg(+\theta_{0})g(-\theta_{0})M_{\bX}^{\textnormal{num}}(G\btheta_{1})R_{b}^{2}},\frac{4LR_{b}\log(\delta^{-1})}{n}\right\}.
    \end{align*}
\end{proposition}

Note that the multiplicative factor $\log(1/g(-|\theta_{0}|)$ for $g(+\theta_{0})g(-\theta_{0})$ must appear for a log-Sobolev inequality for Bernoulli distributions corresponding to $\Gamma_{Y}$; see \citet{boucheron2013concentration}.

\subsubsection{Concentration under strong signal}
We now examine the case with a strong signal such that $\|\btheta_{1}\|\gg1$.
If $\|\btheta_{1}\|$ is large, then the input of $g$ inflate; $\langle \bX_{i},\btheta_{1}\rangle$ takes very large or small values, and $\theta_{0}$ should not affect the residual as long as they are of order $o(\|\btheta_{1}\|)$.
It is well-known that the empirical minimization problem \eqref{eq:empirical} performs well \citep[for example, see][]{kuchelmeister2024finite,hsu2024sample}.
Our result also shows that such a setting makes uniform generalization errors concentrate around their expectation well.

We define the following function:
\begin{equation*}
    \tilde{M}_{\bX}(\bt):=\frac{1}{\E\left[\exp\left(-|\langle\bX,\bt\rangle|\right)\right]},\ \bt\in\R^{d}.
\end{equation*}
Note that $1/\tilde{M}_{\bX}(\bt)\to0$ as $\|\bt\|\to\infty$ in general (we notice that $\lim_{\bt\to\infty}(1/\tilde{M}_{\bX}(\bt))=0$ is equivalent to the statement that for any sequence $\{\bt_{n}\}_{n\in\N}$ with $\|\bt_{n}\|\to\infty$, there exists a subsequence $\{\bt_{n}'\}_{n\in\N}\subset\{\bt_{n}\}_{n\in\N}$ such that $\lim_{n\to\infty}(1/\tilde{M}_{\bX}(\bt_{n}'))=0$; by the dominated convergence theorem, $\lim_{n\to\infty}(1/\tilde{M}_{\bX}(\bt_{n}'))=\E[\lim_{n\to\infty}\exp(-|\langle \bX,\bt_{n}'\rangle|)]$ as long as the limits exist; therefore, it is sufficient to show $1/|\langle \bX,\bt_{n}'\rangle|\to0$ almost surely for some $\{\bt_{n}'\}\subset\{\bt_{n}\}$ for any $\{\bt_{n}\}$; since $P_{\bX}$ is absolutely continuous with respect to the $d$-dimensional standard Gaussian measure, it is sufficient to consider $\bX\sim\Gauss_{d}(\zero_{d},\bfI_{d})$; $\langle \bX,\bt_{n}\rangle\sim\Gauss_{1}(0,\|\bt_{n}\|^{2})$, and the Borel--Cantelli lemma and condition $\|\bt_{n}\|\to\infty$ yield the conclusion).
If $\bX\sim\Gauss_{d}(\zero_{d},\bSigma)$ for some positive definite $\bSigma\in\R^{d\times d}$, then we obtain $1/\tilde{M}_{\bX}(\bt)=2\int_{0}^{\infty}\exp(-t)\exp(-t^{2}/(2\langle\bt,\bSigma\bt\rangle))/(2\pi\langle\bt,\bSigma\bt\rangle)^{1/2}\diff t=2\exp(\langle\bt,\bSigma\bt\rangle/2)\Phi(-\sqrt{\langle\bt,\bSigma\bt\rangle})$.
In addition, we prepare the following notation:
\begin{equation*}
    p_{\btheta_{1},\theta_{0}}:=\Pr\left(\langle\bX,\btheta_{1}\rangle\ge|\theta_{0}|\right).
\end{equation*}
For example, if we consider $\bX\sim\Gauss_{d}(\zero_{d},\bSigma)$, then $\Pr\left(\langle \bX,\btheta_{1}\rangle\ge|\theta_{0}|\right)=1-\Phi(|\theta_{0}|/\sqrt{\langle\btheta_{1},\bSigma\btheta_{1}\rangle})$.
\begin{proposition}[concentration under strong signal and logistic models]\label{prop:concentration:4}
Assume that the potential function $U$ is even and $g(t)=1/(1+\exp(-t))$.

(i) Suppose that Assumption \ref{assum:pi} holds.
    Then, for all $\delta\in(0,1]$, with probability at least $1-\delta$,
    \begin{align*}
        \sup_{(\bw,b)\in\calT}&(\Risk(\bw,b)-\Risk_{n}(\bw,b))- \E\left[\sup_{(\bw,b)\in\calT}(\Risk(\bw,b)-\Risk_{n}(\bw,b))\right]\\
        \le\max&\left\{\sqrt{\frac{L^{2}\left[K_{\Poincare}\left(1+e^{5|\theta_{0}|}/\tilde{M}_{\bX}(\btheta_{1})+\sqrt{e^{5|\theta_{0}|}/\tilde{M}_{\bX}(\btheta_{1})}\right)R_{\bw}^{2}+2\left(1+\sqrt{e^{5|\theta_{0}|}/\tilde{M}_{\bX}(\btheta_{1})}\right)R_{b}^{2}\right]}{n}},\right.\\
        &\quad\left.\frac{2LR_{b}}{n}\right\}\times\log(2\delta^{-1}).
    \end{align*}

    (ii) Suppose that Assumption \ref{assum:lsi} holds.
    Then, for all $\delta\in(0,1]$, with probability at least $1-\delta$,
    \begin{align*}
        &\sup_{(\bw,b)\in\calT}(\Risk(\bw,b)-\Risk_{n}(\bw,b))- \E\left[\sup_{(\bw,b)\in\calT}(\Risk(\bw,b)-\Risk_{n}(\bw,b))\right]\\
        &\le\max\left\{\sqrt{\frac{L^{2}\left(e+\log\left(\frac{1}{p_{\btheta_{1},\theta_{0}}}\right)\right)\left(2K_{\LogSobolev}\left(\frac{5}{4}+e^{5|\theta_{0}|}/\tilde{M}_{\bX}(\btheta_{1})\right)R_{\bw}^{2}+eR_{b}^{2}\right)\log(\delta^{-1})}{n}},\frac{4LR_{b}\log(\delta^{-1})}{n}\right\}.
    \end{align*}
\end{proposition}
\subsection{Comparison of the derived bounds}\label{sec:concentration:comparison}
In this section, we compare the concentration bounds derived above.
Specifically, we focus on three types of comparisons: (i) the general bounds for arbitrary $(\btheta_1,\theta_0)$ (Propositions \ref{prop:concentration:1}, \ref{prop:concentration:3}, and \ref{prop:concentration:4}) versus Proposition \ref{prop:concentration:0}, (ii) specific pairs of results (Propositions \ref{prop:concentration:2} and \ref{prop:concentration:3}, as well as Propositions \ref{prop:concentration:1} and \ref{prop:concentration:4}), and (iii) concrete bounds under Gaussian-logistic models.
Specifically, we only compare the terms of the order $\sqrt{1/n}$ rather than $2LR_{b}\log(\delta^{-1})/n$ and $4LR_{b}\log(\delta^{-1})/n$.

\subsubsection{Comparisons with Proposition \ref{prop:concentration:0}}
We compare Propositions \ref{prop:concentration:1} (weak bias), \ref{prop:concentration:3} (weak signal), and \ref{prop:concentration:4} (strong signal) with Proposition \ref{prop:concentration:0} (no bias) in the limits $|\theta_0|\to0$, $\|\btheta_1\|\to0$, and $\|\btheta_1\|\to\infty$, respectively.
We omit an analysis of Proposition \ref{prop:concentration:2} (strong bias) because its bound, despite its mild dependence on $\theta_0$, can still diverge in the limit $|\theta_0|\to\infty$; we leave its refinement as a future task.

\paragraph{Weak bias case.}
The bound provided by Proposition \ref{prop:concentration:1} remains non-vacuous compared to that of Proposition \ref{prop:concentration:0} as long as $|\theta_{0}|$ is not excessively large.
In particular, when $|\theta_{0}|=0$, the right-hand side of Proposition \ref{prop:concentration:1}-(i) coincides with that of Proposition \ref{prop:concentration:0}-(i), and the right-hand side of Proposition \ref{prop:concentration:1}-(ii) coincides with that of Proposition \ref{prop:concentration:0}-(ii) up to a constant factor of $3$.

\paragraph{Weak signal case.}
Lebesgue's dominated convergence theorem implies that $\lim_{\|\bt\|\to0}M_{\bX}^{\textnormal{num}}(\bt)=1$ and $\lim_{\|\bt\|\to0}M_{\bX}^{\textnormal{den}}(\bt)=1$.
Therefore, when $\btheta_{1}$ is close to zero, we obtain approximately the same upper bounds for the residuals as those in Proposition \ref{prop:concentration:0}-(i) and \ref{prop:concentration:0}-(ii), up to numerical factors of $\sqrt{3}$ and $\sqrt{5(1+(1/2)\log(1/g(-|\theta_{0}|)))}$, respectively, where we use $g(+\theta_{0})g(-\theta_{0})\le 1/4$.

\paragraph{Strong signal case.}
The right-hand side of Proposition \ref{prop:concentration:4}-(i) coincides with that of Proposition \ref{prop:concentration:0}-(i) in the limit $\|\btheta_{1}\|\to\infty$. 
We also observe that the upper bound of Proposition \ref{prop:concentration:4}-(ii) coincides with that of Proposition \ref{prop:concentration:0}-(ii) up to a moderate numerical constant ($\sqrt{(e+\log(2))5/4}\le 2.07$) if $p_{\btheta_{1},\theta_{0}}\approx1/2$ and $1/\tilde{M}_{\bX}(\btheta_{1})\ll1$.
For example, assuming $\bX_{i}\sim\Gauss_{d}(\zero_{d},\bSigma)$, the condition $\langle\btheta_{1},\bSigma\btheta_{1}\rangle \gg|\theta_{0}|^{2}$ is sufficient for these approximations to hold.

\subsubsection{Comparison of selected pairs among Propositions \ref{prop:concentration:1}--\ref{prop:concentration:4}}
We compare the pair of Propositions \ref{prop:concentration:2} and \ref{prop:concentration:3} and that of Propositions \ref{prop:concentration:1} and \ref{prop:concentration:4}, because the situations in which each pair applies are similar.
For these comparisons, we set $g(t)=\sigma(t)=1/(1+\exp(-t))$ (the logistic link function) and assume that $U$ is even.

\paragraph{Bounds for strong bias and weak signal.}
We compare the similar bounds of Propositions \ref{prop:concentration:2} and \ref{prop:concentration:3}, which target strong bias and weak signal, respectively, where the inputs have almost no effect on the outputs.
Since $1+(1/2)\log(1/\sigma(-|\theta_{0}|))\le (e+|\theta_{0}|)/2$ and $\sigma(+\theta_{0})\sigma(-\theta_{0})\le \exp(-|\theta_{0}|)$,
the bounds of Proposition \ref{prop:concentration:3} are better than those of Proposition \ref{prop:concentration:2} in the limit $\|\btheta_{1}\|\to0$. (Here, for simplicity, we use rough upper bounds for both cases; however, the best bounds derived in our proofs yield the same conclusion).
On the other hand, as $\|\btheta_{1}\|$ grows, $M_{\bX}^{\textnormal{num}}(\btheta_{1})M_{\bX}^{\textnormal{num}}(2\btheta_{1})=\omega(M_{\bX}^{2}(\btheta_{1}))$ holds in general (we let $G=1$ for the logistic link); hence, Proposition \ref{prop:concentration:2} provides a tighter bound.

\paragraph{Bounds for weak bias and strong signal.}
We study the differences between the bounds of Propositions \ref{prop:concentration:1} and \ref{prop:concentration:4}, which are derived for situations with weak bias and strong signal, respectively, where the output labels are well-balanced, i.e., $p_{Y}(y)\approx1/2$.
For sufficiently large $\|\btheta_1\|$, we have $\log (1/p_{\btheta_1,\theta_0})\approx \log2\approx0.6931$ and $1/\tilde{M_{\bX}}(\btheta_1)\approx0$.
Under this scenario, the bounds of Proposition \ref{prop:concentration:4} dominate those of Proposition \ref{prop:concentration:1}.
On the other hand, if $\|\btheta_1\|$ is not sufficiently large, Proposition \ref{prop:concentration:1} can yield tighter bounds.
To illustrate this, consider $\theta_0=0$ and compare Propositions \ref{prop:concentration:1}-(ii) and \ref{prop:concentration:4}-(ii) for simplicity; then, $\log (1/p_{\btheta_1,\theta_0})=\log 2$ and Proposition \ref{prop:concentration:1}-(ii) dominates Proposition \ref{prop:concentration:4}-(ii) when $9/2> (e+\log2)(5/4+1/\tilde{M}_{\bX}(\btheta_1))$.
This condition is equivalent to $1/\tilde{M}_{\bX}(\btheta_1)< 9/(2e+2\log2)-5/4\approx0.06910$.

\subsubsection{Comparison of bounds under a Gaussian-logistic model}
We compare the derived bounds under a Gaussian-logistic model, where the input vectors are $\bX_i\sim \calN_{d}(\zero_{d},\bSigma)$ for a positive definite matrix $\bSigma\in\R^{d\times d}$ and the output variables follow $p_{Y|\bX}(y|\bx)=1/(1+\exp(-y(\langle \bx,\btheta_1\rangle+\theta_0)))$, with $L=1$ (Table \ref{tab:summary}).
We consider this setup because it satisfies all the assumptions of the aforementioned propositions.
It is worth noting that some of these propositions offer distinct advantages beyond numerical improvement, such as operating under milder assumptions.
We omit Proposition \ref{prop:concentration:0} from this comparison because it requires $\theta_0=0$, rendering it incomparable with the other bounds that hold for arbitrary $\theta_0$ and $\btheta_1$.

\begin{table}[ht]
    \centering
    \begin{tabular}{ccc}\hline
        Statements & Bounds (given they are greater than $\frac{4LR_{b}\log(\delta^{-1})}{n}$)\\\hline
        Prop.~\ref{prop:concentration:1} & $\displaystyle\sqrt{\frac{\left(3+2|\theta_{0}|\right)\left[2\|\bSigma\|_{\op}\left(\frac{1}{2}+e^{2|\theta_{0}|}\right)R_{\bw}^{2}+eR_{b}^{2}\right]\log(\delta^{-1})}{n}}$\\
        Prop.~\ref{prop:concentration:2} & $\displaystyle\sqrt{\frac{\left(e+|\theta_{0}|\right)\left[\|\bSigma\|_{\op}\left(1/2+8e^{\langle\btheta_1,\bSigma\btheta_1\rangle}\right)R_{\bw}^{2}+4e^{1-|\theta_{0}|}e^{\langle\btheta_1,\bSigma\btheta_1\rangle/2}R_{b}^{2}\right]\log(\delta^{-1})}{n}}$\\
        Prop.~\ref{prop:concentration:3} &  $
        \sqrt{\frac{\left(2+\log(1+e^{|\theta_0|})+\log\frac{1}{M_{\bX}^{\textnormal{den}}(\btheta_1)}\right)\left(5\|\bSigma\|_{\op}M_{\bX}^{\textnormal{num}}(\btheta_1)M_{\bX}^{\textnormal{num}}(2\btheta_1)R_{\bw}^{2}+\frac{4eM_{\bX}^{\textnormal{num}}(\btheta_1)R_{b}^{2}}{(1+\exp(+\theta_0))(1+\exp(-\theta_0))}\right)\log(\delta^{-1})}{n}}$\\
        Prop.~\ref{prop:concentration:4}  & $\sqrt{\frac{\left(e-\log\left(1-\Phi\left(\frac{|\theta_{0}|}{\sqrt{\langle\btheta_{1},\bSigma\btheta_{1}\rangle}}\right)\right)\right)\left(2\|\bSigma\|_{\op}\left(5/4+\frac{e^{5|\theta_{0}|}}{\tilde{M}_{\bX}(\btheta_1)}\right)R_{\bw}^{2}+eR_{b}^{2}\right)\log(\delta^{-1})}{n}}$\\\hline
    \end{tabular}
    \caption{Applications of the derived bounds under the assumption of Gaussian inputs $\calN_{d}(\zero_{d},\bSigma)$, logistic outputs 
    $p_{Y|\bX}(y|\bx)=1/(1+\exp(-y(\langle \bx,\btheta_1\rangle+\theta_0))$, and $L=1$. Regarding Propositions \ref{prop:concentration:3} and \ref{prop:concentration:4}, we have $M_{\bX}^{\textnormal{den}}(\bt)=\exp(\langle \bt,\bSigma\bt\rangle/2)\Phi(-\sqrt{\langle \bt,\bSigma\bt\rangle})+1/2$, $M_{\bX}^{\textnormal{num}}(\bt)=\exp(\langle \bt,\bSigma\bt\rangle/2)\Phi(\sqrt{\langle \bt,\bSigma\bt\rangle}))+1/2$, and $1/\tilde{M}_{\bX}(\bt)=2\exp(\langle\bt,\bSigma\bt\rangle/2)\Phi(-\sqrt{\langle\bt,\bSigma\bt\rangle})$.}
    \label{tab:summary}
\end{table}
\subsubsection{Numerical illustration}
\begin{figure}[ht]
    \centering
    \includegraphics[width=0.95\linewidth]{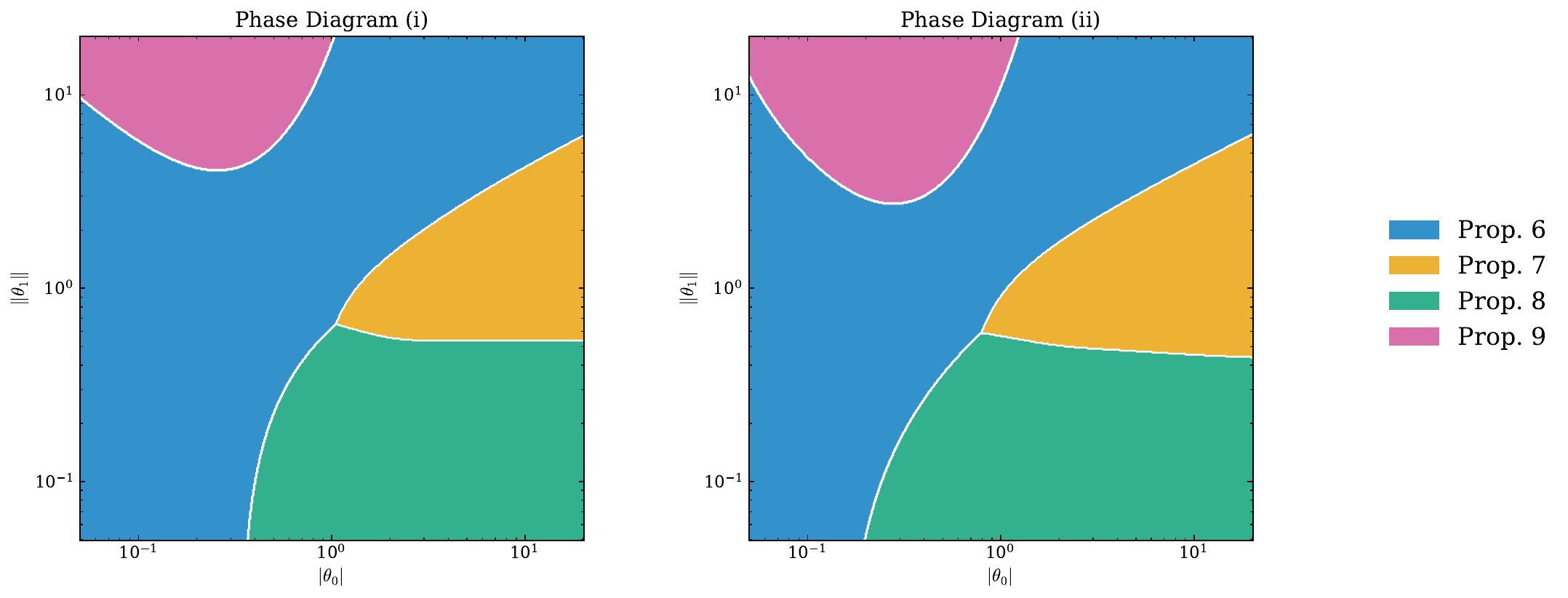}
    \caption{Comparison of the bounds with subexponential tails (i) (left) and subgaussian tails (ii) (right). 
    We set $\bX_i\sim \calN_{d}(0,\bfI_d)$ and $g(t)=\sigma(t)=1/(1+\exp(-t))$ (logistic model).
    The figures display which bound dominates the others by different colours (blue for Proposition \ref{prop:concentration:1}, orange for Proposition \ref{prop:concentration:2}, green for Proposition \ref{prop:concentration:3}, and magenta for Proposition \ref{prop:concentration:4}).}
    \label{fig:phase}
\end{figure}

We also show the phase transitions indicating which proposition yields the dominating bound (Figure \ref{fig:phase}).
In particular, we only compare the terms of order $\calO(\sqrt{\log(\delta^{-1})/n})$ by considering sufficiently large $n/\log(\delta^{-1})$.
We set $\bX_{i}\sim \calN_{d}(\zero_{d},\bfI_d)$, $g(t)=\sigma(t)=1/(1+\exp(-t))$ (the logistic link function), and $R_{\bw}=R_{b}=0$.
The $x$- and $y$-axes represent $|\theta_0|$ and $\|\btheta_1\|$, respectively, ranging from 0.05 to 20.0 on a logarithmic scale.
Each bound exhibits a spreading pattern of dominance over the other upper bounds, particularly toward extreme values; for example, the bound from Proposition \ref{prop:concentration:4} (magenta), designed for large $\|\btheta_1\|$, indeed expands its dominance as $\|\btheta_1\|$ grows (the top regions).
Similarly, the bound from Proposition \ref{prop:concentration:2} (orange) for large $|\theta_0|$ also exhibits expanding dominance toward larger $|\theta_0|$ (the right regions).

\subsection{Extension to non-central input vectors}\label{sec:concentration:noncentral}

We have discussed concentration inequalities based on the evenness of the potential function $U$.
However, one can be interested in situations where the input vectors have the nonzero mean vector $\bmu\neq\zero$.
To address this concern, let us consider the shifted dataset $\{(\bX_{i}+\bmu,Y_{i})\}_{i=1}^{n}$ for $\bmu\in\R^{d}$.
We can redefine the empirical risk function and conditional mass function of $Y_{i}$ given $\bX_{i}+\bmu$ as follows:
\begin{align*}
    \Risk_{n}^{\bmu}(\bw,b)&:=\frac{1}{n}\sum_{i=1}^{n}\ell\left(Y_{i}\left(\langle\bX_{i}+\bmu,\bw\rangle+b\right)\right)=\frac{1}{n}\sum_{i=1}^{n}\ell\left(\langle Y_{i}\bX_{i},\bw\rangle+Y_{i}(b+\langle \bmu,\bw\rangle )\right),\\
    p_{Y|\bX}^{\bmu}(y|\bx)&=g(\langle\bx+\bmu,\btheta_{1}\rangle+\theta_{0})=g(\langle\bx,\btheta_{1}\rangle+\theta_{0}+\langle\bmu,\btheta_{1}\rangle).
\end{align*}
We can immediately obtain non-central versions of the results above using this reparameterization.
The following corollary is an example using Proposition \ref{prop:concentration:1}-(ii); other examples using other propositions can be shown immediately.
\begin{corollary}\label{cor:concentration:noncentral}
    Assume that the potential function $U$ is even, $g$ is non-decreasing, and $\log\circ\ g$ is $G$-Lipschitz.
    Furthermore, suppose that Assumption \ref{assum:pi} holds true.
    Then, for any $\delta\in(0,1]$, with probability $1-\delta$,
\begin{align*}
        \sup_{(\bw,b)\in\calT}&(\Risk^{\bmu}(\bw,b)-\Risk_{n}^{\bmu}(\bw,b))
        -\E\left[\sup_{(\bw,b)\in\calT}(\Risk^{\bmu}(\bw,b)-\Risk_{n}^{\bmu}(\bw,b))\right]\\
        \le\max&\left\{\sqrt{\frac{L^{2}\left[K_{\Poincare}\left(e^{2G|\tilde{\theta}_{0}|}+\sqrt{e^{2G|\tilde{\theta}_{0}|}-1}\right)R_{\bw}^{2}+2(1+\sqrt{e^{2G|\tilde{\theta}_{0}|}-1})\left(\|\bmu\|^{2}R_{\bw}^{2}+R_{b}^{2}\right)\right]}{n}},\frac{2LR_{b}}{n}\right\}\\
        &\quad\times\log(2\delta^{-1}),
\end{align*}
where $\Risk^{\bmu}(\bw,b)=\E[\Risk_{n}^{\bmu}(\bw,b)]$ and $\tilde{\theta}_{0}=\theta_{0}+\langle\bmu,\btheta_{1}\rangle$.
\end{corollary}
Note that a trivial modification of Proposition \ref{prop:rademacher} yields
\begin{equation*}
        \E\left[\sup_{(\bw,b)\in\calT}(\Risk^{\bmu}(\bw,b)-\Risk_{n}^{\bmu}(\bw,b))\right]\le \frac{2L\left(R_{\bw}\sqrt{\tr(\E[(\bX_{i}+\bmu)(\bX_{i}+\bmu)^{\top}])}+R_{b}\right)}{\sqrt{n}}.
\end{equation*}
The order of the upper bound in Corollary \ref{cor:concentration:noncentral} is much milder than the order of this bound.

\section{Applications to asymptotic analysis}\label{sec:appl}

In this section, we apply the concentration results of Section~\ref{sec:concentration} to asymptotic regimes.
While we consider asymptotics and a sequence of statistical experiments indexed by $n$, we omit the explicit dependence of $L,R_{\bw},R_{b},K_{\Poincare},K_{\LogSobolev}$, and other quantities and random variables on $n$ for notational simplicity.

We now observe that the following almost sure convergences hold under some concise and reasonable conditions (for example, $|\theta_{0}|=\calO(1)$, $L=\calO(1)$, and $(K_{\LogSobolev}R_{\bw}^{2}+R_{b}^{2})\log n/n\to 0$ when we use Proposition \ref{prop:concentration:1}-(ii)):
\begin{align}
    \lim_{n\to\infty} &\left|\sup_{(\bw,b)\in\calT}(\Risk(\bw,b)-\Risk_{n}(\bw,b))-\E\left[\sup_{(\bw,b)\in\calT}(\Risk(\bw,b)-\Risk_{n}(\bw,b))\right]\right|=0\label{eq:dulln:1},\\
    \lim_{n\to\infty} &\left|\sup_{(\bw,b)\in\calT}(\Risk_{n}(\bw,b)-\Risk(\bw,b))-\E\left[\sup_{(\bw,b)\in\calT}(\Risk_{n}(\bw,b)-\Risk(\bw,b))\right]\right|=0\label{eq:dulln:2},
\end{align}
almost surely.
This follows immediately from the Borel–Cantelli lemma and the fact that sign changes preserve Lipschitz continuity.
The difference in signs does not matter because $-\ell$ is also $L$-Lipschitz if $\ell$ is.

\subsection{Effective rank and uniform law of large numbers}\label{sec:appl:ulln}

If the expectation terms in Eqs.~\eqref{eq:dulln:1} and \eqref{eq:dulln:2} tend to zero as $n\to\infty$,
the next uniform law of large numbers follows as a direct corollary of the almost sure convergences \eqref{eq:dulln:1} and \eqref{eq:dulln:2}:
\begin{equation}\label{eq:ulln}
    \lim_{n\to\infty} \sup_{(\bw,b)\in\calT}\left|\Risk_{n}(\bw,b)-\Risk(\bw,b)\right|=0.
\end{equation}
Proposition \ref{prop:rademacher} yields a sufficient condition for convergence of the expectation terms by flipping the sign of $\ell$; if $L^{2}R_{\bw}^{2}\tr(\E[\bX_{i}\bX_{i}^{\top}])/n\to0$ and $L^{2}R_{b}^{2}/n\to0$, then the uniform law \eqref{eq:ulln} holds true.
In particular, consider the case where $L,R_{\bw},R_{b},K_{\LogSobolev}$, and $\|\E[\bX_{i}\bX_{i}^{\top}]\|_{\textnormal{op}}$ are fixed in $n$ and $d^{\ast}=\tr(\E[\bX_{i}\bX_{i}^{\top}])/\|\E[\bX_{i}\bX_{i}^{\top}]\|_{\textnormal{op}}$ varies according to $n$.
Then, the sufficient condition for the uniform law is $d^{\ast}/n\to0$.
This generalizes the classical condition $d/n\to0$ by replacing the ambient dimension with the effective rank.

\begin{remark}
    We explain the differences between our asymptotic and non-asymptotic results and those of \citet{nakakita2026dimension} for clarity.
    \citet{nakakita2026dimension} considers a uniform law only for the logistic loss function under the isoperimetry of the marginal distribution of $\bX_{i}$.
    Our results imply uniform laws for arbitrary Lipschitz loss functions under the isoperimetry of the joint distribution of $(\bZ_{i},Y_{i})$.
    Hence, our results and that of \citet{nakakita2026dimension} are complementary in the regime $d^{\ast}/n\to0$, but differ in the form of concentration and the distributions they assume isoperimetry over.
    Moreover, our results can show the almost sure convergences \eqref{eq:dulln:1} and \eqref{eq:dulln:2} even if $d^{\ast}/n=\Theta(1)$, while \citet{nakakita2026dimension} only yields the tightness of the uniform generalization error.
    This is because we consider concentration around means, and \citet{nakakita2026dimension} deals with concentration around zero.
\end{remark}

\subsection{Biased convergence in proportionally high-dimensional settings}\label{sec:appl:proportional}
We also consider proportionally high-dimensional settings following the setup of prior studies such as \citet{sur2019modern} and \citet{salehi2019impact}.
Let us set $\bX_{i}\sim^{\textnormal{i.i.d.}}\Gauss_{d}(\zero_{d},(1/d)\bfI_{d})$ and $g=\sigma$ (logistic link); then $U(\bx)=(d/2)\|\bx\|^{2}$ and $K_{\LogSobolev}=1/d$ since $U$ is $d$-convex.
We set $R_{\bw}=\sqrt{d}R_{1}$ for some fixed $R_{1}>0$.
It reflects the typical assumption that each element of the true value of the parameter is independent and identically distributed.

A Rademacher complexity bound for the expected uniform generalization error is of order $\calO(LR_{1})$ (Proposition \ref{prop:rademacher}), and thus does not vanish asymptotically when  $R_{1}=\Omega(1)$.
This result can be predicted by the result of \citet{sur2019modern}; they show that the maximum likelihood estimator for logistic regression converges with bias in proportional high-dimensional settings.
Since our setting for $\ell$ includes the logistic regression loss, the expected uniform generalization error should not converge to zero.

On the other hand, the residuals in Propositions \ref{prop:concentration:0}--\ref{prop:concentration:4} under Assumption \ref{assum:lsi} are of order $\calO(L(R_{1}+R_{b})(\sqrt{\log(\delta^{-1})/n}+\log(\delta^{-1})/n))$ (ignoring $\theta_{0}$ and $\btheta_{1}$ for a unified treatment).
Therefore, uniform generalization errors rapidly converge to their expectation in proportionally high-dimensional settings.
This does not contradict existing results, which primarily characterize convergence of the maximum likelihood estimator to a biased limit in high dimensions.
The point is that maximum likelihood estimation converges anyway.
Our result is similar to it that the generalization error concentrates around its (possibly biased) expectation, even if the expectation itself does not vanish asymptotically.

Note that we can examine and characterize arbitrary Lipschitz functions of $(\bZ_{i},Y_{i})$ in proportionally high-dimensional settings because our proof strategy can be extended to other Lipschitz functions of them.

\section{Discussion}\label{sec:discussion}
We discuss the advantages and remaining challenges of our results.
Specifically, we compare our results with alternative approaches, such as the tail bound derived by \citet{adamczak2008tail} and a na\"{i}ve approach based on fixing the $Y_i$ variables.
Furthermore, we examine the surrogate loss results used to derive bounds for the 0-1 loss $\mathbbm{1}_{(-\infty,0]}(t)$, as well as their relevance to recent studies on max-margin classifiers.

\subsection{Advantages over general concentration bounds}\label{sec:discussion:advantages}
The approaches of \citet{adamczak2008tail}, \citet{van2013bernstein}, and \citet{lederer2014new} provide sophisticated strategies for establishing high-probability bounds for uniform generalization errors under unbounded losses.
However, in our specific setting, the approach of \citet{van2013bernstein} yields concentration around a value bounded below by the expectation of the uniform generalization error, whereas applying the method of \citet{lederer2014new} results in a heavier tail for the concentration bound.
Therefore, we focus our comparison on the results of \citet{adamczak2008tail}, which necessitates a detailed evaluation to facilitate this analysis.

We first introduce the necessary preliminary definitions.
For $\alpha \in (0,1]$, let $\psi_{\alpha}:[0,\infty)\to[0,\infty)$ be defined by $\psi_{\alpha}(x)=\exp(x^\alpha)-1$, and define the Orlicz norm of a real-valued random variable $X$ as
\begin{equation*}
    \|X\|_{\psi_{\alpha}}=\inf\{c>0:\E[\psi_{\alpha}(|X|/c))]\le 1\};
\end{equation*}
Note that $\|\cdot\|_{\psi_{\alpha}}$ is a quasi-norm rather than a true norm when $\alpha<1$.
This norm satisfies the inequality $\|X\|_{L^2}\le C_{\alpha}\|X\|_{\psi_{\alpha}}$ for a constant $C_\alpha$ depending solely on $\alpha\le 1$ \citep{vladimirova2020sub}.
Then, for all $\alpha\in(0,1]$, $\eta\in(0,1)$, and $\epsilon>0$, there exists a constant $C_{\alpha,\eta,\epsilon}$ depending solely on $\alpha,\eta,\epsilon$ such that for all $t\ge0$,
\begin{align}\label{eq:adamczak2008}
    &\Pr\left(\sup_{(\bw,b)\in\calT}\left|\Risk(\bw,b)-\Risk_{n}(\bw,b)\right|
    \ge (1+\eta)\E\left[\sup_{(\bw,b)\in\calT}\left|\Risk(\bw,b)-\Risk_{n}(\bw,b)\right|\right]+t\right)\notag\\
    &\le \exp\left(-\frac{nt^{2}}{2(1+\epsilon)\sup_{(\bw,b)\in\calT}\E[(\ell(Y_{i}(\langle \bX_i,\bw\rangle+b))-\E[\ell(Y_{i}(\langle \bX_i,\bw\rangle+b))])^{2}]}\right)\notag\\
    &\quad+3\exp\left(-\left(\frac{nt}{C_{\alpha,\eta,\epsilon}\|\max_{i}\sup_{(\bw,b)\in\calT}\left|\ell(Y_{i}(\langle \bX_i,\bw\rangle+b))-\E[\ell(Y_{i}(\langle \bX_i,\bw\rangle+b))]\right|\|_{\psi_{\alpha}}}\right)^{\alpha}\right).
\end{align}
While the first term on the right-hand side is relatively straightforward to control, the second term introduces a heavier dependence on the covariance of $\bX_{i}$.
To explicitly evaluate this dependence, we assume that the marginal distribution of $(\bX_i,Y_i)$ is given by $(\bX_i,Y_i)\sim\calN_{d}(\zero_{d},\bSigma)\otimes\textnormal{Unif}(\{\pm1\})$.
Since it holds that for either the logistic loss $\ell(t)=\ell_{\textnormal{L}}(t)=\log(1+\exp(-t))$ or the hinge loss $\ell_{\textnormal{H}}(t)=\max\{0,1-t\}$, 
\begin{equation*}
    R_{\bw}\|\bX_i\|+R_b \le \sup_{(\bw,b)\in\calT}\left|\ell(Y_{i}(\langle \bX_i,\bw\rangle+b))\right|\le 1+R_{\bw}\|\bX_i\|+R_b
\end{equation*}
and 
\begin{equation*}
    0\le |\ell(Y_{i}(\langle \bX_i,\bw\rangle +b))|\le 1+|\langle \bX_i,\bw\rangle|+R_{b},\ (\bw,b)\in\calT,
\end{equation*}
it follows that
\begin{align*}
    &\left\|\max_{i=1,\ldots,n}\sup_{(\bw,b)\in\calT}\left|\ell(Y_{i}(\langle \bX_i,\bw\rangle+b))-\E[\ell(Y_{i}(\langle \bX_i,\bw\rangle+b))]\right|\right\|_{\psi_{\alpha}}\\
    &\ge C_{\alpha}^{-1}\E\left[\sup_{(\bw,b)\in\calT}\left(\ell(Y_{i}(\langle \bX_i,\bw\rangle+b))-\E[\ell(Y_{i}(\langle \bX_i,\bw\rangle+b))]\right)^{2}\right]^{1/2}\\
    &\ge C_{\alpha}^{-1}\left(\E\left[\sup_{(\bw,b)\in\calT}\ell(Y_{i}(\langle \bX_i,\bw\rangle+b))^{2}\right]^{1/2}-\sup_{(\bw,b)\in\calT}\E[\ell(Y_{i}(\langle \bX_i,\bw\rangle+b))]\right)\\
    &\ge C_{\alpha}^{-1}\left(R_{\bw}\E[\|\bX_i\|^{2}]^{1/2}-2R_{b}-1-\sup_{\bw}\E[|\langle \bX_i,\bw\rangle|]\right)=C_{\alpha}^{-1}\left(R_{\bw}\left(\sqrt{\tr(\bSigma)}-\sqrt{\|\bSigma\|_{\op}}\right)-2R_{b}-1\right),
\end{align*}
where the last two inequalities follow from the triangle inequality.
Consequently,  we obtain the following high-probability bound: with probability at least $1-\delta$,
\begin{align*}
    \sup_{(\bw,b)\in\calT}\left|\Risk(\bw,b)-\Risk_{n}(\bw,b)\right|
    &\le (1+\eta)\E\left[\sup_{(\bw,b)\in\calT}\left|\Risk(\bw,b)-\Risk_{n}(\bw,b)\right|\right]\\
    &\quad+\frac{C_{\alpha,\eta,\delta}}{C_{\alpha}}\frac{R_{\bw}(\sqrt{\tr(\bSigma)}-\sqrt{\|\bSigma\|_{\op}})-2R_{b}-1}{n}\left(\log\frac{6}{\delta}\right)^{1/\alpha}+r,
\end{align*}
where $r=r(\delta,L,R_{\bw},R_{b},\bSigma,n,\alpha,\eta,\epsilon,\delta)\ge0$ is a non-negative number representing the remainder term.
Even if the remainder $r$ is negligibly small, the uniform generalization error on the left-hand side may fail to concentrate around its (scaled) expectation.
For example, the second term on the right-hand side is $\Omega(1)$ if $\bSigma=\bfI_{d}$, $d\gg n^2$, and $R_{\bw}=1$, and our bound still shows a small residual because $K_{\LogSobolev}=1$.
\subsection{Remark on a fixed-label approach}
One might expect that conditioning on $Y_i=y_i\in\{\pm1\}$ and employing the concentration of $(Z_i)_{i=1}^{n}$ given $Y_i=y_{i}$ for each $i$ would suffice to derive the bounds, potentially yielding sharper multiplicative constants.
Specifically, we can consider the following modification of the empirical risk:
\begin{equation*}
    \Risk_{n}(\bw,b|(y_{i})_{i=1}^{n})=\frac{1}{n}\sum_{i=1}^{n}\ell(y_{i}(\langle\bX_i,\bw\rangle+b))=\frac{1}{n}\sum_{i=1}^{n}\ell(\langle\bZ_i,\bw\rangle+by_{i}).
\end{equation*}
This approach is indeed appealing, as its concentration is expected to be sharper than that of the original $\Risk_{n}(\bw,b)$.
However, this introduces a fundamental problem regarding the quantity to which the modified risk concentrates.
Here, we discuss the following four types of candidates: for all $t\ge 0$,
\begin{align*}
    & \max_{(y_i)_i}\Pr\left(\sup_{\bw,b}\left(\Risk(\bw,b)-\Risk_{n}(\bw,b|(y_{i})_i)\right)-\E\left[\sup_{\bw,b}\left(\Risk(\bw,b)-\Risk_{n}(\bw,b|(y_{i})_i)\right)\right]\ge t\right)\le \cdots,\\
    & \max_{(y_i)_i}\Pr\left(\sup_{\bw,b}\left(\Risk(\bw,b|(y_i)_i)-\Risk_{n}(\bw,b|(y_{i})_{i})\right)-\E\left[\sup_{\bw,b}\left(\Risk(\bw,b|(y_i)_{i})-\Risk_{n}(\bw,b|(y_{i})_i)\right)\right]\ge t\right)\le \cdots,\\
    &\Pr\left(\sup_{\bw,b,(y_{i})_i}\left(\Risk(\bw,b)-\Risk_{n}(\bw,b|(y_{i})_i)\right)-\E\left[\sup_{\bw,b,(y_i)_i}\left(\Risk(\bw,b)-\Risk_{n}(\bw,b|(y_{i})_i)\right)\right]\ge t\right)\le \cdots,\\
    & \Pr\left(\sup_{\bw,b,(y_i)_i}\left(\Risk(\bw,b|(y_i)_i)-\Risk_{n}(\bw,b|(y_{i})_{i})\right)-\E\left[\sup_{\bw,b,(y_i)_i}\left(\Risk(\bw,b|(y_i)_{i})-\Risk_{n}(\bw,b|(y_{i})_i)\right)\right]\ge t\right)\le \cdots,
\end{align*}
where $\Risk(\bw,b|(y_i)_{i})=\E_{(\bZ_i)_i}[\Risk_{n}(\bw,b|(y_i)_{i})]$.
Although all of these formulations should exhibit rapid concentration, their corresponding expectations lack meaningful interpretations.
Because $\Risk(\bw,b|(y_i)_{i})$ does not characterize the predictive performance of $(\bw,b)$ on unseen data, the second and fourth formulations offer limited utility.
Similarly, the remaining first and third types of bounds lack practical significance because the expectation terms can become excessively large due to pathological choices of $(y_i)_i$. Examples include $y_i=\argmin_{y_{i}}p_{Y}(y_i)$, or $y_i=\argmin_{y_{i}}p_{Y|\bZ}(y_i|\bZ_i)$ in the third bound.
Furthermore, it should be noted that the estimation of these terms cannot be reduced to their Rademacher complexities due to the fixed $(y_i)_i$ and their asymmetric structures \citep[see, e.g.,][]{wainwright2019high}.
\subsection{Applications to the 0-1 loss}\label{sec:discussion:01loss}
We can establish dimension-free excess risk bounds for the 0-1 loss $\ell(t)=\mathbbm{1}_{(-\infty,0]}(t)$ ($t\in\R$) by combining well-established results on surrogate losses \citep{zhang2004statistical,bartlett2006convexity} with our generalization error analysis for the hinge loss.
Since the 0-1 loss is bounded, McDiarmid's inequality ensures concentration for the difference between the corresponding uniform generalization error and its expectation.
However, bounds on the expected uniform generalization error for the 0-1 loss are typically derived using the Vapnik--Chervonenkis dimension and are therefore dimension-dependent.

Let $\Risk_{\textnormal{L}}$ and $\Risk_{\textnormal{L},n}$ denote the population and empirical risk functions under the logistic loss.
Furthermore, we define $\Risk_{01}(\bw,b)=\Pr(Y_i(\langle \bX_i,\bw\rangle+b))$ as the population risk under the 0-1 loss.
By Theorem 2.1 and Section 3.5 of \citet{zhang2004statistical}, we obtain that for all $(\bw,b)\in\R^{d+1}$,
\begin{equation*}
    \left(\Risk_{01}(\bw,b)-\Risk_{01}^\ast\right)^2\le 2\left(\Risk_{\textnormal{L}}(\bw,b)-\Risk_{\textnormal{L}}^\ast\right),
\end{equation*}
where $\eta(\bX_i)=\Pr(Y_i=+1|\bX_i)=p_{Y|\bX}(+1|\bX_i)$, $\Risk_{01}^\ast=\E[\min\{\eta(\bX_i),1-\eta(\bX_i)\}]$, and $\Risk_{\textnormal{L}}^\ast=\E[-\eta(\bX_i)\log\eta(\bX_i)]$, which represent the corresponding optimal risks (i.e., Bayes risks).
If we assume that $\eta(\bX_i)=1/(1+\exp(-(\langle \bX_i,\btheta_1\rangle+\theta_0))$ for $(\btheta_1,\theta_0)\in\calT$, then the minimum of the right-hand side is zero.
Our uniform generalization error bounds guarantee that (i) the empirical risk minimization problem closely approximates the population risk minimization problem, and thus (ii) the empirical risk minimizer achieves a small 0-1 risk as well.

Alternatively, one can consider the hinge loss instead of the logistic loss.
Let $\Risk_{\textnormal{H}}$ and $\Risk_{\textnormal{H},n}$ denote the population and empirical risk functions under the hinge loss.
Theorem 1 of \citet{bartlett2006convexity} implies that for all $(\bw,b)\in\R^{d+1}$,
\begin{equation*}
    \Risk_{01}(\bw,b)-\Risk_{01}^\ast\le \Risk_{\textnormal{H}}(\bw,b)-\Risk_{\textnormal{H}}^\ast,
\end{equation*}
where $\Risk_{\textnormal{H}}^\ast=\E[2\min\{\eta(\bX_i),1-\eta(\bX_i)\}]$ denotes the corresponding Bayes risk.
If $\min_{(\bw,b)\in\calT}\Risk_{\textnormal{H}}(\bw,b)-\Risk_{\textnormal{H}}^\ast\ll1$, then empirical risk minimization results in faster convergence of the population misclassification probability to the Bayes risk compared to that obtained via the logistic loss.
\subsection{Overparameterized regime}
We consider a binary linear classification problem in the overparameterized regime.
If the ratio $d/n$ exceeds certain thresholds, there exist directions $\bw$ such that $Y_i\langle \bX_i,\bw\rangle>0$ for all $i=1,\ldots,n$ with high probability; this geometric property of the data is referred to as linear separability \citep{cover1964geometrical,cover1965geometrical,sur2019modern}.
The generalization error of such linearly separating vectors $\bw$ has been widely investigated; for instance, \citet{liang2022precise} and \citet{montanari2025generalization} studied the generalization performance of max-$\ell_1$-margin and max-$\ell_2$-margin classifiers under the 0-1 loss, respectively.
In particular, their analyses operate within a proportionally high-dimensional regime, which requires a slow decay of the eigenvalues of the covariance matrix $\bSigma$ of the centred Gaussian input vectors $\bX_i$.
In contrast, we examine the generalization error under a fast decay of the eigenvalues; specifically, we consider the setting where $\tr(\bSigma)\ll n$.
Even under this condition, linearly separating directions can still exist.

Our bounds, combined with the discussion in Section \ref{sec:discussion:01loss}, imply that if $(\bX_i,Y_i)$ has the marginal distribution $\bX_i\sim \calN_d(\zero_{d},\bSigma)$ with $\tr(\bSigma)\ll n$
and satisfies certain mild conditions (e.g., $\Pr(Y_i=+1|\bX_i)=1/(1+\exp(-Y_i\langle\bX_i,\btheta_1\rangle))$ for some $\btheta_1$ with $\|\btheta_1\|\le R_{\bw}$ and $\max\{R_{\bw},R_{b}\}=\calO(1)$ to employ Proposition \ref{prop:concentration:0}), 
the uniform generalization errors $\sup_{\bw:\|\bw\|_2\le 1}(\Risk(\bw)-\Risk_n(\bw))$ under the 0-1 and logistic losses decay with high probability (see Proposition \ref{prop:rademacher}).
Consequently, $\bw$ with a small empirical risk $\Risk_n(\bw)\ll 1$ also achieves a small population risk $\Risk(\bw)\ll1$.

\subsection{Towards further improvements}\label{sec:discussion:further}
We present numerical experiments that support our conjecture in Section \ref{sec:concentration:largebias} (i.e., that $|\theta_0|\gg1$ also drives the bound to converge to that of Proposition \ref{prop:concentration:0} up to certain numerical multiplicative constants) and suggest further possibilities for improvement.
We generate data according to $\bX_i\sim\calN_{d}(\zero_{d},\bfI_d)$ and $\Pr(Y_i=y|\bX_i)=1/(1+\exp(-Y_i(\langle \bX_i,\btheta_1\rangle+\theta_0)))$, setting $d=500$ and $n=500$ across different configurations of $\btheta_1$ and $\theta_0$.
For the estimation step, we set $R_{\bw}=R_b=1$ and consider the logistic regression problem with the loss function $\ell(t)=\ell_{\textnormal{L}}(t)=\log(1+\exp(-t))$.
By generating $500$ validation samples, we approximate the maximum absolute gap between the population and empirical risks using the absolute difference between the empirical risks evaluated on the training and validation datasets.
Since the maximum absolute gap is non-convex, we maximize it using the gradient descent algorithm initialized with $500$ different random starting values at each iteration.
Finally, to evaluate the concentration phenomenon, we compute the variance of the $1000$ maximal absolute gaps generated across $1000$ independent iterations for each setting.
The numerical experiments are conducted under two scenarios: (a) $\theta_0=0.0, 0.5, 1.0,\ldots,14.5,15.0$ with a fixed vector $\btheta_1=(1,0,\ldots,0)$, and (b) $\|\btheta_1\|=0.0, 0.5, 1.0,\ldots,14.5,15.0$ with a fixed bias $\theta_0=15.0$.

\begin{figure}[ht]
    \centering
    
    \includegraphics[width=0.95\linewidth]{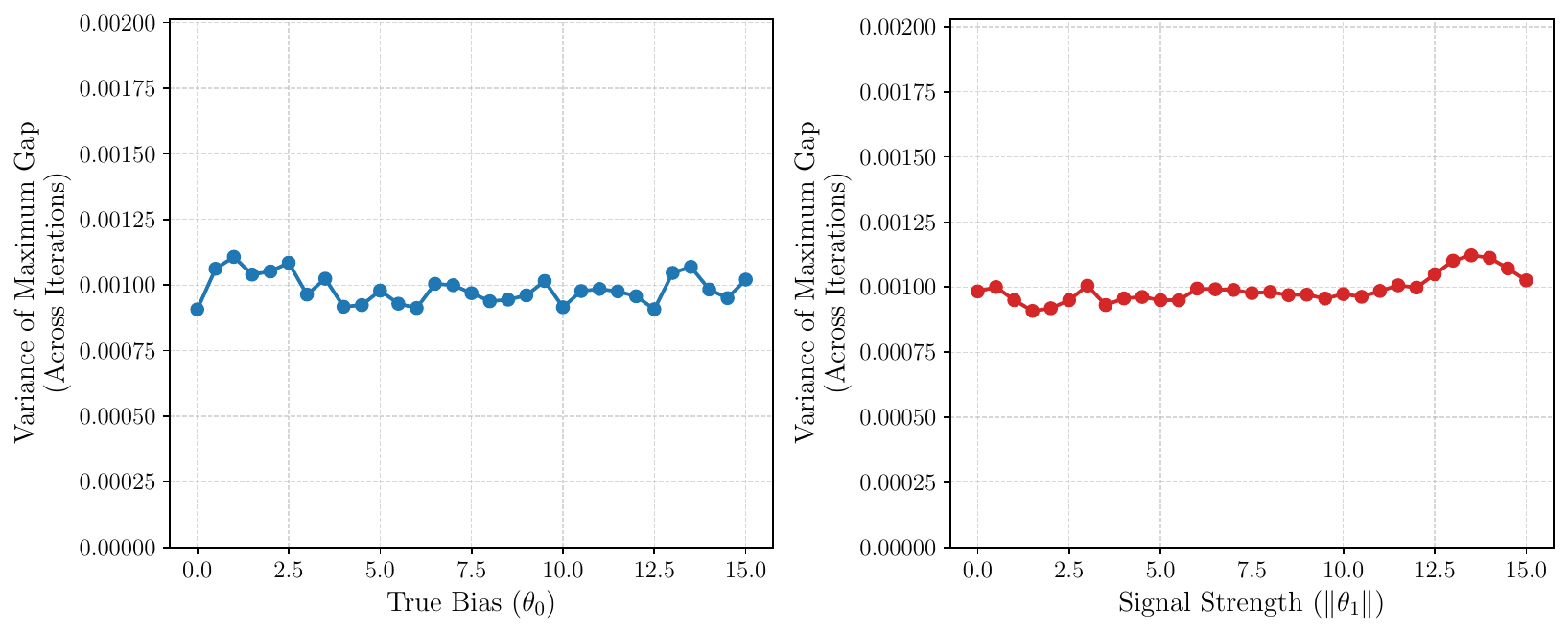}
    
    \caption{The transition of the variance of empirical generalization errors with different true bias coefficients $\theta_0=0.0,0.5,1.0,\ldots,14.5,15.0$ (left) and different true signal coefficients $\|\btheta_1\|=0.0,0.5,1.0,\ldots,14.5,15.0$ (right).}
    \label{fig:towards}
\end{figure}

Figure \ref{fig:towards} displays the experimental results.
We observe that the variance, reflecting the degree of concentration, remains stable across the different configurations of $\btheta_1$ and $\theta_0$.
This finding suggests two directions for future research: first, deriving an improved bound over Proposition \ref{prop:concentration:2} for large $|\theta_0|$, as conjectured; second, establishing a unified bound independent of $\btheta_1$ and $\theta_0$, which represents a more challenging and ambitious objective.

\section{Concluding remarks}\label{sec:concluding}

We have developed functional inequalities tailored to binary linear classification, and derived improved concentration bounds under a variety of bias and signal regimes.
The small numerical constants in our results allow for refined evaluation of uncertainty in binary linear classification tasks.
Moreover, our proof strategy can be adapted to different problem setups.
In particular, the idea to evaluate the isoperimetry of some problem-specific random vectors should be helpful in other supervised and unsupervised learning problems.

We refer to potential future research directions.
As \citet{johnson2017discrete} considers the Bakry--\'{E}mery condition for a discrete space $\Z_{+}$, more tractable criteria for functional inequalities on hybrid continuous--discrete spaces such as $\R^{d}\times\{\pm1\}$ and derivation of sufficient conditions generalizing Theorem \ref{thm:functional} should be of interest.
Extension to nonlinear classification problems (in particular, classification with neural networks) is also an important and topical problem.
Furthermore, the studies on logistic regression under the regime $d/n\to0$ \citep{kuchelmeister2024finite,hsu2024sample} derive the phase transitions in the rates of convergence and the sample complexities; its dimension-free versions should be an important direction.

\section*{Acknowledgements}
This work is supported by JSPS KAKENHI Grant Number JP24K02904.
The author gratefully acknowledges the associate editor and two anonymous reviewers for helpful comments.
The author used ChatGPT and Google Gemini to improve the grammatical accuracy of the manuscript, as a non-native English speaker, and to write the source code for numerical illustrations (Figures \ref{fig:phase} and \ref{fig:towards}), which is checked and corrected by the author.

\bibliographystyle{apalike}
\bibliography{bibliography}

\appendix

\section{Proofs of the results in Section \ref{sec:functional}}
\subsection{Proof of Theorem \ref{thm:functional}}\label{sec:proof:functional}
We cite a technical result used in the proof of Theorem \ref{thm:functional}.
\begin{lemma}[\citealp{chen2021dimension}]\label{lem:ccn21lem1}
    Let $\pi$ and $\rho$ be two probability measures.
    For any non-negative function $f$ with $\E_{\pi}[f]<\infty$,
    We have
    \begin{equation*}
        \E_{\pi}\left[f\log\left(\frac{\E_{\pi}[f]}{\E_{\rho}[f]}\right)\right]\le \Ent_{\pi}(f)+\E_{\pi}[f]\log\left(1+\chi^{2}(\pi\|\rho)\right).
    \end{equation*}
\end{lemma}

\begin{proof}[Proof of Theorem \ref{thm:functional}]
    As the basis of our proof strategy, we employ the proof of  \citet{chen2021dimension} partially and introduce a decoupling-like argument.

(Step 1: Poincar\'{e} inequality)
We set a random variable $Y'$, satisfying the following two conditions:
\begin{align*}
    Y'\overset{\calL}{=}Y,\ (\bZ,Y)\perp Y'.
\end{align*}
Note the following decomposition:
\begin{equation*}
    \Var(f(\bZ,Y))=\E[\Var(f(\bZ,Y)|Y)]+\Var(\E[f(\bZ,Y)|Y]).
\end{equation*}
The first term on the right-hand side is bounded as
\begin{equation*}
    \E[\Var(f(\bZ,Y)|Y)]\le \E\left[K_{\Poincare}\Gamma_{\bZ}(f)(\bZ,Y)\right].
\end{equation*}
Regarding the second term, by definition, we have
\begin{equation*}
    \Var(\E[f(\bZ,Y)|Y])=\E\left[\left|\E[f(\bZ,Y)|Y]-\E[f(\bZ,Y)]\right|^{2}\right].
\end{equation*}
We have for each $y\in\{\pm1\}$,
\begin{equation*}
    \E[f(\bZ,Y)|Y=y]-\E[f(\bZ,Y)]=\left(\E_{\bZ|Y}[f(\bZ,Y)|Y=y]-\E_{\bZ}[f(\bZ,y)]\right)+\E_{\bZ Y}[f(\bZ,y)-f(\bZ,Y)].
\end{equation*}
Therefore, Young's inequality yields that for any $c,c^{\ast}\in[1,\infty]$ with $1/c+1/c^{\ast}=1$,
\begin{align}
    \Var(\E[f(\bZ,Y)|Y])
    &\le c\underbrace{\E_{Y'}\left[\left|\E_{\bZ|Y}[f(\bZ,Y)|Y=Y']-\E_{\bZ}[f(\bZ,Y')]\right|^{2}\right]}_{\text{How $Y$ is informative to $\bZ$}}\notag\\
    &\quad+c^{\ast}\underbrace{\E_{Y'}\left[\left|\E_{\bZ Y}[f(\bZ,Y)-f(\bZ,Y')]\right|^{2}\right]}_{\text{How $\bZ$ is informative to $Y$}}.\label{eq:thm1:proof:decom1}
\end{align}
For the first term on the right-hand side of \eqref{eq:thm1:proof:decom1}, since for each $y\in\{\pm1\}$ and $c\in\R$,
\begin{align*}
    \E_{\bZ |Y}[f(\bZ,Y)|Y=y]-\E_{\bZ}[f(\bZ,y)]&=\int_{\R^{d}} f(\bz,y)\left(1-\frac{P_{\bZ}(\diff\bz)}{P_{\bZ|Y}(\diff\bz|y)}\right)P_{\bZ|Y}(\diff\bz|y)\\
    &=\int_{\R^{d}} (f(\bz,y)-c)\left(1-\frac{P_{\bZ}(\diff\bz)}{P_{\bZ|Y}(\diff\bz|y)}\right)P_{\bZ|Y}(\diff\bz|y),
\end{align*}
the Cauchy--Schwarz inequality yields
\begin{align*}
    \E_{Y'}\left[\left|\E_{\bZ |Y}[f(\bZ,Y)|Y=Y']-\E_{\bZ}[f(\bZ,Y')]\right|^{2}\right]
    &\le \E_{Y'}\left[\Var(f(\bZ,Y)|Y=Y')\chi^{2}(P_{\bZ}(\cdot)\|P_{\bZ|Y}(\cdot|Y'))\right]\\
    &\le  \E_{Y'}\left[\Var(f(\bZ,Y)|Y=Y')\max_{y\in\{\pm1\}}\chi^{2}(P_{\bZ}(\cdot)\|P_{\bZ|Y}(\cdot|y))\right]\\
    &= K_{\chi^{2}}\E\left[\Var(f(\bZ,Y)|Y)\right]\\
    &\le K_{\chi^{2}}K_{\Poincare}\E\left[\Gamma_{\bZ}(f)(\bZ,Y)\right].
\end{align*}
For the second term on the right-hand side of \eqref{eq:thm1:proof:decom1},  we have
\begin{align*}
    &\E_{Y'}\left[\left|\E_{\bZ Y}[f(\bZ,Y)-f(\bZ,Y')]\right|^{2}\right]\\
    &=\sum_{y'\in\{\pm1\}}p_{Y}(y')\left(\sum_{y\in\{\pm1\}}\int_{\R^{d}}(f(\bz,y)-f(\bz,y'))P_{\bZ Y}(\diff\bz,\{y\})\right)^{2}\\
    &=\sum_{y'\in\{\pm1\}}p_{Y}(y')\left(\int_{\R^{d}}(f(\bz,-y')-f(\bz,y'))P_{\bZ Y}(\diff\bz,\{-y'\})\right)^{2}\\
    &=\sum_{y'\in\{\pm1\}}p_{Y}(y')\left(\int_{\R^{d}} 2\nabla_{y}f(\bz,y')P_{\bZ Y}(\diff\bz,\{-y'\})\right)^{2}\\
    &=\sum_{y'\in\{\pm1\}}p_{Y}(y')p_{Y}(-y')^{2}\left(\frac{1}{p_{Y}(-y')}\int_{\R^{d}} 2\nabla_{y}f(\bz,y')P_{\bZ Y}(\diff\bz,\{-y'\})\right)^{2}\\
    &=p_{Y}(+1)p_{Y}(-1)\sum_{y'\in\{\pm1\}}p_{Y}(-y')\left(\frac{1}{p_{Y}(-y')}\int_{\R^{d}} 2\nabla_{y}f(\bz,y')P_{\bZ Y}(\diff\bz,\{-y'\})\right)^{2}\\
    &\le p_{Y}(+1)p_{Y}(-1)\sum_{y'\in\{\pm1\}}p_{Y}(-y')\frac{1}{p_{Y}(-y')}\int_{\R^{d}} 4(\nabla_{y}f(\bz,y'))^{2}P_{\bZ Y}(\diff\bz,\{-y'\})\\
    &=4p_{Y}(+1)p_{Y}(-1)\sum_{y\in\{\pm1\}}\int_{\R^{d}} (\nabla_{y}f(\bz,y'))^{2}P_{\bZ Y}(\diff\bz,\{y\})\\
    &=4p_{Y}(+1)p_{Y}(-1)\E\left[(\nabla_{y}f(\bZ,Y))^{2}\right],
\end{align*}
where Jensen's inequality for $P_{\bZ|Y}(\diff\bz|-y')=P_{\bZ Y}(\diff\bz,\{-y'\})/p_{Y}(-y')$ yields the inequality above.
Therefore, we obtain that
\begin{align*}
    \Var(f(\bZ,Y))&=\E[\Var(f(\bZ,Y)|Y)]+\Var(\E[f(\bZ,Y)|Y])\\
    &\le K_{\Poincare}\left(1+cK_{\chi^{2}}\right)\E\left[\Gamma_{\bZ}(f)(\bZ,Y)\right]+c^{\ast}K_{\textnormal{V}}\E\left[(\nabla_{y}f(\bZ,Y))^{2}\right]\\
    &\le \E\left[\Gamma_{\Poincare}(f)(\bZ,Y)\right].
\end{align*}

(Step 2: log-Sobolev inequality) Note the following decomposition:
\begin{equation*}
    \Ent(f^{2}(\bZ,Y))=\E\left[\Ent_{\bZ|Y}(f^{2}(\bZ,Y)|Y)\right]+\Ent_{Y}\left(\E_{\bZ|Y}[f^{2}(\bZ,Y)|Y]\right).
\end{equation*}
The first term can be bounded as
\begin{equation*}
        \E[\Ent_{\bZ|Y}(f^{2}(\bZ,Y)|Y)]\le \E\left[2K_{\LogSobolev}\Gamma_{\bZ}(f)(\bZ,Y)\right].
\end{equation*}
We analyse the second term on the right-hand side in detail.
We have
\begin{align*}
    &\Ent_{Y}\left(\E_{\bZ|Y}[f^{2}(\bZ,Y)|Y]\right)\\
    &=\E_{Y}\left[\E_{\bZ|Y}[f^{2}(\bZ,Y)|Y]\log\frac{ \E_{\bZ|Y}[f^{2}(\bZ,Y)|Y]}{\E_{\bZ Y}[f^{2}(\bZ,Y)]}\right]\\
    &\le \E_{Y}\left[\Ent_{\bZ|Y}(f^{2}(\bZ,Y)|Y)\right]+\E_{Y}\left[\E_{\bZ|Y}[f^{2}(\bZ,Y)|Y]\log\left(1+\left.\chi^{2}\left(P_{\bZ|Y}(\cdot|y)\delta_{y}(\cdot)\|P_{\bZ Y}(\cdot)\right)\right|_{y=Y}\right)\right]\\
    &=\E_{Y}\left[\Ent_{\bZ|Y}(f^{2}(\bZ,Y)|Y)\right]+\E_{Y}\left[\E_{\bZ|Y}[f^{2}(\bZ,Y)|Y]\log\frac{1}{p_{Y}(Y)}\right]\\
    &\le \E_{Y}\left[\Ent_{\bZ|Y}(f^{2}(\bZ,Y)|Y)\right]+\log K_{\textnormal{U}}\E[f^{2}],
\end{align*}
where the first inequality holds via Lemma \ref{lem:ccn21lem1} with fixing $\pi(\diff\bz',\diff y')=P_{\bZ|Y}(\diff\bz'|y)\delta_{y}(\diff y')$ and $\rho=P_{\bZ Y}(\diff\bz',\diff y')$ for each $y\in\{\pm1\}$, and the second identity is given as follows: for each $y\in\{\pm1\}$,
\begin{align*}
     1+\chi^{2}\left(P_{\bZ|Y}(\cdot|y)\delta_{y}(\cdot)\|P_{\bZ Y}(\cdot)\right)
    &=\int_{\R^{d}}\frac{P_{\bZ|Y}(\diff\bz'|y)}{P_{\bZ Y}(\diff \bz', \{y\})}P(\diff\bz'|y)\\
    &=\frac{1}{p_{Y}(y)^{2}}\int_{\R^{d}}\frac{P_{\bZ Y}(\diff \bz', \{y\})}{P_{\bZ Y}(\diff \bz', \{y\})}P_{\bZ Y}(\diff \bz', \{y\})\\
    &=\frac{1}{p_{Y}(y)}.
\end{align*}
In total, we have
\begin{equation*}
    \Ent(f^{2}(\bZ,Y))\le 2\E\left[\Ent_{\bZ|Y}(f^{2}(\bZ,Y)|Y)\right]+\log K_{\textnormal{U}}\E[f^{2}].
\end{equation*}
By letting $\hat{f}=f-\E f$ and owing to Rothaus' lemma \citep[Lemma 5.1.4 of][]{bakry2014analysis}, we obtain
\begin{equation*}
    \Ent(f^{2})\le \Ent(\hat{f}^{2})+2\Var(f)
\end{equation*}
Combining the bound on $\Ent(f^2(\bZ,Y))$ above with the Poincar\'{e} inequality which we also derive,
\begin{align*}
    \Ent(f^{2})&\le \Ent(\hat{f}^{2})+2\Var(f)\le 2\E\left[\Ent_{\bZ|Y}(\hat{f}^{2}|Y)\right]+\log K_{\textnormal{U}}\E[\hat{f}^{2}]+2\Var(f)\\
    &\le 2\left(\E\left[\Ent_{\bZ|Y}(\hat{f}^{2}|Y)\right]+\left(1+\frac{\log K_{\textnormal{U}}}{2}\right)\Var(f)\right)\\\\
    &\le 2\left(2K_{\LogSobolev}\E\left[\Gamma_{\bZ}(f)\right]+\left(1+\frac{\log K_{\textnormal{U}}}{2}\right)\E\left[\Gamma_{\Poincare}(f)\right]\right).
\end{align*}
This is the desired conclusion.
\end{proof}

\subsection{Proof of Lemma \ref{lem:tensor}}
Though it is identical to Theorems 2.3 and Theorem 3.14 of \citet{van2014probability}, we present the proof for self-containedness.

We prepare the variational representation of entropy.
\begin{lemma}[Lemma 3.15 of \citealp{van2014probability}]\label{lem:entropy}
    For nonnegative random variable $U$, we have
    \begin{equation*}
        \Ent(U)=\sup\left\{E[UV]: \text{$V$ is a random variable satisfying $\E[e^{V}]=1$}\right\}.
    \end{equation*}
\end{lemma}

\begin{proof}
    Fix a random variable $V$ with $\E[e^{V}]=1$ and define the new probability measure $Q$ with $Q(A)=\E[e^{V}\mathbbm{1}_{A}]$. Then we have
    \begin{align*}
        \Ent(U)-\E[UV]&=\E[U\log U]-\E[U\log e^{V}]-\E[U]\log\E[U]\\
        &=\E_{Q}[e^{-V}U\log(e^{-V}U)]-\E_{Q}[e^{-V}U]\log\E_{Q}[e^{-V}U].
    \end{align*}
    Since $x\mapsto x\log x$ is convex, Jensen's inequality leads to $\Ent(U)\ge \E[UV]$.
    Furthermore, we see that $V=V^\ast=\log(Z/\E[Z])$ achieves $\Ent(Y)=\E[UV]$. This is the desired conclusion.
\end{proof}
Lemma \ref{lem:tensor} is almost immediate by the following tensorization for more general spaces.
\begin{lemma}[tensorization;  Theorems 2.3 and 3.14 of \citealp{van2014probability}]\label{lem:tensor:general}
    Fix a measurable space $(\bbS,\calS)$. 
    For arbitrary independent random variables $\zeta_{i}$ for $i=1,\ldots,n$ defined on the measurable space $(\bbS,\calS)$ and measurable functions $f:\bbS^{n}\to\R$ and $g:\bbS^{n}\to[0,\infty)$,
    \begin{align*}
        \Var(f(\zeta_{1},\ldots,\zeta_{n}))&\le \E\left[\sum_{i=1}^{n}\Var\left(f(\zeta_{1},\ldots,\zeta_{n})|\zeta_{1},\ldots,\zeta_{i-1},\zeta_{i+1},\ldots,\zeta_{n}\right)\right],\\
        \Ent(g(\zeta_{1},\ldots,\zeta_{n}))&\le \E\left[\sum_{i=1}^{n}\Ent\left(g(\zeta_{1},\ldots,\zeta_{n})|\zeta_{1},\ldots,\zeta_{i-1},\zeta_{i+1},\ldots,\zeta_{n}\right)\right],
    \end{align*}
    where $\Ent(U|\calG)=\E[U\log U|\calG]-\E[U|\calG]\log\E[U|\calG]$ for a nonnegative random variable $U$ and sub-$\sigma$-algebra $\calG$.
\end{lemma}
    
\begin{proof}
    (Step 1: Variance)
    Define $T=f(\zeta_{1},\ldots,\zeta_{n})$ for brevity and prove $\Var(T)\le \E[\sum_{i=1}^{n}\Var(T|(\zeta_{j})_{j\neq i})]$.
    Let $\Delta_{i}$ for $i=1,\ldots,n$ be random variables defined as
    \begin{equation*}
        \Delta_{i}=
            \E\left[T|(\zeta_{j})_{j\le i}\right]-\E\left[T|(\zeta_{j})_{j\le i-1}\right],
    \end{equation*}
    where we set $\E[T|(\zeta_{j})_{j\le 0}]=\E[T]$ by convention.
    Then $T-\E[T]=\sum_{i=1}^{n}\Delta_{i}$ and $\E[\Delta_{i}\Delta_{j}]=0$ for $i\neq j$.
    Hence we have
    \begin{equation*}
        \Var(T)=\E\left[\left(\sum_{i=1}^{n}\Delta_{i}\right)^{2}\right]=\sum_{i=1}^{n}\E\left[\Delta_{i}^{2}\right].
    \end{equation*}
    It suffices to show $\E[\Delta_{i}^{2}]\le \E[\Var(T|(\zeta_{j})_{j\neq i})]$.
    Using the tower property of conditional expectation and the independence of $(\zeta_{i})_{i=1}^{n}$, we obtain
    \begin{equation*}
        \E\left[T|(\zeta_{j})_{j\le i-1}\right]=\E\left[\E\left[\left.T|(\zeta_{j})_{j\neq i}\right]\right|(\zeta_{j})_{j\le i-1}\right]
        =\E\left[\E\left[\left.T|(\zeta_{j})_{j\neq i}\right]\right|(\zeta_{j})_{j\le i}\right]
    \end{equation*}
    since the random variable $\E[T|(\zeta_{j})_{j\neq i}]$ is independent of $\zeta_{i}$.
    Note that $\Delta_{i}=\E[\tilde{\Delta}_{i}|(\zeta_{j})_{j\le i}]$, where $\tilde{\Delta}_{i}$ is defined as
    \begin{equation*}
        \tilde{\Delta}_{i}=T-\E\left[T|(\zeta_{j})_{j\neq i}\right]
    \end{equation*}
    Since $\Var(T|(\zeta_{j})_{j\neq i})=\E[\tilde{\Delta}_{i}^{2}|(\zeta_{j})_{j\neq i}]$ due to the independence of $(\zeta_{i})_{i=1}^{n}$, Jensen's inequality leads to
    \begin{equation*}
        \E\left[\Delta_{i}^{2}\right]=\E\left[\E[\tilde{\Delta}_{i}|(\zeta_{j})_{j\le i}]^{2}\right]\le \E\left[\tilde{\Delta}_{i}^{2}\right]=\E\left[\Var(T|(\zeta_{j})_{j\neq i})\right].
    \end{equation*}

    (Step 2: Entropy) Define $U=g(\zeta_{1},\ldots,\zeta_{n})$ and for $i=1,\ldots,n$,
    \begin{equation*}
        V_{i}=
            \log\E\left[U|(\zeta_{j})_{j\le i}\right]-\log\E\left[U|(\zeta_{j})_{j\le i-1}\right],
    \end{equation*}
    where we let $\E[U|(\zeta_{j})_{j\le 0}]=\E[U]$ by convention.
    Then, we see that
    \begin{equation*}
        \Ent\left(U\right)=\E\left[U\left(\log U - \log\E\left[U\right]\right)\right]
        =\sum_{i=1}^{n}\E\left[UV_{i}\right].
    \end{equation*}
    The independence of $(\zeta_{i})_{i=1}^{n}$ yields
    \begin{align*}
        \E\left[e^{V_{i}}|(\zeta_{j})_{j\neq i}\right]&=\frac{\E\left[\E\left[U|(\zeta_{j})_{j\le i}\right]|(\zeta_{j})_{j\neq i}\right]}{\E\left[U|(\zeta_{j})_{j\le  i-1}\right]}=\frac{\E\left[\E\left[U|(\zeta_{j})_{j\le i}\right]|(\zeta_{j})_{j\le i-1}\right]}{\E\left[U|(\zeta_{j})_{j\le  i-1}\right]}=1.
    \end{align*}
    Therefore, Lemma \ref{lem:entropy} yields
    \begin{equation*}
        \E\left[UV_{i}|(\zeta_{j})_{j\neq i}\right]\le \Ent\left(U|(\zeta_{j})_{j\neq i}\right),
    \end{equation*}
    which is the desired conclusion.
\end{proof}

\begin{proof}[Proof of Lemma \ref{lem:tensor}]
We derive the conclusion by setting $\bbS=\R^{d}\times\{\pm1\}$ and applying Lemma \ref{lem:tensor:general}; in particular, as the proof of Theorem \ref{thm:functional}, Poincar\'{e} inequalities and log-Sobolev inequalities for each $\zeta_{i}=(\bZ_{i},Y_{i})$ are applicable to $\Var(f(\zeta_{1},\ldots,\zeta_{n})|\zeta_{1},\ldots,\zeta_{i-1},\zeta_{i+1},\ldots,\zeta_{n})$ and $\Ent\left(g(\zeta_{1},\ldots,\zeta_{n})|\zeta_{1},\ldots,\zeta_{i-1},\zeta_{i+1},\ldots,\zeta_{n}\right)$.
\end{proof}

\subsection{Proofs of Lemmas \ref{lem:bobkov} and \ref{lem:herbst}}

We introduce subexponential concentration and subgaussian concentration for Lipschitz maps under a Poincar\'{e} inequality and log-Sobolev inequality respectively.

We employ the following known inequality: for all $a,b\in\R$,
\begin{align}
    \left(e^{a/2}-e^{b/2}\right)^{2}&=\left(\frac{a-b}{2}\int_{0}^{1}e^{au/2+b(1-u)/2}\diff u\right)^{2}
    \le \frac{(a-b)^2}{4}\int_{0}^{1}e^{au+b(1-u)}\diff u\notag\\
    &\le \frac{(a-b)^2}{4}\int_{0}^{1}(ue^{a}+(1-u)e^{b})\diff u\le \frac{\left(a-b\right)^{2}}{4}\frac{e^{a}+e^{b}}{2}.\label{eq:dchain}
\end{align}
We also often use a truncation argument $f_{k}=\min\{\max\{f,-k\},k\}$ to guarantee integrability and Fatou's lemma; note that $\Gamma_{\bZ_{i}}(f_{k})\le \Gamma_{\bZ_{i}}(f)$ and $\Gamma_{Y_{i}}(f_{k})\le \Gamma_{Y_{i}}(f)$.

\begin{proof}[Proof of Lemma \ref{lem:bobkov}]
    (Step 1: Integrability) We first show that $\E_{P}[f]<\infty$ for unbounded and Lipschitz $f$. Let $f_{k}=\min\{|f|,k\}$, since $\Gamma_{\bZ_{i}}(f_k)\le \Gamma_{\bZ_{i}}(|f|)\le \Gamma_{\bZ_{i}}(f)\le L_{\bZ}^{2}$ and $\Gamma_{Y_{i}}(f_k)\le \Gamma_{Y_{i}}(|f|)\le \Gamma_{Y_{i}}(f)\le L_{Y}^{2}$, 
    \begin{equation*}
        \E_{P}[(f_{k}-\E_{P}[f_{k}])^2]\le C,
    \end{equation*}
    and thus, Markov's inequality yields that for all $k\ge1$,
    \begin{equation*}
        P\left(|f_{k}-\E_{P}[f_{k}]|\ge \sqrt{2C}\right)\le \frac{1}{2}.
    \end{equation*}
    Fix $D>0$ such that $P(|f|\ge D)<1/2$.
    It implies that $P(f_k\ge D)<1/2$ for any $k\ge1$. 
    Therefore, $P(\{|f_{k}-\E_{P}[f_{k}]|<\sqrt{2C}\}\cap\{f_{k}< D\})>0$, and thus, for some $((\bz_{i},y_{i}))_{i=1}^{n}$, $|f_{k}(((\bz_{i},y_{i}))_{i=1}^{n})-\E_{P}[f_{k}]|\le \sqrt{2C}$ and $f_{k}(((\bz_{i},y_{i}))_{i=1}^{n})<D$.
    It indicates that
    \begin{equation*}
        \E_{P}[f_{k}]\le \sqrt{2C}+D,
    \end{equation*}
    and thus Fatou's lemma yields $\E_{P}[|f|]\le \lim_{k\to\infty}\E_{P}[f_{k}]\le \sqrt{2C}+D$.

    Therefore, we have $\E_{P}[f]=\lim_{k\to\infty}\E_{P}[\min\{\max\{f,-k\},k\}]$ by this result and Lebesgue's donminated convergence theorem.
    By this convergence of expectations, the truncation argument, and Fatou's lemma applied for the exponential moment below, we can suppose that $f$ is bounded without loss of generality.

    (Step 2: Exponential integrability) Set $C=C_{\bZ}L_{\bZ}^{2}+2C_{Y} L_{Y}^{2}$ for brevity.
    We show that for all $s< \min\{\log3/L_{Y},2/\sqrt{nC}\}$,
    \begin{equation*}
        \E_{P}[e^{sf}]\le e^{s\E_{P}[f]}\prod_{\ell=0}^{\infty}\left(1-\frac{nCs^2}{4^{\ell+1}}\right)^{-2^\ell}.
    \end{equation*}
    The Poincar\'{e} inequality in the statement and Eq.~\eqref{eq:dchain} give that
    \begin{align*}
        \Var_{P}(e^{sf/2})&\le \E_{P}\left[C_{\bZ}\sum_{i=1}^{n}\Gamma_{\bZ_{i}}(f)+C_{Y}\sum_{i=1}^{n}\Gamma_{Y_{i}}(f)\right]\le s^2\E_{P}\left[\frac{C_{\bZ}}{4}e^{sf}\sum_{i=1}^{n}\Gamma_{\bZ_{i}}(f)+\frac{C_{Y}}{4}\sum_{i=1}^{n}\frac{e^{sf}+e^{sT_if}}{2}\Gamma_{Y_i}(f)\right]\\
        &\le  \frac{n s^2}{4}\left(C_{\bZ}L_{\bZ}^{2}+C_{Y} L_{Y}^{2}\frac{1+e^{sL_{Y}}}{2}\right)\E_{P}\left[e^{sf}\right],
    \end{align*}
    where $T_if((\bz_1,y_1),\ldots,(\bz_{i},y_{i}),\ldots,(\bz_{n},y_{n}))=f((\bz_1,y_1),\ldots,(\bz_{i},-y_{i}),\ldots,(\bz_{n},y_{n}))$ and we used $e^{sT_{i}f}\le e^{sf+2sL_{Y}}$.
    Since $e^{2sL_{Y}}\le 3$ for all $s\in[0,\log 3/(2L_{Y})]$, we derive that for all $s< \min\{\log3/(2L_{Y}),2/\sqrt{nC}\}$
    \begin{equation*}
        \E_{P}[e^{sf}]-\E_{P}[e^{sf/2}]^{2}\le \frac{n s^2}{4}\left(C_{\bZ}L_{\bZ}^{2}+2C_{Y} L_{Y}^{2}\right)\E_{P}\left[e^{sf}\right]=\frac{nCs^2}{4}\E_{P}\left[e^{sf}\right].
    \end{equation*}
    Setting $F(s)=\E_{P}[e^{sf}]$ for $s< \min\{\log3/(2L_{Y}),2/\sqrt{nC}\}$, we have that
    \begin{equation*}
        \left(1-\frac{nCs^2}{4}\right)F(s)\le F\left(\frac{s}{2}\right)^{2}.
    \end{equation*}
    Since $1-\frac{nCs^2}{4}>0$, 
    \begin{equation*}
        F(s)\le  \left(1-\frac{nCs^2}{4}\right)^{-1}F\left(\frac{s}{2}\right)^{2}.
    \end{equation*}
    Repeating this operation and taking the limit, we derive that
    \begin{align*}
        F(s)&\le \lim_{L\to\infty} F\left(\frac{s}{2^{L+1}}\right)^{2^{L+1}}\prod_{\ell=0}^{\infty}\left(1-\frac{nCs^2}{4^{\ell+1}}\right)^{-2^{\ell}}=\lim_{m\to\infty}\E_{P}\left[e^{(sf)/m}\right]^{m}\prod_{\ell=0}^{\infty}\left(1-\frac{nCs^2}{4^{\ell+1}}\right)^{-2^{\ell}}\\
        &=e^{s\E_{P}[f]}\prod_{\ell=0}^{\infty}\left(1-\frac{nCs^2}{4^{\ell+1}}\right)^{-2^{\ell}}
    \end{align*}
    as $\log (\lim_{m\to\infty}\E_{P}[e^{(sf)/m}]^{m})=\lim_{h\to0}h^{-1}\log(\E_{P}[e^{hsf}]-\log e^{0})=\E_{P}[sf]$.

    (Step 3: Tail bound) Since $-\log(1-u)\le u/(1-u)$ for all $u\in(0,1)$, we have that for all $x\in(0,1)$, 
    \begin{align*}
        &\log\prod_{\ell=0}^{\infty}\left(1-\frac{x}{4^{\ell}}\right)^{-2^\ell}=\sum_{\ell=0}^{\infty}2^{\ell}\left[-\log\left(1-\frac{x}{4^{\ell}}\right)\right]\le \sum_{\ell=0}^{\infty}2^{\ell}\frac{\frac{x}{4^{\ell}}}{1-\frac{x}{4^{\ell}}}\le \sum_{\ell=0}^{\infty}\frac{\frac{x}{2^{\ell}}}{1-x}\le \frac{2x}{1-x}.
    \end{align*}
    For every $s<\min\{\log 3/(2L_{Y}), 2/\sqrt{nC}\}$, we set $x=nCs^{2}/4$ in the above inequality and have
    \begin{equation*}
        \E_{P}[e^{sf-s\E_{P}[f]}]\le \prod_{\ell=0}^{\infty}\left(1-\frac{nCs^2}{4^{\ell+1}}\right)^{-2^\ell}\le \exp\left(\frac{2nCs^2}{4-nCs^2}\right).
    \end{equation*}
    Fixing $s=\min\{1/(2L_{Y}), 1/\sqrt{nC}\}$ and employing Chernoff's bound, we obtain that
    \begin{equation*}
        P\left(f-\E_{P}[f]\ge t\right)\le \exp\left(\frac{2}{4-1}\right)\exp(-st)\le2\exp\left(-t\min\left\{\frac{1}{2L_{Y}},\frac{1}{\sqrt{nC}}\right\}\right),
    \end{equation*}
    where we used $e^{2/3}\le 2$.
    This is the desired conclusion.
\end{proof}

\begin{proof}[Proof of Lemma \ref{lem:herbst}]
    By the truncation argument, suppose that $f$ is bounded without loss of generality.
    Using \eqref{eq:dchain}, we have that for all $s\ge0$,
    \begin{align*}
        \Ent_{P}(e^{sf})&=\Ent_{P}\left((e^{(sf/2)})^2\right)\le 2s^2\E_{P}\left[\frac{C_{\bZ}}{4}e^{sf}\sum_{i=1}^{n}\Gamma_{\bZ_{i}}(f)+\frac{C_{Y}}{4}\sum_{i=1}^{n}\frac{e^{sf}+e^{sT_if}}{2}\Gamma_{Y_i}(f)\right]\\
        &\le  \frac{n s^2}{2}\left(C_{\bZ}L_{\bZ}^{2}+C_{Y} L_{Y}^{2}\frac{1+e^{2sL_{Y}}}{2}\right)\E_{P}\left[e^{sf}\right],
    \end{align*}
    where $T_if((\bz_1,y_1),\ldots,(\bz_{i},y_{i}),\ldots,(\bz_{n},y_{n}))=f((\bz_1,y_1),\ldots,(\bz_{i},-y_{i}),\ldots,(\bz_{n},y_{n}))$.
    Hence, we obtain that for $M(s)=\E[e^{sf}]$ for all $s\ge0$,
    \begin{equation*}
        sM'(s)\le M(s)\log M(s)+\frac{n s^2}{2}\left(C_{\bZ}L_{\bZ}^{2}+C_{Y} L_{Y}^{2}\frac{1+e^{2sL_{Y}}}{2}\right)M(s).
    \end{equation*}
    Define $F(s)=(1/s)\log M(s)$ with $s>0$ and $F(0)=\E_{P}[f]$ (an unbounded Lipschitz $f$ is integrable as a log-Sobolev inequality implies a Poincar\'{e} inequality and Step 1 of the proof of Lemma \ref{lem:bobkov} yields it); then $F$ is continuous at $s=0$ and obtain
    \begin{equation*}
        F'(s)=\frac{1}{s}\frac{M'(s)}{M(s)}-\frac{1}{s^2}\log M(s)\le \frac{n}{2}\left(C_{\bZ}L_{\bZ}^{2}+C_{Y} L_{Y}^{2}\frac{1+e^{2sL_{Y}}}{2}\right).
    \end{equation*}
    Therefore, we have that for all $s>0$,
    \begin{align*}
        F(s)-F(0)&\le \int_{0}^{s}\frac{n}{2}\left(C_{\bZ}L_{\bZ}^{2}+C_{Y} L_{Y}^{2}\left(\frac{1+e^{2uL_{Y}}}{2}\right)\right)\diff u\\
        &= \frac{ns}{2}\left(C_{\bZ}L_{\bZ}^{2}+C_{Y} L_{Y}^{2}\left(\frac{1+\frac{1}{2sL_{Y}}(e^{2sL_{Y}}-1)}{2}\right)\right).
    \end{align*}
    It leads to
    \begin{equation*}
        \E_{P}\left[e^{s(f-\E_{P}[f])}\right]\le \exp\left(\frac{ns^{2}}{2}\left(C_{\bZ}L_{\bZ}^{2}+C_{Y} L_{Y}^{2}\left(\frac{1+\frac{1}{2sL_{Y}}(e^{2sL_{Y}}-1)}{2}\right)\right)\right).
    \end{equation*}
    Chernoff's bound derives that for all $t\ge0$ and $s>0$,
    \begin{equation*}
        P\left(f-\E_{P}[f]\ge t\right)\le \exp\left(-st+\frac{ns^{2}}{2}\left(C_{\bZ}L_{\bZ}^{2}+C_{Y} L_{Y}^{2}\left(\frac{1+\frac{1}{2sL_{Y}}(e^{2sL_{Y}}-1)}{2}\right)\right)\right).
    \end{equation*}
    We constrain $s\le 1/(2L_{Y})$, so that $2sL_{Y}\le 1$. Using $(e^x-1)\le (e-1)x$ for all $x\in[0,1]$, we obtain
    \begin{equation*}
        P\left(f-\E_{P}[f]\ge t\right)\le \exp\left(-st+\frac{ns^{2}}{2}\left(C_{\bZ}L_{\bZ}^{2}+\frac{e}{2}C_{Y}L_{Y}^{2}\right)\right).
    \end{equation*}
    If $t\le \frac{n(C_{\bZ}L_{\bZ}^{2}+(e/2)C_{Y}L_{Y}^{2})}{2L_{Y}}$, we can choose $s=\frac{t}{n(C_{\bZ}L_{\bZ}^{2}+\frac{e}{2}C_{Y}L_{Y}^{2})}$ bounded above by $1/(2L_{Y})$ and obtain
    \begin{equation*}
        P\left(f-\E_{P}[f]\ge t\right)\le \exp\left(-\frac{t^{2}}{n\left(2C_{\bZ}L_{\bZ}^{2}+eC_{Y}L_{Y}^{2}\right)}\right).
    \end{equation*}
    Otherwise (i.e., if $t>\frac{n(C_{\bZ}L_{\bZ}^{2}+\frac{e}{2}C_{Y}L_{Y}^{2})}{2L_{Y}}$), we choose $s=1/2L_{Y}$ and have that 
    \begin{equation*}
        P\left(f-\E_{P}[f]\ge t\right)\le \exp\left(-\frac{t}{2L_{Y}}+\frac{n}{8L_{Y}^{2}}\left(C_{\bZ}L_{\bZ}^{2}+\frac{e}{2}C_{Y}L_{Y}^{2}\right)\right)\le \exp\left(-\frac{t}{4L_{Y}}\right).
    \end{equation*}
    This is the desired result.
\end{proof}

\section{Proofs of the results in Section \ref{sec:concentration}}\label{sec:proof:concentration}

\subsection{Proof of the result in Section \ref{sec:concentration:settings}}
Though the proof of Proposition \ref{prop:rademacher} is a very standard one, we present it for readers' convenience.

\begin{proof}[Proof of Proposition \ref{prop:rademacher}]
    Proposition 4.2 of \citet{bach2024learning} (symmetrization) gives
    \begin{equation*}
        \E\left[\sup_{(\bw,b)\in\calT}(\Risk(\bw,b)-\Risk_{n}(\bw,b))\right]\le 2\E\left[\sup_{(\bw,b)\in\calT}\frac{1}{n}\sum_{i=1}^{n}\varepsilon_{i}\ell(Y_{i}(\langle\bX_{i},\bw\rangle+b))\right],
    \end{equation*}
    where $\varepsilon_{i}\sim^{\textnormal{i.i.d.}}\Unif(\{\pm1\})$ independent of $\{\bX_{i},Y_{i}\}_{i=1}^{n}$.
    Proposition 4.3 of \citet{bach2024learning} (contraction) yields
    \begin{equation*}
        \E\left[\sup_{(\bw,b)\in\calT}\frac{1}{n}\sum_{i=1}^{n}\varepsilon_{i}\ell(Y_{i}(\langle\bX_{i},\bw\rangle+b))\right]\le L\E\left[\sup_{(\bw,b)\in\calT}\frac{1}{n}\sum_{i=1}^{n}\varepsilon_{i}(Y_{i}(\langle\bX_{i},\bw\rangle+b))\right].
    \end{equation*}
    Since
    \begin{align*}
        \E\left[\sup_{(\bw,b)\in\calT}\frac{1}{n}\sum_{i=1}^{n}\varepsilon_{i}(Y_{i}(\langle\bX_{i},\bw\rangle+b))\right]&=\frac{1}{n}\E\left[R_{\bw}\left\|\sum_{i=1}^{n}\varepsilon_{i}Y_{i}\bX_{i}\right\|+R_{b}\left|\sum_{i=1}^{n}\varepsilon_{i}Y_{i}\right|\right]\\
        &= \frac{1}{n}\left(R_{\bw}\E\left[\left\|\sum_{i=1}^{n}\varepsilon_{i}Y_{i}\bX_{i}\right\|\right]+R_{b}\E\left[\left|\sum_{i=1}^{n}\varepsilon_{i}Y_{i}\right|\right]\right)\\
        &\le  \frac{1}{n}\left(R_{\bw}\E\left[\left\|\sum_{i=1}^{n}\varepsilon_{i}Y_{i}\bX_{i}\right\|^{2}\right]^{1/2}+R_{b}\E\left[\left(\sum_{i=1}^{n}\varepsilon_{i}Y_{i}\right)^{2}\right]^{1/2}\right)\\
        &=  \frac{1}{n}\left(R_{\bw}\E\left[\sum_{i=1}^{n}\left\|Y_{i}\bX_{i}\right\|^{2}\right]^{1/2}+R_{b}\E\left[\sum_{i=1}^{n}Y_{i}^{2}\right]^{1/2}\right)\\
        &=  \frac{1}{n}\left(R_{\bw}\sqrt{n}\E\left[\left\|\bX_{i}\right\|^{2}\right]^{1/2}+R_{b}\sqrt{n}\right),
    \end{align*}
    we obtain the desired result.
\end{proof}

\subsection{Proofs of the results in Section \ref{sec:concentration:base}}

\begin{proof}[Proof of Proposition \ref{prop:independence}]
        
    Let $P_{\bX|Y}(\diff \bx|y)$ be the conditional measure of $\bX_{i}$ given $Y_{i}=y\in\{\pm 1\}$. 
    For each $y\in\{\pm1\}$,
    \begin{equation*}
        P_{\bX|Y}(\diff \bx|y)=\frac{\exp\left(-U\left(\bx\right)-\left(-\log(g(y\langle \bx,\btheta_{1}\rangle\right)\right)\diff\bx}{\int_{\R^{d}}\exp\left(-U\left(\bx\right)-\left(-\log(g(y\langle \bx,\btheta_{1}\rangle\right)\right)\diff \bx}.
    \end{equation*}
    We consider $P_{\bZ|Y}$, the conditional measure of $\bZ_{i}$ given $Y_{i}=y\in\{\pm1\}$.
    We have
    \begin{equation*}
        P_{\bZ|Y}(\diff \bz|y)=\frac{\exp\left(-U\left(y\bz\right)-\left(-\log(g(\langle \bz,\btheta_{1}\rangle\right)\right)\diff\bz}{\int_{\R^{d}}\exp\left(-U\left(y\bz\right)-\left(-\log(g(\langle \bz,\btheta_{1}\rangle\right)\right)\diff \bz}=\frac{\exp\left(-U\left(\bz\right)-\left(-\log(g(\langle \bz,\btheta_{1}\rangle\right)\right)\diff\bz}{\int_{\R^{d}}\exp\left(-U\left(\bz\right)-\left(-\log(g(\langle \bz,\btheta_{1}\rangle\right)\right)\diff \bz},
    \end{equation*}
    where the evenness of $U$ gives the second identity.
    Hence, $\bZ_{i}$ is independent of $Y_{i}$.
\end{proof}

\begin{proof}[Proof of Proposition \ref{prop:concentration:0}]
    We only prove the statement (ii) since (i) can be shown by Lemma \ref{lem:bobkov} and a parallel manner.
    
    The conditional distribution of $\bZ_{i}$ given $Y_{i}$ is equal to the marginal distribution of $\bZ_{i}$ due to Proposition \ref{prop:independence}, and $Y_{i}\sim \Unif(\{\pm1\})$ holds.
    Tensorization of log-Sobolev inequalities (Lemma \ref{lem:tensor}) and Assumption \ref{assum:lsi} yield the following log-Sobolev inequality: for all $f\in\calA_{n}$,
    \begin{equation*}
        \Ent(f^{2})\le 2\E\left[\sum_{i=1}^{n}\left(K_{\LogSobolev}\Gamma_{\bZ_{i}}(f)+\Gamma_{Y_{i}}(f)\right)\right]=:2\E[\Gamma_{\textnormal{i}}(f)].
    \end{equation*}
    What we need is an almost sure upper bound on $\Gamma_{\textnormal{i}}$, owing to Lemma \ref{lem:herbst}.
    We easily see that for arbitrary $\bz_{i},\bz_{i}'\in\R^{d}$ and $y,y'\in\{\pm1\}$,
    \begin{align*}
        \sup_{(\bw,b)\in\calT}\left(\Risk(\bw,b)-\ell(\langle\bz_{i},\bw\rangle+y_{i}b)\right)-\sup_{(\bw,b)\in\calT}\left(\Risk(\bw,b)-\ell(\langle\bz_{i}',\bw\rangle+y_{i}b)\right)&\le LR_{\bw}\|\bz-\bz'\|,\\
        \sup_{(\bw,b)\in\calT}\left(\Risk(\bw,b)-\ell(\langle\bz_{i},\bw\rangle+y_{i}b)\right)-\sup_{(\bw,b)\in\calT}\left(\Risk(\bw,b)-\ell(\langle\bz_{i},\bw\rangle+y_{i}'b)\right)&\le 2LR_{b},
    \end{align*}
    and
    \begin{align}
        \Gamma_{\bZ_{i}}\left(\sup_{(\bw,b)\in\calT}\left(\Risk(\bw,b)-\frac{1}{n}\sum_{i=1}^{n}\ell(\langle\bz_{i},\bw\rangle+y_{i}b)\right)\right)
        &\le \frac{L^{2}R_{\bw}^{2}}{n^{2}},\label{eq:cdc:z}\\
        \Gamma_{Y_{i}}\left(\sup_{(\bw,b)\in\calT}\left(\Risk(\bw,b)-\frac{1}{n}\sum_{i=1}^{n}\ell(\langle\bz_{i},\bw\rangle+y_{i}b)\right)\right)
        &\le \frac{L^{2}R_{b}^{2}}{n^{2}}.\label{eq:cdc:y}
    \end{align}
    Hence, we obtain the desired conclusion by Lemma \ref{lem:herbst}.
\end{proof}

\subsection{Proofs of the results in Section \ref{sec:concentration:main}}
Let us define the normalizing constants of conditional measures as follows: 
\begin{equation*}
    \calZ_{+\theta_{0}}:=\int_{\R^{d}}g(\langle \bz,\btheta_{1}\rangle+\theta_{0})\exp\left(-U\left(\bz\right)\right)\diff \bz,\ \calZ_{-\theta_{0}}:=\int_{\R^{d}}g(\langle \bz,\btheta_{1}\rangle-\theta_{0})\exp\left(-U\left(\bz\right)\right)\diff \bz.
\end{equation*}
Note that $\calZ_{+\theta_{0}}+\calZ_{-\theta_{0}}=\calZ$ under the evenness of the potential function $U$.
\begin{proof}[Proof of Proposition \ref{prop:concentration:1}]
    We only prove (ii) since the proof of (i) is analogous and more straightforward (we use Theorem \ref{thm:functional} with $c=1+1/\sqrt{e^{2G|\theta_{0}|}-1}$ and $c^{\ast}=1+\sqrt{e^{2G|\theta_{0}|}-1}$).
    
    We use that $K_{\Poincare}\le K_{\LogSobolev}$.
    Therefore, we need estimates for the following constants:
    \begin{align*}
        K_{\chi^{2}}&:=\max_{y\in\{\pm1\}}\chi^{2}\left(P_{\bZ}(\cdot)\|P_{\bZ|Y}(\cdot|y)\right),\\
        K_{\textnormal{V}}&:=4p_{Y}(+1)p_{Y}(-1),\\
        K_{\textnormal{U}}&:=\max\left\{\frac{1}{p_{Y}(+1)},\frac{1}{p_{Y}(-1)}\right\}.
    \end{align*}

The assumption yields that for all $\theta_{0}\in\R$ and $t\in\R$,
\begin{equation}\label{eq:loglikdiff}
    \left|\log(g(t+\theta_{0}))-\log(g(t-\theta_{0}))\right|\le 2G|\theta_{0}|.
\end{equation}
Let us assume $\theta_{0}\ge0$ without loss of generality.
Equation \eqref{eq:loglikdiff} gives $g(t+\theta_{0})/g(t-\theta_{0})\le e^{2G\theta_{0}}$, which is frequently used in the subsequent discussions.
Note that $\calZ_{+\theta_{0}}/\calZ_{-\theta_{0}}\in[1,e^{2G\theta_{0}}]$ for any $\theta_{0}\ge0$ since the inequality \eqref{eq:loglikdiff} yields
\begin{align*}
    \calZ_{+\theta_{0}}
    &=\int_{\R^{d}}\exp\left(-U\left(\bz\right)+\log(g(\langle \bz,\btheta_{1}\rangle+\theta_{0}))\right)\diff \bz\\
    &= \int_{\R^{d}}\exp\left(-U\left(\bz\right)+\log(g(\langle \bz,\btheta_{1}\rangle+\theta_{0}))-\log(g(\langle \bz,\btheta_{1}\rangle-\theta_{0}))+\log(g(\langle \bz,\btheta_{1}\rangle-\theta_{0}))\right)\diff \bz\\
    &\le \int_{\R^{d}}\exp\left(-U\left(\bz\right)+2G\theta_{0}+\log(g(\langle \bz,\btheta_{1}\rangle-\theta_{0}))\right)\diff \bz\\
    &= e^{2G\theta_{0}}\calZ_{-\theta_{0}},
\end{align*}
and the monotonicty of $g$ gives $\calZ_{+\theta_{0}}\ge \calZ_{-\theta_{0}}$.
Similarly, $\calZ_{-\theta_{0}}/\calZ_{+\theta_{0}}\in[e^{-2G\theta_{0}},1]$ holds.

Since $g(t-\theta_{0})/g(t+\theta_{0})\le 1$, we obtain
\begin{align*}
    \chi^{2}\left(P_{\bZ}(\cdot)\|P_{\bZ|Y}(\cdot|+1)\right)
    &=\int_{\R^{d}}\frac{\calZ_{+\theta_{0}}\sum_{y'\in\{\pm1\}}g(\langle \bz,\btheta_{1}\rangle+y'\theta_{0})}{\calZ g(\langle \bz,\btheta_{1}\rangle+\theta_{0})}P_{\bZ}\left(\diff \bz\right)-1\\
    &\le 2\frac{\calZ_{+\theta_{0}}}{\calZ}-1
    =2\frac{\calZ_{+\theta_{0}}}{\calZ_{+\theta_{0}}+\calZ_{-\theta_{0}}}-1
    =2\frac{\calZ_{+\theta_{0}}/\calZ_{-\theta_{0}}}{1+\calZ_{+\theta_{0}}/\calZ_{-\theta_{0}}}-1\\
    &\le \calZ_{+\theta_{0}}/\calZ_{-\theta_{0}}-1
    \le e^{2G\theta_{0}}-1.
\end{align*}
In addition, the fact that $\calZ_{-\theta}/\calZ\le 1/2$ along with $g(t+\theta_{0})/g(t-\theta_{0})\le e^{2G\theta_{0}}$ gives  
\begin{align*}
    \chi^{2}\left(P_{\bZ}(\cdot)\|P_{\bZ|Y}(\cdot|-1)\right)
    &=\int_{\R^{d}}\frac{\calZ_{-\theta_{0}}\sum_{y'\in\{\pm1\}}g(\langle \bz,\btheta_{1}\rangle+y'\theta_{0})}{\calZ g(\langle \bz,\btheta_{1}\rangle-\theta_{0})}P_{\bZ}\left(\diff \bz\right)-1\\
    &\le \frac{\calZ_{-\theta_{0}}}{\calZ}\int_{\R^{d}}\left(1+\frac{g(\langle \bz,\btheta_{1}\rangle+\theta_{0})}{g(\langle \bz,\btheta_{1}\rangle-\theta_{0})}\right)P_{\bZ}\left(\diff \bz\right)-1\\
    &= \frac{1}{2}(1+e^{2G\theta_{0}})-1\le e^{2G\theta_{0}}-1.
\end{align*}
Therefore, we obtain
\begin{equation*}
    K_{\chi^{2}}=\max_{y\in\{\pm1\}}\chi^{2}\left(P_{\bZ}(\cdot)\|P_{\bZ|Y}(\cdot|y)\right)\le e^{2G\theta_{0}}-1.
\end{equation*}
The remaining constants can be evaluated as follows:
\begin{align*}
    K_{\textnormal{V}}=\frac{4\calZ_{+\theta_{0}}\calZ_{-\theta_{0}}}{(\calZ_{+\theta_{0}}+\calZ_{-\theta_{0}})^{2}}\le 1,
    K_{U}=\frac{\calZ_{+\theta_{0}}+\calZ_{-\theta_{0}}}{\calZ_{-\theta_{0}}}\le 1+e^{2G\theta_{0}}.
\end{align*}

Hence, we obtain the following log-Sobolev inequality by Theorem \ref{thm:functional} ($c=c^{\ast}=2$) and Lemma \ref{lem:tensor}: for all $f\in\calA_{n}$,
\begin{equation*}
    \Ent(f^{2}((\bZ_{i},Y_{i})_{i=1}^{n}))\le 2\E[\Gamma_{\textnormal{ii}}(f)((\bZ_{i},Y_{i})_{i=1}^{n})],
\end{equation*}
where
\begin{align*}
    \Gamma_{\textnormal{ii}}(f)&:= \left(\left(1+\frac{1}{2}\log\left( 1+e^{2G\theta_{0}}\right)\right)K_{\LogSobolev}\left(1+2\left(e^{2G\theta_{0}}-1\right)\right)+2K_{\LogSobolev}\right)\sum_{i=1}^{n}\Gamma_{\bZ_{i}}(f)\\
    &\quad+2\left(1+\frac{1}{2}\log\left( 1+e^{2G\theta_{0}}\right)\right)\sum_{i=1}^{n}\Gamma_{Y_{i}}(f)\\
    &\le K_{\LogSobolev}\left(3+2G\theta_{0}\right)\left(\frac{1}{2}+e^{2G\theta_{0}}\right)\sum_{i=1}^{n}\Gamma_{\bZ_{i}}(f)+\left(3+2G\theta_{0}\right)\sum_{i=1}^{n}\Gamma_{Y_{i}}(f);
\end{align*}
here, we use that $2+\log(1+e^{2G\theta_{0}})\le 3+2G\theta_{0}$, and
\begin{align*}
    \left(1+\frac{1}{2}\log\left( 1+e^{2G\theta_{0}}\right)\right)\left(1+2\left(e^{2G\theta_{0}}-1\right)\right)+2\le \left(1+\frac{1}{2}\log\left( 1+e^{2G\theta_{0}}\right)\right)\left(1+2e^{2G\theta_{0}}\right).
\end{align*}
Therefore, using Equations \eqref{eq:cdc:z} and \eqref{eq:cdc:y} and Lemma \ref{lem:herbst}, we obtain that for all $\delta\in(0,1]$, with probability at least $1-\delta$,
    \begin{align*}
        &\sup_{(\bw,b)\in\calT}(\Risk(\bw,b)-\Risk_{n}(\bw,b))
        -\E\left[\sup_{(\bw,b)\in\calT}(\Risk(\bw,b)-\Risk_{n}(\bw,b))\right]\\
        &\le \max\left\{\sqrt{\frac{L^{2}\left(3+2G\theta_{0}\right)\left[K_{\LogSobolev}\left(1+2e^{2G\theta_{0}}\right)R_{\bw}^{2}+eR_{b}^{2}\right]\log(\delta^{-1})}{n}},\frac{4LR_{b}\log(\delta^{-1})}{n}\right\}.
    \end{align*}
    Hence, we obtain the desired conclusion.
\end{proof}

\begin{proof}[Proof of Proposition \ref{prop:concentration:2}]
    We only prove (ii) since we can prove (i) in a similar and simpler manner.
    
    Again, we consider $\theta_{0}>0$ and focus on estimate for $K_{\chi^{2}},K_{\textnormal{V}},K_{\textnormal{U}}$.
    We obtain the following bound for $\calZ_{-\theta_{0}}$:
    \begin{align*}
        \calZ_{-\theta_{0}}&\le \int_{\R^{d}}\exp(\langle \bz,\btheta_{1}\rangle-\theta_{0})\exp\left(-U(\bz)\right)\diff\bz=e^{-\theta_{0}}\calZ M_{\bX}(\btheta_{1}),\\
        \calZ_{-\theta_{0}}&=\int_{\R^{d}}\frac{1}{1+\exp(-\langle \bz,\btheta_{1}\rangle+\theta_{0})}\exp\left(-U(\bz)\right)\diff\bz\\
        &\ge \int_{\R^{d}}\frac{1}{e^{\theta_{0}}+\exp(-\langle \bz,\btheta_{1}\rangle+\theta_{0})}\exp\left(-U(\bz)\right)\diff\bz\\
        &\ge e^{-\theta_{0}}\int_{\R^{d}}\frac{1}{1+\exp(-\langle \bz,\btheta_{1}\rangle)}\exp\left(-U(\bz)\right)\diff\bz=e^{-\theta_{0}}\frac{\calZ}{2},
    \end{align*}
    where we use $1/(1+\exp(-t))\le \exp(t)$ for the upper bound.
    Since $\calZ_{+\theta_{0}}\le \calZ$,
    \begin{align*}
    \chi^{2}\left(P_{\bZ}(\cdot)\|P_{\bZ|Y}(\cdot|+1)\right)
    &=\int_{\R^{d}}\frac{\calZ_{+\theta_{0}}\sum_{y'\in\{\pm1\}}\sigma(\langle \bz,\btheta_{1}\rangle+y'\theta_{0})}{\calZ \sigma(\langle \bz,\btheta_{1}\rangle+\theta_{0})}P_{\bZ}\left(\diff \bz\right)-1\\
    &\le 2\frac{\calZ_{+\theta_{0}}}{\calZ}-1\le 1.
\end{align*}
In addition, the inequality $\sigma(t+\theta_{0})+\sigma(t-\theta_{0})\le 2$ gives
\begin{align*}
    &\chi^{2}\left(P_{\bZ}(\cdot)\|P_{\bZ|Y}(\cdot|-1)\right)\\
    &=\int_{\R^{d}}\frac{\calZ_{-\theta_{0}}\sum_{y'\in\{\pm1\}}\sigma(\langle \bz,\btheta_{1}\rangle+y'\theta_{0})}{\calZ \sigma(\langle \bz,\btheta_{1}\rangle-\theta_{0})}P_{\bZ}\left(\diff \bz\right)-1\\
    &\le 2e^{-\theta_{0}}M_{\bX}(\btheta_{1})\int_{\R^{d}}\frac{1}{\sigma(\langle \bz,\btheta_{1}\rangle-\theta_{0})}P_{\bZ}\left(\diff \bz\right)-1\\
    &\le 2e^{-\theta_{0}}M_{\bX}(\btheta_{1})\int_{\R^{d}}\exp\left(-\langle\bz,\btheta_{1}\rangle+\theta_{0}\right)P_{\bZ}\left(\diff \bz\right)-1\\
    &= 2e^{-\theta_{0}}M_{\bX}(\btheta_{1})\int_{\R^{d}}\exp\left(-\langle\bz,\btheta_{1}\rangle+\theta_{0}\right)\sum_{y\in\{\pm1\}}\sigma(\langle\bz,\btheta_{1}\rangle+y\theta_{0})\calZ^{-1}e^{-U(\bz)}\diff \bz-1\\
    &\le 4e^{-\theta_{0}}M_{\bX}(\btheta_{1})\int_{\R^{d}}\exp\left(-\langle\bz,\btheta_{1}\rangle+\theta_{0}\right)\calZ^{-1}e^{-U(\bz)}\diff \bz-1\\
    &= 4M_{\bX}^{2}(\btheta_{1})-1.
\end{align*}
Therefore, we obtain
\begin{equation*}
    K_{\chi^{2}}=\max_{y\in\{\pm1\}}\chi^{2}\left(P_{\bZ}(\cdot)\|P_{\bZ|Y}(\cdot|y)\right)\le 4M_{\bX}^{2}(\btheta_{1})-1;
\end{equation*}
here, we use $M_{\bX}(\btheta_{1})=\E[\exp(\langle\btheta_{1},\bX\rangle)]\ge \exp(\E[\langle\btheta_{1},\bX\rangle])=\exp(0)=1$ by the evenness of $U$.
For the other constants, we have
\begin{equation*}
    K_{\textnormal{V}}\le 4e^{-\theta_{0}}M_{\bX}(\btheta_{1}),\ K_{U}= \frac{\calZ}{\calZ_{-\theta_{0}}}\le 2e^{\theta_{0}}
\end{equation*}

Hence, we obtain the following log-Sobolev inequality by Theorem \ref{thm:functional} ($c=c^{\ast}=2$) and Lemma \ref{lem:tensor}: for any $f\in\calA_{n}$,
\begin{equation*}
    \Ent(f^{2}((\bZ_{i},Y_{i})_{i=1}^{n}))\le 2\E[\Gamma_{\textnormal{iii}}(f)((\bZ_{i},Y_{i})_{i=1}^{n})],
\end{equation*}
where
\begin{align*}
    \Gamma_{\textnormal{iii}}(f)&:= \left(\left(1+\frac{1}{2}\log\left(2e^{\theta_{0}}\right)\right)K_{\LogSobolev}\left(1+2(4M_{\bX}^{2}(\btheta_{1})-1)\right)+2K_{\LogSobolev}\right)\sum_{i=1}^{n}\Gamma_{\bZ_{i}}(f)\\
    &\quad+8e^{-\theta_{0}}M_{\bX}(\btheta_{1})\left(1+\frac{1}{2}\log\left( 2e^{\theta_{0}}\right)\right)\sum_{i=1}^{n}\Gamma_{Y_{i}}(f)\\
    &\le K_{\LogSobolev}\frac{e+\theta_{0}}{2}\left(\frac{1}{2}+8M_{\bX}^{2}(\btheta_{1})\right)\sum_{i=1}^{n}\Gamma_{\bZ_{i}}(f)+4e^{-\theta_{0}}M_{\bX}(\btheta_{1})\left(e+\theta_{0}\right)\sum_{i=1}^{n}\Gamma_{Y_{i}}(f);
\end{align*}
here, we use $2+\log2\le e$ and
\begin{align*}
    &\left(1+\frac{1}{2}\log\left(2e^{\theta_{0}}\right)\right)\left(8M_{\bX}^{2}(\btheta_{1})-1\right)+2\le \left(\frac{e}{2}+\frac{\theta_{0}}{2}\right)\left(8M_{\bX}^{2}(\btheta_{1})-1\right)+2\\
    &\le \left(\frac{e}{2}+\frac{\theta_{0}}{2}\right)\left(8M_{\bX}^{2}(\btheta_{1})-1+\frac{4}{e}\right)\le \left(\frac{e}{2}+\frac{\theta_{0}}{2}\right)\left(8M_{\bX}^{2}(\btheta_{1})+\frac{1}{2}\right).
\end{align*}

Therefore, using Equations \eqref{eq:cdc:z} and \eqref{eq:cdc:y} and Lemma \ref{lem:herbst}, we obtain that for all $\delta\in(0,1]$, with probability at least $1-\delta$,
    \begin{align*}
        &\sup_{(\bw,b)\in\calT}(\Risk(\bw,b)-\Risk_{n}(\bw,b))- \E\left[\sup_{(\bw,b)\in\calT}(\Risk(\bw,b)-\Risk_{n}(\bw,b))\right]\\
        &\le\max\left\{\sqrt{\frac{L^{2}\left(e+\theta_{0}\right)\left[K_{\LogSobolev}\left(1/2+8M_{\bX}^{2}(\btheta_{1})\right)R_{\bw}^{2}+4e^{1-\theta_{0}}M_{\bX}(\btheta_{1})R_{b}^{2}\right]\log(\delta^{-1})}{n}},\frac{4LR_{b}\log(\delta^{-1})}{n}\right\}.
    \end{align*}
    This is the desired conclusion.
\end{proof}

\begin{proof}[Proof of Proposition \ref{prop:concentration:3}]
    We only prove (ii) as the proof of (i) is analogous and simpler.
    
    We assume that $\theta_{0}\ge0$ without loss of generality.
    Using Equation \ref{eq:loglikdiff}, we obtain
    \begin{align*}
        \calZ_{+\theta_{0}}&=\int_{\R^{d}}g(\langle \bz,\btheta_{1}\rangle+\theta_{0})\exp\left(-U(\bz)\right)\diff \bz
        =g(+\theta_{0})\int_{\R^{d}}\frac{g(\langle \bz,\btheta_{1}\rangle+\theta_{0})}{g(+\theta_{0})}\exp\left(-U(\bz)\right)\diff \bz\\
        &\le g(+\theta_{0})\int_{\R^{d}}\exp\left(\max\{0,\langle\bz,G\btheta_{1}\rangle\}\right)\exp\left(-U(\bz)\right)\diff \bz
        =g(+\theta_{0})M_{\bX}^{\textnormal{num}}(G\btheta_{1})\calZ.
    \end{align*}
    Similarly, we have
    \begin{equation*}
        \calZ_{+\theta_{0}}\ge g(+\theta_{0})M_{\bX}^{\textnormal{den}}(G\btheta_{1})\calZ,\ 
        \calZ_{-\theta_{0}}\le g(-\theta_{0})M_{\bX}^{\textnormal{num}}(G\btheta_{1})\calZ,\ 
        \calZ_{-\theta_{0}}\ge g(-\theta_{0})M_{\bX}^{\textnormal{den}}(G\btheta_{1})\calZ.
    \end{equation*}
    In total, $\calZ_{+\theta_{0}}/(g(+\theta_{0})\calZ),\calZ_{-\theta_{0}}/(g(-\theta_{0})\calZ)\in[M_{\bX}^{\textnormal{den}}(G\btheta_{1}),M_{\bX}^{\textnormal{den}}(G\btheta_{1})]$.
    Then, since $g(t-\theta_{0})/g(t+\theta_{0})\le 1$ and $g(t+\theta_{0})+g(t-\theta_{0})\le 2$ for any $t\in\R$,
    \begin{align*}
        &\chi^{2}(P_{\bZ}(\cdot)\|P_{\bZ|Y}(\cdot|+1))\\
        &=\int_{\R^{d}}\frac{\calZ_{+\theta_{0}}\sum_{y'\in\{\pm1\}}g(\langle\bz,\btheta_{1}\rangle+y'\theta_{0})}{\calZ g(\langle\bz,\btheta_{1}\rangle+\theta_{0})}P_{\bZ}(\diff\bz)-1\\
        &\le g(+\theta_{0})M_{\bX}^{\textnormal{num}}(G\btheta_{1})\int_{\R^{d}}\left(1+\frac{g(\langle\bz,\btheta_{1}\rangle-\theta_{0})}{g(\langle\bz,\btheta_{1}\rangle+\theta_{0})}\right)P_{\bZ}(\diff\bz)-1\\
        &\le g(+\theta_{0})M_{\bX}^{\textnormal{num}}(G\btheta_{1})\int_{\R^{d}}\left(1+\frac{g(-\theta_{0})}{g(+\theta_{0})}\frac{g(\langle\bz,\btheta_{1}\rangle-\theta_{0})/g(-\theta_{0})}{g(\langle\bz,\btheta_{1}\rangle+\theta_{0})/g(+\theta_{0})}\right)P_{\bZ}(\diff\bz)-1\\
        &\le M_{\bX}^{\textnormal{num}}(G\btheta_{1})\int_{\R^{d}}\left(g(+\theta_{0})+g(-\theta_{0})\exp\left(2\langle\bz,G\btheta_{1}\rangle\right)\right)P_{\bZ}(\diff\bz)-1\\
        &= M_{\bX}^{\textnormal{num}}(G\btheta_{1})\int_{\R^{d}}\left(g(+\theta_{0})+g(-\theta_{0})\exp\left(2\langle\bz,G\btheta_{1}\rangle\right)\right)\sum_{y\in\{\pm1\}}g(\langle\bz,\btheta_{1}\rangle+y\theta_{0})\calZ^{-1}e^{-U(\bz)}\diff \bz-1\\
        &\le 2M_{\bX}^{\textnormal{num}}(G\btheta_{1})\left(g(+\theta_{0})+g(-\theta_{0})M_{\bX}^{\textnormal{num}}(2G\btheta_{1})\right)-1.
    \end{align*}
    Moreover,
   \begin{align*}
        \chi^{2}(P_{\bZ}(\cdot)\|P_{\bZ|Y}(\cdot|-1))&=\int_{\R^{d}}\frac{\calZ_{-\theta_{0}}\sum_{y'\in\{\pm1\}}g(\langle\bz,\btheta_{1}\rangle+y'\theta_{0})}{\calZ g(\langle\bz,\btheta_{1}\rangle-\theta_{0})}P_{\bZ}(\diff\bz)-1\\
        &\le g(-\theta_{0})M_{\bX}^{\textnormal{num}}(G\btheta_{1})\int_{\R^{d}}\left(1+\frac{g(\langle\bz,\btheta_{1}\rangle+\theta_{0})}{g(\langle\bz,\btheta_{1}\rangle-\theta_{0})}\right)P_{\bZ}(\diff\bz)-1\\
        &\le g(-\theta_{0})M_{\bX}^{\textnormal{num}}(G\btheta_{1})\int_{\R^{d}}\left(1+\frac{g(+\theta_{0})}{g(-\theta_{0})}\frac{g(\langle\bz,\btheta_{1}\rangle+\theta_{0})/g(+\theta_{0})}{g(\langle\bz,\btheta_{1}\rangle-\theta_{0})/g(-\theta_{0})}\right)P_{\bZ}(\diff\bz)-1\\
        &\le M_{\bX}^{\textnormal{num}}(G\btheta_{1})\int_{\R^{d}}\left(g(-\theta_{0})+g(+\theta_{0})\exp\left(2\langle\bz,G\btheta_{1}\rangle\right)\right)P_{\bZ}(\diff\bz)-1\\
        &\le 2M_{\bX}^{\textnormal{num}}(G\btheta_{1})\left(g(-\theta_{0})+g(+\theta_{0})M_{\bX}^{\textnormal{num}}(2G\btheta_{1})\right)-1.
    \end{align*}
    Therefore, since $g(+\theta_{0})\ge g(-\theta_{0})$ and $M_{\bX}^{\textnormal{num}}(\bt)\ge 1$ for $\bt\in\R^{d}$,
    \begin{equation*}
        K_{\chi^{2}}\le 2M_{\bX}^{\textnormal{num}}(G\btheta_{1})\left(g(-\theta_{0})+g(+\theta_{0})M_{\bX}^{\textnormal{num}}(2G\btheta_{1})\right)-1\le 2M_{\bX}^{\textnormal{num}}(G\btheta_{1})M_{\bX}^{\textnormal{num}}(2G\btheta_{1})-1.
    \end{equation*}
    We also have
    \begin{equation*}
        K_{\textnormal{V}}=4g(+\theta_{0})g(-\theta_{0})M_{\bX}^{\textnormal{num}}(G\btheta_{1}),\ K_{\textnormal{U}}=\frac{1}{g(-\theta_{0})M_{\bX}^{\textnormal{den}}(G\btheta_{1})}.
    \end{equation*}

    Hence, Theorem \ref{thm:functional} ($c=c^{\ast}=2$) along with Lemma \ref{lem:tensor} gives the following log-Sobolev inequality: for arbitrary $f\in\calA_{n}$,
    \begin{equation*}
        \Ent(f^{2}((\bZ_{i},Y_{i})_{i=1}^{n}))\le 2\E[\Gamma_{\textnormal{iv}}(f)((\bZ_{i},Y_{i})_{i=1}^{n})],
    \end{equation*}
    where
    \begin{align*}
        \Gamma_{\textnormal{iv}}(f)&:= \left(\left(1-\frac{1}{2}\log\left(g(-\theta_{0})M_{\bX}^{\textnormal{den}}(G\btheta_{1})\right)\right)K_{\LogSobolev}\left(4M_{\bX}^{\textnormal{num}}(G\btheta_{1})M_{\bX}^{\textnormal{num}}(2G\btheta_{1})-1\right)+2K_{\LogSobolev}\right)\sum_{i=1}^{n}\Gamma_{\bZ_{i}}(f)\\
        &\quad+8g(+\theta_{0})g(-\theta_{0})M_{\bX}^{\textnormal{num}}(G\btheta_{1})\left(1-\frac{1}{2}\log\left(g(-\theta_{0})M_{\bX}^{\textnormal{den}}(G\btheta_{1})\right)\right)\sum_{i=1}^{n}\Gamma_{Y_{i}}(f)\\
        &\le K_{\LogSobolev}\left(1-\frac{1}{2}\log\left(g(-\theta_{0})M_{\bX}^{\textnormal{den}}(G\btheta_{1})\right)\right)\left(5M_{\bX}^{\textnormal{num}}(G\btheta_{1})M_{\bX}^{\textnormal{num}}(2G\btheta_{1})\right)\sum_{i=1}^{n}\Gamma_{\bZ_{i}}(f)\\
        &\quad+8g(+\theta_{0})g(-\theta_{0})M_{\bX}^{\textnormal{num}}(G\btheta_{1})\left(1-\frac{1}{2}\log\left(g(-\theta_{0})M_{\bX}^{\textnormal{den}}(G\btheta_{1})\right)\right)\sum_{i=1}^{n}\Gamma_{Y_{i}}(f).
    \end{align*}
    
    Therefore, Equations \eqref{eq:cdc:z} and \eqref{eq:cdc:y} and Lemma \ref{lem:herbst} give that for all $\delta\in(0,1]$, with probability at least $1-\delta$,
    \begin{align*}
        \sup_{(\bw,b)\in\calT}(&\Risk(\bw,b)-\Risk_{n}(\bw,b))-\E\left[\sup_{(\bw,b)\in\calT}(\Risk(\bw,b)-\Risk_{n}(\bw,b))\right]\\
        \le \max&\left\{
        \sqrt{\frac{2L^{2}\log(\delta^{-1})}{n}}\sqrt{1+\frac{1}{2}\left(\log\frac{1}{g(-|\theta_{0}|)}+\log\frac{1}{M_{\bX}^{\textnormal{den}}(G\btheta_{1})}\right)}\right.\\
            &\times\left.\sqrt{5K_{\LogSobolev}M_{\bX}^{\textnormal{num}}(G\btheta_{1})M_{\bX}^{\textnormal{num}}(2G\btheta_{1})R_{\bw}^{2}+4eg(+\theta_{0})g(-\theta_{0})M_{\bX}^{\textnormal{num}}(G\btheta_{1})R_{b}^{2}},\frac{4LR_{b}\log(\delta^{-1})}{n}\right\}.
    \end{align*}
        Hence, the statement holds true.
\end{proof}

\begin{proof}[Proof of Proposition \ref{prop:concentration:4}]
    We give the proof of (ii) and omit that of (i), as that of (i) can be given with fewer steps by applying Theorem \ref{thm:functional} with $c=1+1/\sqrt{e^{5|\theta_{0}|}/\tilde{M}_{\bX}(\btheta_{1})}$ and $c^{\ast}=1+\sqrt{e^{5|\theta_{0}|}/\tilde{M}_{\bX}(\btheta_{1})}$.
    
    Let $\theta_{0}\ge 0$ without loss of generality. 
    Since $\sigma(t)\le \exp(t)$ for any $t\in\R$,
    \begin{align*}
        \calZ_{+\theta_{0}}&=\int_{\R^{d}} \sigma(\langle\bz,\btheta_{1}\rangle+\theta_{0})\exp(-U(\bz))\diff\bz\\
        &\le \int_{\langle \bz,\btheta_{1}\rangle \ge 0}\exp(-U(\bz))\diff\bz+\int_{\langle \bz,\btheta_{1}\rangle < 0}\sigma(\langle \bz,\btheta_{1}\rangle +\theta_{0})\exp(-U(\bz))\diff\bz\\
        &= \int_{\langle \bz,\btheta_{1}\rangle \ge 0}\exp(-U(\bz))\diff\bz+\exp(+\theta_{0})\int_{\langle \bz,\btheta_{1}\rangle < 0}\exp(\langle \bz,\btheta_{1}\rangle)\exp(-U(\bz))\diff\bz\\
        &\le \frac{\calZ}{2}\left(1+ \exp(+\theta_{0})\E\left[\exp\left(-|\langle \bX,\btheta_{1}\rangle|\right)\right]\right),\\
        \calZ_{-\theta_{0}}&\le \frac{\calZ}{2}\left(1+ \exp(-\theta_{0})\E\left[\exp\left(-|\langle \bX,\btheta_{1}\rangle|\right)\right]\right),
    \end{align*}
    and by $\sigma(t)\ge 1/2$ for $t\ge0$,
    \begin{align*}
        \calZ_{-\theta_{0}}&=\int_{\R^{d}} \sigma(\langle\bz,\btheta_{1}\rangle-\theta_{0})\exp(-U(\bz))\diff\bz
        \ge \int_{\langle \bz,\btheta_{1}\rangle -\theta_{0}\ge 0}\sigma(\langle \bz,\btheta_{1}\rangle -\theta_{0})\exp(-U(\bz))\diff\bz\\
        &\ge \frac{\calZ}{2}\Pr\left(\langle \bX,\btheta_{1}\rangle \ge \theta_{0}\right).
    \end{align*}
    We obtain
    \begin{align*}
        \chi^{2}(P_{\bZ}(\cdot)\|P_{\bZ|Y}(\cdot|+1))&=\int_{\R^{d}}\frac{\calZ_{+\theta_{0}}\sum_{y'\in\{\pm1\}}\sigma(\langle\bz,\btheta_{1}\rangle+y'\theta_{0})}{\calZ \sigma(\langle\bz,\btheta_{1}\rangle+\theta_{0})}P_{\bZ}(\diff\bz)-1\\
        &\le \frac{1}{2}\left(1+ \exp(+\theta_{0})/\tilde{M}_{\bX}(\btheta_{1})\right)\int_{\R^{d}}\left(1+\frac{\sigma(\langle\bz,\btheta_{1}\rangle-\theta_{0})}{\sigma(\langle\bz,\btheta_{1}\rangle+\theta_{0})}\right)P_{\bZ}(\diff\bz)-1\\
        &\le 1+ \exp(+\theta_{0})/\tilde{M}_{\bX}(\btheta_{1})-1\\
        &\le \exp(+\theta_{0})/\tilde{M}_{\bX}(\btheta_{1})
    \end{align*}
    by the monotonicity of $\sigma$, and
    \begin{align*}
        \chi^{2}(P_{\bZ}(\cdot)\|P_{\bZ|Y}(\cdot|-1))&=\int_{\R^{d}}\frac{\calZ_{-\theta_{0}}\sum_{y'\in\{\pm1\}}\sigma(\langle\bz,\btheta_{1}\rangle+y'\theta_{0})}{\calZ \sigma(\langle\bz,\btheta_{1}\rangle-\theta_{0})}P_{\bZ}(\diff\bz)-1\\
        &\le \frac{1}{2}\left(1+ \exp(-\theta_{0})/\tilde{M}_{\bX}(\btheta_{1})\right)\int_{\R^{d}}\left(1+\frac{\sigma(\langle\bz,\btheta_{1}\rangle+\theta_{0})}{\sigma(\langle\bz,\btheta_{1}\rangle-\theta_{0})}\right)P_{\bZ}(\diff\bz)-1\\
        &\le \left(1+ \exp(-\theta_{0})/\tilde{M}_{\bX}(\btheta_{1})\right)-1\\
        &\quad+\frac{1}{2}e^{+\theta_{0}}(e^{+4\theta_{0}}-1)/\tilde{M}_{\bX}(\btheta_{1})\left(1+ \exp(-\theta_{0})/\tilde{M}_{\bX}(\btheta_{1})\right)\\
        &=(1/\tilde{M}_{\bX}(\btheta_{1}))\left(e^{-\theta_{0}}+\frac{1}{2}e^{+\theta_{0}}(e^{+4\theta_{0}}-1)\left(1+ e^{-\theta_{0}}/\tilde{M}_{\bX}(\btheta_{1})\right)\right)
    \end{align*}
    because
    \begin{align*}
        \frac{\sigma(t+\theta_{0})}{\sigma(t-\theta_{0})}-1
        &=\frac{\sigma(t+\theta_{0})}{\sigma(t-\theta_{0})}-1=\frac{1+e^{-t}e^{+\theta_{0}}}{1+e^{-t}e^{-\theta_{0}}}-1=\frac{e^{-t}(e^{+\theta_{0}}-e^{-\theta_{0}})}{1+e^{-t}e^{-\theta_{0}}}=(e^{+2\theta_{0}}-1)\frac{e^{-t-\theta_{0}}}{1+e^{-t-\theta_{0}}}\\
        &=(e^{+2\theta_{0}}-1)\frac{1}{1+e^{t+\theta_{0}}}=(e^{+2\theta_{0}}-1)\sigma(-t-\theta_{0}),
    \end{align*}
    and thus the facts $\sigma(t)\sigma(-t)\le \sigma(-|t|)$ and $\sigma(t)\le \exp(t)$ give
    \begin{align*}
         &\int_{\R^{d}}\frac{\sigma(t+\theta_{0})}{\sigma(t-\theta_{0})}P_{\bZ}(\diff\bz) -1\\
         &=(e^{+2\theta_{0}}-1)\int_{\R^{d}}\sigma(-\langle\bz,\btheta_{1}\rangle-\theta_{0})P_{\bZ}(\diff\bz)\\
         &=(e^{+2\theta_{0}}-1)\int_{\R^{d}}\sigma(-\langle\bx,\btheta_{1}\rangle-\theta_{0})\left(\sigma(\langle\bx,\btheta_{1}\rangle+\theta_{0})+\sigma(\langle\bx,\btheta_{1}\rangle-\theta_{0})\right)P_{\bX}(\diff\bx)\\
         &\le(e^{+2\theta_{0}}-1)\int_{\R^{d}}\sigma(-\langle\bx,\btheta_{1}\rangle-\theta_{0}) (1+e^{+2\theta_{0}})\sigma(\langle\bx,\btheta_{1}\rangle+\theta_{0})P_{\bX}(\diff\bx)\\
         &\le(e^{+4\theta_{0}}-1)\int_{\R^{d}}\frac{1}{1+\exp(-\langle\bx,\btheta_{1}\rangle-\theta_{0})}\frac{1}{1+\exp(+\langle\bx,\btheta_{1}\rangle+\theta_{0})}P_{\bX}(\diff\bx)\\
         &\le(e^{+4\theta_{0}}-1)\int_{\R^{d}}\sigma(-|\langle\bx,\btheta_{1}\rangle+\theta_{0}|)P_{\bX}(\diff\bx)\\
         &\le (e^{+4\theta_{0}}-1)\E\left[\exp(-|\langle\bX,\btheta_{1}\rangle+\theta_{0}|)\right]\\
         &\le e^{+\theta_{0}}(e^{+4\theta_{0}}-1)\E\left[\exp(-|\langle\bX,\btheta_{1}\rangle|\right].
    \end{align*}
    In total, since both $e^{-\theta_{0}}\le 1$ and $1/\tilde{M}_{\bX}(\btheta_{1})\le 1$ by definition, we have
    \begin{align*}
        K_{\chi^{2}}&\le e^{+\theta_{0}}/\tilde{M}_{\bX}(\btheta_{1})\left(1+\frac{1}{2}(e^{+4\theta_{0}}-1)\left(1+ e^{-\theta_{0}}/\tilde{M}_{\bX}(\btheta_{1})\right)\right)\\
        &\le 
        e^{+\theta_{0}}/\tilde{M}_{\bX}(\btheta_{1})\left(1+(e^{+4\theta_{0}}-1)\right) \le e^{+5\theta_{0}}/\tilde{M}_{\bX}(\btheta_{1}),
    \end{align*}
    and
    \begin{align*}
        K_{\textnormal{U}}&\le 1,\ K_{\textnormal{V}}\le \frac{2}{p_{\btheta_{1},\theta_{0}}}.
    \end{align*}
    Therefore, the following log-Sobolev inequality holds by Theorem \ref{thm:functional} ($c=c^{\ast}=2$) and Lemma \ref{lem:tensor}: for any $f\in\calA_{n}$,
    \begin{equation*}
        \Ent(f^{2}((\bZ_{i},Y_{i})_{i=1}^{n}))\le 2\E[\Gamma_{\textnormal{v}}(f)((\bZ_{i},Y_{i})_{i=1}^{n})],
    \end{equation*}
    where
    \begin{align*}
        \Gamma_{\textnormal{v}}(f)
        &:= K_{\LogSobolev}\left(\left(1+\frac{1}{2}\log\left(\frac{2}{p_{\btheta_{1},\theta_{0}}}\right)\right)\left(1+2e^{+5\theta_{0}}/\tilde{M}_{\bX}(\btheta_{1})\right)+2\right)\sum_{i=1}^{n}\Gamma_{\bZ_{i}}(f)\\
        &\quad+2\left(1+\frac{1}{2}\log\left(\frac{2}{p_{\btheta_{1},\theta_{0}}}\right)\right)\sum_{i=1}^{n}\Gamma_{Y_{i}}(f)\\
        &\le K_{\LogSobolev}\left(\frac{e}{2}+\frac{1}{2}\log\left(\frac{1}{p_{\btheta_{1},\theta_{0}}}\right)\right)\left(\frac{5}{2}+2e^{+5\theta_{0}}/\tilde{M}_{\bX}(\btheta_{1})\right)\sum_{i=1}^{n}\Gamma_{\bZ_{i}}(f)\\
        &\quad+\left(e+\log\left(\frac{1}{p_{\btheta_{1},\theta_{0}}}\right)\right)\sum_{i=1}^{n}\Gamma_{Y_{i}}(f),
    \end{align*}
    where we use $1+(1/2)\log(2)\le e/2$ and $1+4/e\le 5/2$.
    Therefore, as the proofs of the previous propositions above, for all $\delta\in(0,1]$, with probability at least $1-\delta$,
    \begin{align*}
        &\sup_{(\bw,b)\in\calT}(\Risk(\bw,b)-\Risk_{n}(\bw,b))- \E\left[\sup_{(\bw,b)\in\calT}(\Risk(\bw,b)-\Risk_{n}(\bw,b))\right]\\
        &\le\max\left\{\sqrt{\frac{L^{2}\left(e+\log\left(\frac{1}{p_{\btheta_{1},\theta_{0}}}\right)\right)\left(2K_{\LogSobolev}\left(\frac{5}{4}+e^{5|\theta_{0}|}/\tilde{M}_{\bX}(\btheta_{1})\right)R_{\bw}^{2}+eR_{b}^{2}\right)\log(\delta^{-1})}{n}},\frac{4LR_{b}\log(\delta^{-1})}{n}\right\}.
    \end{align*}
        Hence, the statement holds true.
\end{proof}

\end{document}